%% file: root.tex
\documentclass[lettersize,journal]{IEEEtran}
\usepackage{amsmath,amssymb,amsfonts}
\usepackage{graphicx}
\usepackage{pifont}
\usepackage{wrapfig}
\usepackage{amsthm}
\usepackage{xcolor}
\usepackage[figuresright]{rotating}

\usepackage{graphicx}
\graphicspath{{/}{fig/}}

\usepackage{array}
\usepackage{textcomp}
\usepackage{xcolor}
\usepackage{multirow}
\usepackage{booktabs}

\usepackage{mathtools}
\usepackage{breqn}
\usepackage{float}

\usepackage{pgfplots}
\pgfplotsset{compat=1.7}
\usepgfplotslibrary{groupplots}

\usepackage{caption}
\usepackage{subcaption}
\usepackage{enumerate}

\usepackage{multirow,tabularx}
\usepackage{hyperref}
\usepackage{flushend}
\usepackage[vlined, ruled, shortend]{algorithm2e}

\usepackage{epstopdf}
\newlength\figureheight
\newlength\figurewidth
\SetAlCapNameFnt{\footnotesize}
\SetAlCapFnt{\footnotesize}

\DeclareMathOperator*{\argmax}{arg\,max}

\usepackage{algorithmicx}
\usepackage{algpseudocode}
\usepackage{tabularx}
\usepackage{bm}
\newtheorem{assumption}{Assumption}
\newtheorem{proposition}{Proposition}
\newtheorem{remark}{Remark}
\newtheorem{theorem}{Theorem}
\usepackage{url}

\usepackage[T1]{fontenc}
\usepackage{microtype}
\usepackage{graphicx}
\usepackage{epstopdf}
\newtheorem{definition}{Definition}

\algnewcommand{\Inputs}[1]{%
  \State \textbf{Inputs:}
  \Statex \hspace*{\algorithmicindent}\parbox[t]{.8\linewidth}{\raggedright #1}
}
\algnewcommand{\Initialize}[1]{%
  \State \textbf{Initialize:}
  \Statex \hspace*{\algorithmicindent}\parbox[t]{.8\linewidth}{\raggedright #1}
}

\title{
Exploiting Local Observations for Robust Robot Learning
}

\author{Wenshuai Zhao$^{1*}$,  Eetu-Aleksi Rantala$^{2*}$, Sahar Salimpour$^{2}$, Zhiyuan Li$^{1}$, \\Joni Pajarinen$^{1}$ and Jorge Pe\~{n}a-Queralta$^{3}$% <-this % stops a space
\thanks{$*$ These authors contributed equally to this work.}% <-this % stops a space
\thanks{$^{1}$ Wenshuai Zhao, Zhiyuan Li and Joni Pajarinen are with the Department of Electrical Engineering and Automation, Aalto University, Finland.}
\thanks{$^{2}$ Eetu-Aleksi Rantala and Sahar Salimpour are with the Turku Intelligent Embedded and Robotic Systems (TIERS) Lab, University of Turku, Finland.}%
\thanks{$^{3}$ Jorge Pe\~{n}a-Queralta is with the Institute of Robotics and Intelligent Systems, ETH Zurich, Switzerland.}%
}

\begin{document}

% \markboth{Journal of \LaTeX\ Class Files,~Vol.~18, No.~9, September~2020}%
% {How to Use the IEEEtran \LaTeX \ Templates}

\maketitle
% \thispagestyle{empty}
% \pagestyle{empty}

%%%%%%%%%%%%%%%%%%%%%%%%%%%%%%%%%%%%%%%%%%%%%%%%%%%%%%%%%%%%%%%%%%%%%%%%%%%%%%%%
% \begin{abstract}

% In many robot tasks, controllers can be designed centrally or decentrally, using either single-agent reinforcement learning (SARL) or multi-agent reinforcement learning (MARL). However, the relationship between these two paradigms is not well-studied. This work aims to systematically investigate the robustness and performance of SARL and MARL in the same task. We first analytically show that independent Gaussian policies optimized by policy-gradient based SARL and MARL are equivalent under full-state observations. Following, we empirically show that in certain inherently single-agent tasks, we can use multiple agents to control a robot such that each agent only has access to local observations. Since in these cases an agent does not depend on full state information multi-agent policies can provide additional robustness to perturbations and failures. Experiments on an illustrative decentralized control task and a mobile manipulation task with a real robot show that multiple agents with access to local observations outperform a single agent when parts of the system fail.

% \end{abstract}

\begin{abstract}
While many robotic tasks can be addressed using either centralized single-agent control with full state observation or decentralized multi-agent control, clear criteria for choosing between these approaches remain underexplored. This paper systematically investigates how multi-agent reinforcement learning (MARL) with local observations can improve robustness in complex robotic systems compared to traditional centralized control. Through theoretical analysis and empirical validation, we show that in certain tasks, decentralized MARL can achieve performance comparable to centralized methods while exhibiting greater resilience to perturbations and agent failures. By analytically demonstrating the equivalence of single-agent reinforcement learning (SARL) and MARL under full observability, we identify observability as the critical factor distinguishing the two paradigms. We further derive bounds quantifying performance degradation under external perturbations for locally observable policies. Empirical results on standard MARL benchmarks confirm that MARL with limited observations can maintain competitive performance. Finally, real-world experiments with a mobile manipulator demonstrate that decentralized MARL controllers achieve markedly improved robustness to agent malfunctions and environmental disturbances relative to centralized baselines. Together, these findings highlight MARL with local observations as a robust and practical alternative to conventional centralized control in complex robotic systems.

\end{abstract}


%%%%%%%%%%%%%%%%%%%%%%%%%%%%%%%%%%%%%%%%%%%%%%%%%%%%%%%%%%%%%%%%%%%%%%%%%%%%%%%%
\section{Introduction}\label{sec:introduction}

Multi-agent reinforcement learning (MARL) provides a paradigm where multiple agents learn simultaneously and has been shown promising in simulated tasks and video games~\cite{yu2022surprising, vinyals2019grandmaster}. Nevertheless, many existing MARL benchmark tasks can also be addressed by single-agent reinforcement learning (SARL) with a centralized controller for all the agents given full state observation~\cite{terry2021pettingzoo}. Conversely, certain single-agent-like tasks may be reframed as multi-agent learning problems~\cite{de2020deep, peng2021facmac}. For instance, in a classic legged-robot locomotion task, typically controlled by a centralized controller managing all joints, joints can be divided among different agents, allowing MARL algorithms to learn decentralized controllers. Despite this flexibility, the differences between these two approaches and the criteria for choosing the optimal method for new robotic tasks remain unclear. 

\begin{figure*}[t]
    \centering
    \includegraphics[width=.85\textwidth]{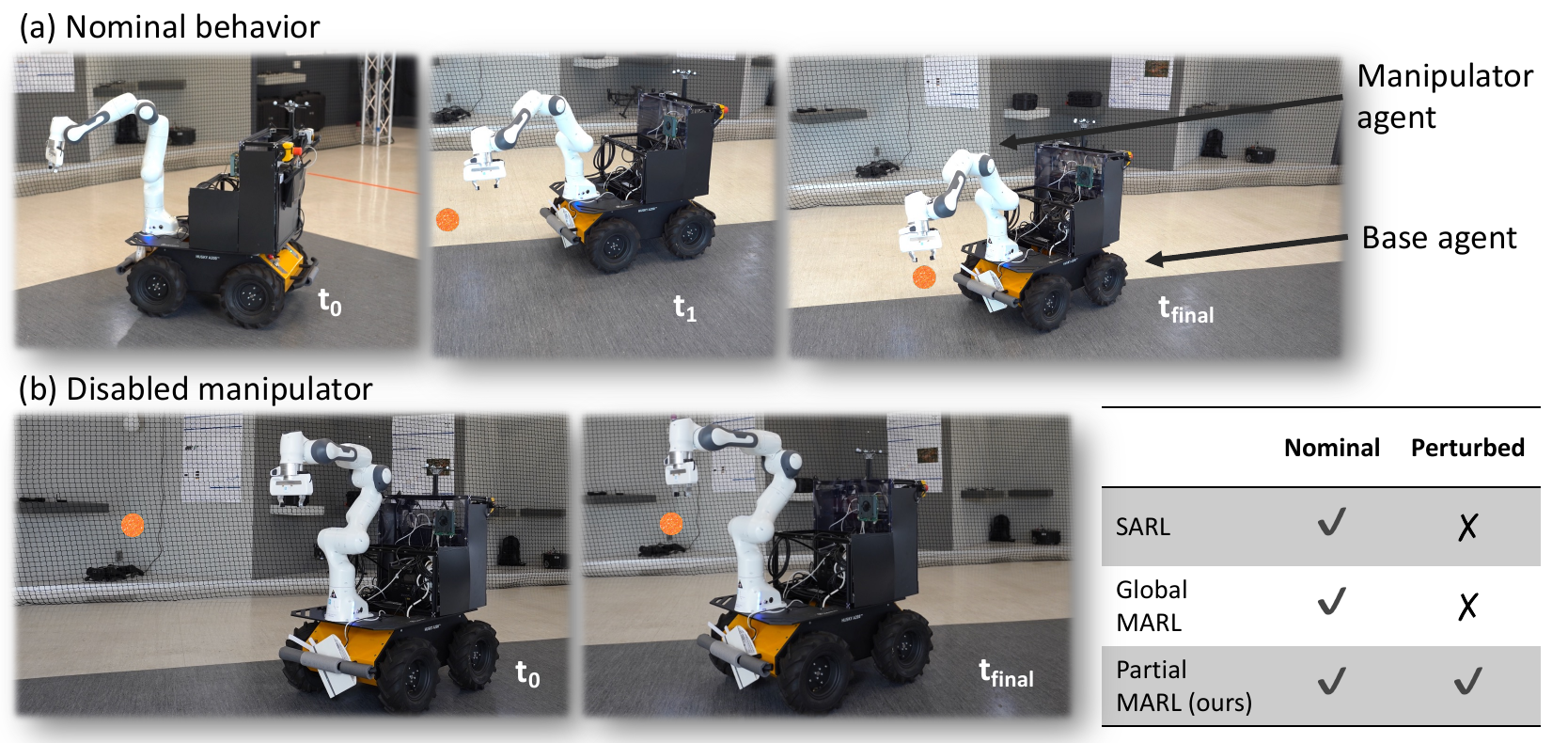}
    \caption{The real-world mobile manipulator experiments examine the robustness of learned controllers via three different methods: (i) \textit{SARL}, (ii) \textit{MARL (global)} and (iii) \textit{MARL (partial)}. In the setting shown in Figure (a), both the mobile base and the manipulator arm work normally while in (b) the manipulator is disabled to test the robustness of learned controllers. The results in different settings in the bottom right table show that although all three methods can work well with nominal behaviors, only \textit{MARL (partial)} succeeds when the manipulator is disabled and perturbed, demonstrating improved robustness of the controller empowered by locally observed MARL.}
    \label{fig:topfig}
    % \vspace{-1em}
\end{figure*}

Decentralized control with local observations has been extensively studied to achieve performance comparable to centralized control in large-scale systems~\cite{lavaei2011decentralized, davison2020decentralized}. However, these studies often focus on mitigating the limitations of local information in decentralized controllers rather than leveraging this structure. Recent research has demonstrated that decentralized controllers for complex robotic systems can achieve similar or even superior performance and robustness compared to centralized controllers~\cite{schilling2020decentralized, tao2023multi, guo2023decentralized}. Nonetheless, these studies are typically confined to specific simulated tasks and do not provide a general investigation of design choices. 

In this paper, we argue that, for many robotic tasks, modeling a complex robot system as a collection of multiple agents and applying multi-agent reinforcement learning with local observations to learn decentralized controllers can not only achieve performance comparable to that of centralized controllers with full observations, but also enhance the robustness of the system against perturbations and agent failures. To the best of our knowledge, this is the first work to both theoretically and empirically investigate how the flexibility of MARL can be leveraged to improve robustness through the use of local observation settings.

Our contributions are summarized as follows:
\begin{itemize}
    \item [i)] We both theoretically and empirically investigate single-agent and multi-agent reinforcement learning under varying observation settings, identifying observability as the critical factor that differentiates the two frameworks.
    \item [ii)] Empirical results show that MARL with limited local observations performs comparably to full observability.
    \item [iii)] We derive an upper bound on the performance loss of locally observable policies under external perturbations.
    \item [iv)] In real-world experiments with a mobile manipulator, our decentralized MARL controllers with local observations show markedly greater robustness to agent failures and environmental disturbances, as illustrated in Fig.~\ref{fig:topfig}.
\end{itemize}

We believe this systematic investigation offers important insights for the design of robust and generalizable control strategies in complex robotic systems.

% Our contributions are as follows: (i) We analytically show the equivalence between SARL and MARL under global full observation conditions and show that the observability is the key factor to distinguish the two frameworks. (ii) Our empirical experiments on two common MARL tasks further indicate that MARL with limited local observations can achieve performance comparable to that with full observations. (iii) We provide an upper bound the performance degradation of locally observable policies when facing perturbations. (iv) In experiments with a real-world mobile manipulation robot, our decentralized MARL controllers with local observations exhibit improved robustness to agent failure and perturbations, as shown in Fig.~\ref{fig:topfig}. We believe this systematic investigation provides valuable insights for the design of general robotic system control.

% The remainder of this document is organized as follows. Section II introduces the state of the art in MARL and robotics use cases. Section III then describes the background concepts for Dec-MDP and Dec-POMDP. We introduce the equivalence between SARL and MARL in Section IV and empirically show the near-optimal performance of MARL agents with limited observations in Section V. Section VI provides an illustrative example to analytically show the robustness via decentralized PI controllers. Section VII describes our real robot experiments. We conclude the work and outline limitations in Section VIII.

\section{Related Work}
% We first introduce common MARL methods and then specify works dealing with partial observation. Finally, we discuss current works using MARL for robotic tasks.
\subsection{Multi-Agent Reinforcement Learning}

Deep multi-agent reinforcement learning has exhibited success in various tasks~\cite{gronauer2022multi, zhao2024optimistic, zhaolearning}. One notable example is the Multi-agent Mujoco~\cite{peng2021facmac} domain, which splits the joints of Mujoco robot tasks~\cite{brockman2016openai} into different agents and formulates it as a multi-agent cooperative task. Although such Mujoco tasks were traditionally used to benchmark single-agent RL methods, MARL approaches also show competitive performance~\cite{zhao2024optimistic, li2025agentmixer}. Therefore, it motivates us to investigate the connection between MARL and SARL paradigms and their applicability to complex robotic systems.

%methods and distinguish their respective advantage for robot learning. 

% A number of works try to overcome the non-stationarity problem in MARL through appealing properties in SARL such as the policy monotonic improvement guarantee~\cite{kuba2021trust, sun2022monotonic}. However, to the best of our knowledge, this work is the first to analytically show the policy gradient equivalence between SARL and MARL under two common assumptions, which supports that the strong performance of multi-agent policy gradient (MAPG) methods on various tasks is a direct result of equivalently to single-agent policy gradient (SAPG) methods. 

\subsection{Partially Observable MARL}

Existing research on partial observability in multi-agent reinforcement learning (MARL) has predominantly sought to mitigate its negative impact~\cite{li2025agentmixer}. Typical strategies include enhancing information sharing among agents~\cite{liu2023partially, li2025agentmixer} or approximating the full-state belief via mean-field methods~\cite{he2021many} and agent modeling~\cite{papoudakis2021agent}. In contrast, motivated by evidence that many multi-agent problems exhibit local independence and can attain near-optimal performance using only local information~\cite{lavaei2011decentralized, deweese2024locally}, we propose exploiting partial observability as an advantage rather than a limitation by leveraging the inherent structure of complex robotic tasks. On the theoretical side, prior work has shown that in certain subclasses of decentralized partially observable Markov decision processes (Dec-POMDPs), decentralized MARL with local observations can enhance tractability by capitalizing on local structures in observations~\cite{qu2022scalable} and interactions~\cite{witwicki2011abstracting, deweese2024locally}. Building on this foundation, our work further provides formal guarantees on bounded robustness under local observation settings.

\subsection{MARL for Robotics}

MARL methods have been naturally used in multi-robot control tasks~\cite{jana2022deep, leottau2018decentralized, zhao2020sim} due to their intrinsic decentralized control mechanism. However, some robot tasks can adopt either SARL or MARL methods to learn controllers. For example,  in bi-arm manipulation, studies such as~\cite{liu2021collaborative} and~\cite{ding2020distributed} employ MARL methods to learn decentralized controllers, whereas works like~\cite{chitnis2020efficient} and recent imitation learning studies~\cite{zhao2023learning, shi2023waypoint} develop a centralized controller for all arms. Most existing works predominantly use centralized controllers for mobile manipulation~\cite{yokoyama2023adaptive, wang2020learning}, and few studies have explored the comparative advantages of MARL over SARL when both approaches are applicable. The closest work to ours~\cite{tao2023multi} compares SARL and MARL methods for in-hand manipulation. However, unlike our approach, they allow the decentralized controllers to observe the neighbor agents' actions, which provides critical information for the agents to adapt to malfunctions compared to SARL baselines. In this paper, we instead seek the benefits of using less state information with MARL and comprehensively investigate the connection between SARL and MARL.

\section{Problem Formulation} \label{sec:background}

\begin{figure}[t]
    \centering
    % \vskip 0.1in
    \includegraphics[width=0.96\linewidth]{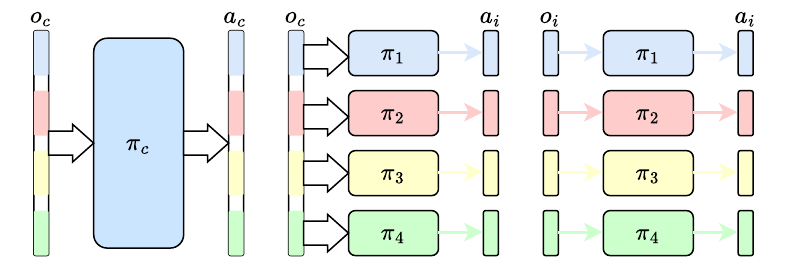}
    % \vskip -0.14in
    \makebox[\linewidth][c]{%
      \makebox[0.32\linewidth][c]{(a)}%
      \makebox[0.32\linewidth][c]{(b)}%
      \makebox[0.32\linewidth][c]{(c)}%
    }
    \caption{Three different configurations for designing controllers to perform the same task: (a) A centralized controller with full observation, denoted as $\pi_{c}(a_c \vert o_c)$, outputs the complete set of actions. (b) Decentralized controllers with full observations, $\pi_{i}(a_i \vert o_c)$, each output a subset of the overall action space. (c) Decentralized controllers with local observations, $\pi_{i}(a_i \vert o_i)$, produce respective subsets of actions based on partial, agent-specific observations.}
    % \vskip -0.15in
    \label{fig:pg}
\end{figure}

% \subsection{Decentralized Markov Decision Process}
We study the fully cooperative multi-agent sequential decision-making tasks which can be formulated as a \textit{decentralized Markov decision process} (Dec-MDP)~\cite{bernstein2002complexity} represented as a tuple $(\mathcal{S}, \{\mathcal{A}^i\}_{i \in \mathcal{N}}, r, \mathcal{P}, \gamma)$. $\mathcal{N}=\{1, \cdots, n\}$ denotes a set of agents. At time step $t$ of, each agent $i$ observes the full state $s_t$ in the state space $\mathcal{S}$ of the environment and performs an action $a^i_t$ in the action space $\mathcal{A}^i$ from its policy $\pi^i(\cdot \vert s_t)$. The joint policy consists of all the individual policies $\bm{\pi}(\cdot \vert s_t)=\pi^1\times \cdots \times \pi^n$. The environment takes the joint action of all agents $\mathbf{a}_t=\{a_t^1, \cdots, a_t^n\}$, changes its state following the dynamics function $\mathcal{P}: \mathcal{S}\times \mathcal{A} \times \mathcal{S} \mapsto [0, 1]$ and generates a common reward $r: \mathcal{S}\times \mathcal{A} \mapsto \mathbb{R}$ for all the agents. $\gamma \in [0, 1)$ is a reward discount factor. The agents learn their individual policies and maximize the expected return: $\bm{\pi}^{\ast}=\argmax_{\bm{\pi}}\mathbb{E}_{s,\mathbf{a}\sim\bm{\pi}, \mathcal{P}}[\sum_{t=0}^{\infty}\gamma^{t}r(s_t, \mathbf{a}_t)]$. When agents only partially observe the state, the problem can be reformulated as \textit{decentralized partially observable Markov decision process} (Dec-POMDP)~\cite{amato2013decentralized} where agents can only access partial observation $\mathcal{O}^i$ from the full state $\mathcal{S}$.

In this work, we explore the properties of controllers learned via SARL and MARL for the same robot task. As illustrated in Fig.~\ref{fig:pg}, the centralized controller $\pi_{c}$ takes all the information and outputs all actions with one controller. By contrast, we can also design a set of decentralized controllers with different local observations for separate action dimensions. We seek to systematically analyze and compare these two approaches for control design, from perspectives of both classical control theory and reinforcement learning. 

\section{observability matters}

Single-agent and multi-agent reinforcement learning are typically investigated separately. Here, we first analytically establish their equivalence with policy gradient methods~\cite{mnih2016asynchronous} under specific assumptions including full observability, as shown in Fig~\ref{fig:pg}. (a) and (b). This explicit connection forms the basis for exploring when and why their behaviors diverge, and how the distinct strengths of MARL can be exploited. We further identify that in many weakly coupled multi-agent tasks\cite{witwicki2010influence}, local observability is sufficient to achieve near-optimal decision making.

\subsection{Equivalence Between SARL and MARL}

\begin{proposition}\label{prop:equivalence}
Under the assumptions stated below, the policy gradient of single-agent reinforcement learning is mathematically equivalent to the sum of policy gradients in multi-agent reinforcement learning.
\end{proposition}

\begin{assumption}\label{assumption:equivalence}
We assume the following conditions hold:
\begin{enumerate}
    \item All agents observe the full global state $s$;
    \item The single-agent joint policy factorizes as $\pi_{\theta}(\mathbf{a}|s) = \prod_{i=1}^{N}\pi_{\theta}^i(a^i|s)$, where each component policy is either Gaussian with diagonal covariance for continuous actions or categorical for discrete actions;
    \item Each agent $i$ in MARL uses policy $\pi_{\theta^i}^i(a^i|s)$ with the same distributional form as the corresponding component in SARL;
    \item All agents share the same reward function and use identical advantage estimation methods such as the temporal difference estimation~\cite{schulman2015high}, $A(s, \mathbf{a})=r(s, \mathbf{a})+\gamma V(s^{\prime})-V(s)$.
\end{enumerate}
\end{assumption}

\begin{proof}
The policy gradient for SARL is given by:
\begin{equation}
    \frac{\partial J}{\partial\theta}=\mathbb{E}_{(s, \mathbf{a})\sim \pi}[\nabla_{\theta}\log \pi_{\theta}(\mathbf{a}|s)A(s, \mathbf{a})],
\end{equation}
where $J=\mathbb{E}_{\pi}[\sum_{t=0}^{\infty}\gamma^t r_t]$ is the expected return.

For MARL, the policy gradient of agent $i$ is:
\begin{equation}
    \frac{\partial J}{\partial\theta^{i}}=\mathbb{E}_{(s, \mathbf{a})\sim \pi}[\nabla_{\theta^i}\log \pi_{\theta^i}^i(a^i|s)A^i(s, \mathbf{a})].
\end{equation}

From Assumption~\ref{assumption:equivalence}.4, under shared rewards and identical advantage estimation methods, we have $A^i(s, \mathbf{a}) = A(s, \mathbf{a})$ for all agents $i$ since $r_t(s, \mathbf{a})=r_t(s,a^i)$. By Assumption~\ref{assumption:equivalence}.2, the logarithmic derivative of the joint policy factorizes:
\begin{equation}
    \nabla_{\theta}\log \pi_{\theta}(\mathbf{a}|s) = \sum_{i=1}^{N}\nabla_{\theta^i} \log\pi_{\theta^i}^i(a^i|s).
\end{equation}

Substituting this factorization into the SARL policy gradient, we have:
\begin{align}
    \frac{\partial J}{\partial\theta} &= \mathbb{E}_{(s, \mathbf{a})\sim \pi}\left[\sum_{i=1}^{N}\nabla_{\theta^i} \log\pi_{\theta^i}^i(a^i|s)A(s, \mathbf{a})\right]\nonumber\\
    &= \sum_{i=1}^{N}\mathbb{E}_{(s, \mathbf{a})\sim \pi}[\nabla_{\theta^i} \log\pi_{\theta^i}^i(a^i|s)A(s, \mathbf{a})]\nonumber\\
    &= \sum_{i=1}^{N}\frac{\partial J}{\partial\theta^{i}}.
\end{align}

Therefore, the SARL policy gradient equals the sum of individual MARL policy gradients.
\end{proof}

\begin{remark}
This proposition establishes that under full observability, factorized policies, and shared rewards, SARL and MARL are mathematically equivalent in terms of their policy gradient updates. Therefore, the fundamental distinction between these two frameworks lies in their observability settings.
\end{remark}

\subsection{How Much Local Information do Individual Agents Need?}\label{sec:multi-agent}

Inspired by the concept of \textit{information state} (IS) introduced by Jayakumar et al.~\cite{subramanian2022approximate}, which establishes that a representation sufficient for predicting both rewards and state transitions is also sufficient for optimal decision-making, we posit that in certain cases, MARL with local observations can achieve performance comparable to that of fully observed settings. This is because, in many MARL tasks~\cite{witwicki2010influence}, agents are only weakly coupled: their interactions are largely localized, and the global reward often decomposes naturally into a sum of local rewards determined mainly by these local interactions.

% In this section, we empirically demonstrate that agents can make near-optimal decisions despite having limited local observations. We validate this claim on two widely used multi-agent reinforcement learning benchmark tasks: \textit{simple-spread}~\cite{lowe2017multi} and \textit{pursuit}~\cite{terry2021pettingzoo}. Our empirical findings align with the concept of the \textit{information state} (IS) introduced by Jayakumar et al.~\cite{subramanian2022approximate}, which states that if a representation is sufficient to predict both the reward and the next state, it is also sufficient for optimal decision-making. 

\begin{definition}[\textit{Information state}~\cite{subramanian2022approximate}]
    Let $\{Z_t\}_{t=1}^{T}$ be a pre-specified collection of Banach spaces. A collection $\{\sigma_t: \mathcal{H}_t \to \mathcal{Z}_t\}_{t=1}^{T}$ of history compression functions is called an \textit{information state generator} if the process $\{Z_t\}_{t=1}^{T}$, where $Z_t = \sigma_t(H_t)$, satisfies the following properties:
    
    \begin{enumerate}
        \item[(i)] sufficient for performance evaluation, i.e., for any time $t$, any realization $h_t$ of $\mathcal{H}_t$ and any choice $a_t$ of $A_t$, we have
        \begin{align*}
            &\mathbb{E}[R_t \mid H_t = h_t, A_t = a_t] \\
            &= \mathbb{E}[R_t \mid Z_t = \sigma_t(h_t), A_t = a_t].
        \end{align*}

        \item[(ii)] sufficient to predict itself, i.e., for any time $t$, any realization $h_t$ of $\mathcal{H}_t$ and any choice $a_t$ of $A_t$, we have that for any Borel subset $\mathcal{B}$ of $\mathcal{Z}_{t+1}$,
        \begin{align*}
            &\mathbb{P}(Z_{t+1} \in \mathcal{B} \mid H_t = h_t, A_t = a_t) \\
            &= \mathbb{P}(Z_{t+1} \in \mathcal{B} \mid Z_t = \sigma_t(h_t), A_t = a_t).
        \end{align*}
    \end{enumerate}
\end{definition}

Intuitively, in many multi-agent learning tasks, the global state and reward function are often decomposable into local states and local rewards~\cite{witwicki2010influence}. Consequently, local observations may be sufficient to serve as \textit{information states} for optimal decentralized decision-making.

% As shown in the weakly coupled Dec-POMDPs~\cite{witwicki2010influence}, agents are only weakly coupled, allowing the global state and reward to be linearly decomposed into local observations and local rewards. This decomposition naturally satisfies the assumptions of the information state (IS), ensuring that each agent can make optimal decisions based solely on its local information since the local observations are sufficient to predict the local rewards, as evidenced by our experiments. 

\subsection{Experiments}

\begin{figure}[ht]
    \centering
    \includegraphics[width=.7\linewidth]{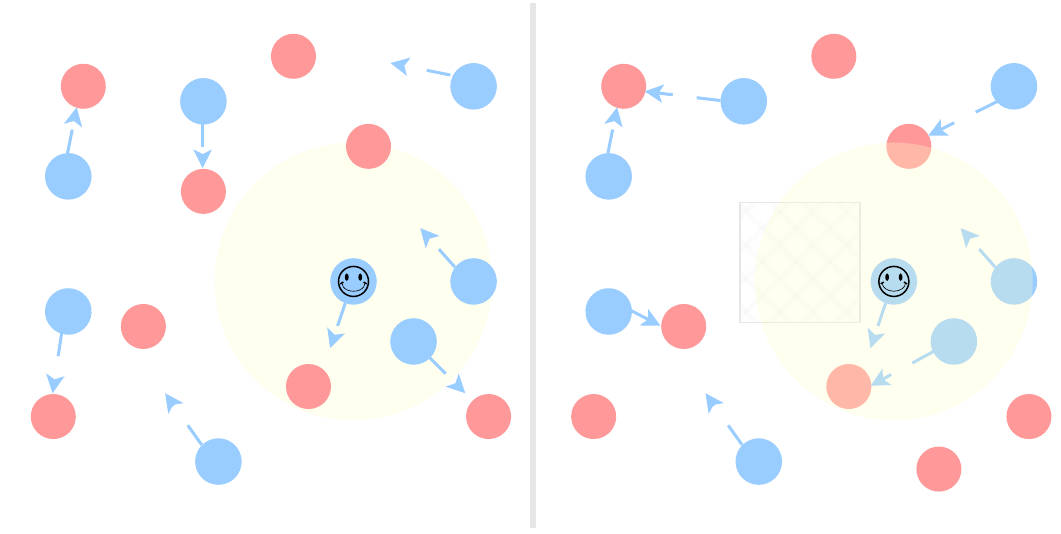}

    \vspace{0.5em} % space between image and subcaptions
    \makebox[\linewidth][c]{%
      \makebox[0.35\linewidth][c]{(a) \textit{simple-spread}}%
      % \hspace{2em}%
      \makebox[0.35\linewidth][c]{(b) \textit{pursuit}}%
    }

    \caption{Illustration of (a) \textit{simple-spread} task and (b) \textit{pursuit} task. Orange shaded regions represent the local observation range of the ego agent marked with smiley face.}
    \label{fig:task_show}
\end{figure}

\begin{figure}[ht]
    \centering
    \includegraphics[width=.48\textwidth]{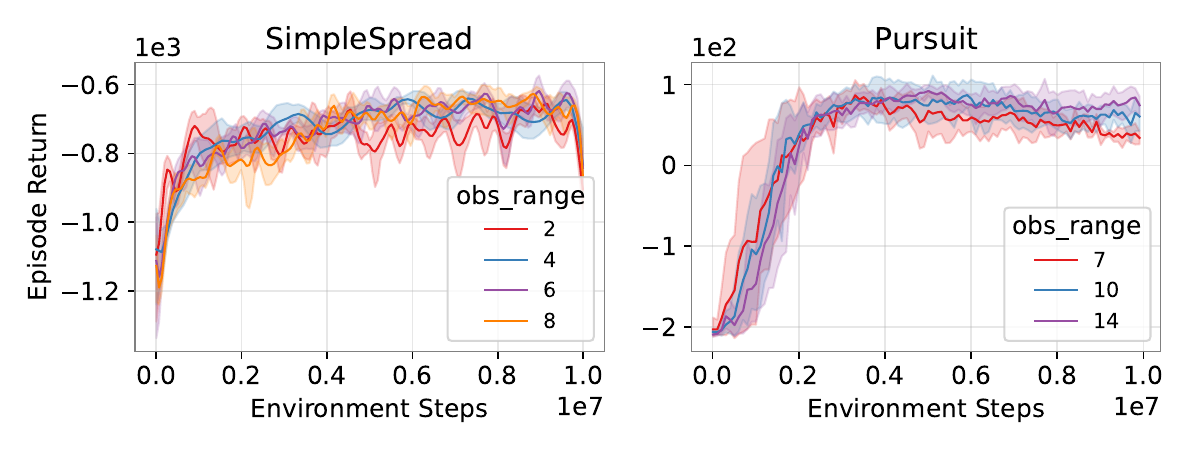}
    \caption{Evaluation results for the \textit{simple-spread} task on the left and the \textit{pursuit} task on the right. The results show that performance remains comparable across a wide range of observation configurations. Each curve represents the average return over five random seeds, and the shaded regions indicate 95\% confidence intervals.}
    \label{fig:training_marl}
    % \vskip -0.15in
\end{figure}

As illustrated in Fig.~\ref{fig:task_show}, we evaluate our approach on two benchmark tasks: \textit{simple-spread} (a)~\cite{lowe2017multi} and \textit{pursuit} (b)~\cite{terry2021pettingzoo}. In the \textit{simple-spread} task (left), each agent observes its own velocity and the relative positions of its $k$ nearest neighboring agents.We vary $k \in \{2, 4, 6, 8\}$ and analyze how different levels of local information affect decision-making performance. In the \textit{pursuit} task (right), each agent observes its surroundings within a fixed-radius circle, and we investigate different observation ranges by setting $k \in \{7, 10, 14\}$. We employ multi-agent proximal policy optimization (MAPPO)~\cite{yu2022surprising} as the policy learning algorithm. As shown in Fig.~\ref{fig:training_marl}, the learning curves reveal consistently strong performance across different levels of local observability, suggesting that even limited local information can suffice for near-optimal decision making in these cooperative multi-agent settings.

% We employ multi-agent proximal policy optimization (MAPPO)~\cite{yu2022surprising} as our policy learning algorithm. MAPPO works in a similar way as independent PPO~\cite{de2020independent} except that MAPPO learns the critic using a concatenated global state in order to stabilize the learning process. However, during execution, the agent only uses its local observations. The learning curves are shown in Fig.~\ref{fig:training_marl}. The consistently close performance across different observation ranges indicates that in these tasks, quite limited local observation is sufficient to perform near-optimal decision making. Therefore, based on this empirical evidence, we are motivated to explore the advantages equipped by partially observable MARL or decentralized control. 

% We employ multi-agent proximal policy optimization (MAPPO)~\cite{yu2022surprising} as our policy learning algorithm. The learning curves are shown in Fig.~\ref{fig:training_marl}. The consistently close performance across different observation ranges indicates that in these tasks, quite limited local observation is sufficient to perform near-optimal decision making. 

% Therefore, based on this empirical evidence, we are motivated to explore the advantages equipped by partially observable MARL or decentralized control. 

%%%%%%%%%%%%% Theory part %%%%%%%%%%%%%%%%
\section{Provable Robustness}\label{sec:theory}
Building on the insights above, we hypothesize that the local observability enabled by the flexible MARL framework can be leveraged to achieve robust robot learning. Based on the networked multi-agent system model proposed by Qu et al.~\cite{qu2022scalable}, we derive an upper bound on the performance degradation when locally observed MARL policies are perturbed. 

\subsection{Networked Multi-Agent System Model}\label{subsec:model}

We consider a network of $n$ agents that are associated with an underlying undirected graph $G = (N, E)$, where $N = \{1, \ldots, n\}$ is the set of agents and $E \subset N \times N$ is the set of edges. Each agent $i$ is associated with state $s_i \in S_i$, action $a_i \in A_i$, where $S_i$ and $A_i$ are finite sets. The global state is denoted as $s = (s_1, \ldots, s_n) \in S := S_1 \times \cdots \times S_n$ and similarly the global action $a = (a_1, \ldots, a_n) \in A := A_1 \times \cdots \times A_n$. At time $t$, given current state $s(t)$ and action $a(t)$, the next individual state $s_i(t+1)$ is independently generated and is only dependent on neighbors:
\begin{equation}
P(s(t+1) | s(t), a(t)) = \prod_{i=1}^n P(s_i(t+1) | s_{N_i}(t), a_i(t)),
\end{equation}
where notation $N_i$ means the neighborhood of $i$ (including $i$ itself) and notation $s_{N_i}$ means the states of the agents in $N_i$. In addition, for integer $\kappa \geq 0$, we use $N_i^{\kappa}$ to denote the $\kappa$-hop neighborhood of $i$; that is, the nodes whose graph distance to $i$ has length less than or equal to $\kappa$. 

Each agent is associated with a class of localized policies $\pi_i^{\theta_i}$ parameterized by $\theta_i$. The localized policy $\pi_i^{\theta_i}(a_i | s_i)$ is a distribution on the local action $a_i$ conditioned on the local state $s_i$, and each agent, conditioned on observing $s_i(t)$, takes an action $a_i(t)$ independently drawn from $\pi_i^{\theta_i}(\cdot | s_i(t))$. We use $\theta = (\theta_1, \ldots, \theta_n)$ to denote the tuple of the localized policies $\pi_i^{\theta_i}$, and also use $\pi^{\theta}(a|s) = \prod_{i=1}^n \pi_i^{\theta_i}(a_i|s_i)$ to denote the joint policy, which is a product distribution of the localized policies as each agent acts independently.

Furthermore, each agent is associated with a stage reward function $r_i(s_i, a_i)$ that depends on the local state and action, and the global stage reward is defined as: 
\begin{equation}
    r(s, a) = \frac{1}{n} \sum_{i=1}^n r_i(s_i, a_i).
\end{equation}

The objective is to find localized policy tuple $\theta$ such that the discounted global stage reward is maximized, starting from some initial state distribution $\pi_0$:

\begin{align}
\max_{\pi_\theta} J(\pi_\theta) := 
&\ \mathbb{E}_{s \sim \pi_0} \mathbb{E}_{a(t) \sim \pi^{\theta}(\cdot|s(t))} \nonumber \\
&\left[ \sum_{t=0}^{\infty} \gamma^t r(s(t), a(t))\vert s(0) = s \right].
\end{align}

Importantly, following Qu et al.~\cite{qu2022scalable}, we introduce the concept of $Q_i^{\pi_\theta}$, which is the $Q$ function for the individual reward $r_i$. $Q_i^{\pi_\theta}$ is formally defined from the the global $Q$ function $Q^{\pi_\theta}$:
\begin{align}
&Q^{\pi_\theta}(s,a)=\mathbb{E}_{s, a \sim \pi^\theta} \left[\sum_{t=0}^{\infty} \gamma^t r(s(t), a(t))\vert s(0) = s, a_0=a \right] \nonumber\\
&\quad=\frac{1}{n}\sum_{i=1}^{n}\mathbb{E}_{s, a \sim \pi^\theta} \left[\sum_{t=0}^{\infty} \gamma^t r_i(s(t), a(t))\vert s(0) = s, a_0=a \right] \nonumber\\
&\quad:=\frac{1}{n}\sum_{i=1}^{n}Q_i^{\pi_\theta}(s,a).
\label{equ:individual_Q}
\end{align}
Intuitively, $Q_i^{\pi_\theta}$ can be considered as the expected accumulated local rewards of agent $i$ over the global state-action distribution, which is determined by all the agents and the environment.

\begin{remark}
    Thanks to the locally decomposable reward function of the networked multi-agent system, the global $Q$ function $Q^{\pi_\theta}$ can be decomposed into individual local $Q$ functions $Q_i^{\pi_\theta}$.
\end{remark}

\begin{definition}[$(c, \rho)-$\textit{Exponential decay property}~\cite{qu2022scalable}]

    We say the $(c, \rho)-$exponential decay property holds if, for any agent $i\in \mathcal{N}$ with policy $\pi_{\theta}$, for any $s_{\mathcal{N}_i^{\kappa}} \in \mathcal{S}_{\mathcal{N}_i^{\kappa}}$, $s_{\mathcal{N}_{-i}^{\kappa}}$ and $\bar{s}_{\mathcal{N}_{-i}^{\kappa}} \in \mathcal{S}_{\mathcal{N}_{-i}^{\kappa}}$, $a_{\mathcal{N}_i^{\kappa}} \in \mathcal{A}_{\mathcal{N}_i^{\kappa}}$, $a_{\mathcal{N}_{-i}^{\kappa}}$ and $\bar{a}_{\mathcal{N}_{-i}^{\kappa}} \in \mathcal{A}_{\mathcal{N}_{-i}^{\kappa}}$, the Q-function $Q_i^{\pi_{\theta}}$ satisfies:

    \begin{equation}
    \begin{split}
    &|Q_i^{\pi_{\theta}}(s_{\mathcal{N}_i^{\kappa}}, s_{\mathcal{N}_{-i}^{\kappa}}, a_{\mathcal{N}_i^{\kappa}}, a_{\mathcal{N}_{-i}^{\kappa}}) \\
    &\quad - Q_i^{\pi_{\theta}}(s_{\mathcal{N}_i^{\kappa}}, \bar{s}_{\mathcal{N}_{-i}^{\kappa}}, a_{\mathcal{N}_i^{\kappa}}, \bar{a}_{\mathcal{N}_{-i}^{\kappa}})| \leq c\rho^{\kappa+1}.
    \end{split}
    \end{equation}
    
    % $|Q_i^{\pi_{\theta}}(s_{\mathcal{N}_i^{\kappa}}, s_{\mathcal{N}_{-i}^{\kappa}}, a_{\mathcal{N}_i^{\kappa}}, a_{\mathcal{N}_{-i}^{\kappa}}) - Q_i^{\pi_{\theta}}(s_{\mathcal{N}_i^{\kappa}}, \bar{s}_{\mathcal{N}_{-i}^{\kappa}}, a_{\mathcal{N}_i^{\kappa}}, \bar{a}_{\mathcal{N}_{-i}^{\kappa}})| \leq c\rho^{\kappa+1}$.
\end{definition}

Note that $c > 0$ is a constant and $\rho \in (0,1)$ is a decay factor with $\rho \leq \gamma$, where $\gamma \in (0,1)$ is the discount factor used in the reinforcement learning problem. Lemma 2(a) in~\cite{qu2022scalable} shows that the exponential decay property holds generally with $\rho=\gamma$.

\subsection{Bounded Robustness}

\begin{theorem}[\textit{Robustness via locality}]
   
    Consider the networked MDP described in Section~\ref{subsec:model} with the $(c,\rho)$-exponential decay property. Let $(s, a) = (s_{N_i^{\kappa}}, s_{N_{-i}^{\kappa}}, a_{N_i^{\kappa}}, a_{N_{-i}^{\kappa}})$ and $(s', a') = (s_{N_i^{\kappa}}, s'_{N_{-i}^{\kappa}}, a_{N_i^{\kappa}}, a'_{N_{-i}^{\kappa}})$ be two state-action profiles that differ only outside agent $i$'s $\kappa$-hop neighborhood. Under a fixed policy $\pi = (\pi_1, \ldots, \pi_n)$, if the system starts from these two different initial state-action pairs, then the performance loss is bounded:
    $$|J^\pi(s, a) - J^\pi(s', a')| \leq \frac{c\rho^{\kappa+1}}{1-\gamma},$$
    where $J^\pi(s, a)$ denotes the expected discounted reward starting from initial state-action pair $(s, a)$ under policy $\pi$.
\end{theorem}

\begin{proof}

% \textbf{Step 1: Setup and Notation}

Let $d_{t,i}$ and $d'_{t,i}$ be the distributions of $(s_i(t), a_i(t))$ conditioned on starting state-action pairs $(s(0), a(0)) = (s, a)$ and $(s(0), a(0)) = (s', a')$ respectively, under policy $\pi$.

By the local dependence structure and the localized policy structure, we have $d_{t,i} = d'_{t,i}$ for all $t \leq \kappa$. This is because:
\begin{itemize}
\item[i)] The distribution of $s_i(t)$ depends only on $(s_{N_i^{t-1}}, a_{N_i^{t-1}})$ from the previous time step;
\item[ii)] For $t \leq \kappa$, the $t$-hop neighborhoods $N_i^t$ are contained within $N_i^\kappa$;
\item[iii)] The initial states and actions are identical within $N_i^\kappa$: $(s_{N_i^\kappa}, a_{N_i^\kappa}) = (s'_{N_i^\kappa}, a'_{N_i^\kappa}).$
\end{itemize}

% \textbf{Step 2: Bound on Q-function Difference}

% Following the proof of Lemma 2(a)~\cite{qu2022scalable}, we know that the exponential decay property gives us:
% $$|Q_i^\pi(s, a) - Q_i^\pi(s', a')| \leq c\rho^{\kappa+1}$$

% Expanding the Q-function definition:
% $$Q_i^\pi(s, a) = \mathbb{E}_{a(t) \sim \pi(\cdot|s(t))}\left[\sum_{t=0}^{\infty} \gamma^t r_i(s_i(t), a_i(t)) \bigg| s(0) = s, a(0) = a\right]$$
Assuming $\forall i, r$ is upper bounded by $\bar{r}$, the difference between $Q_i^\pi(s, a)$ and $Q_i^\pi(s', a')$ arises only from time steps $t > \kappa$:
\begin{align}
&|Q_i^\pi(s, a) - Q_i^\pi(s', a')| \nonumber\\
&= \left|\sum_{t=0}^{\infty} \left[\mathbb{E}_{(s_i,a_i) \sim d_{t,i}} \gamma^tr_i(s_i, a_i) - \mathbb{E}_{(s_i,a_i) \sim d'_{t,i}} \gamma^tr_i(s_i, a_i)\right]\right|\nonumber\\
&\leq\sum_{t=0}^{\infty}\left|\left[\mathbb{E}_{(s_i,a_i) \sim d_{t,i}} \gamma^tr_i(s_i, a_i) - \mathbb{E}_{(s_i,a_i) \sim d'_{t,i}} \gamma^tr_i(s_i, a_i)\right]\right|\nonumber\\
&=\sum_{t=\kappa+1}^{\infty}\left|\left[\gamma^t\mathbb{E}_{(s_i,a_i) \sim d_{t,i}} r_i(s_i, a_i) - \gamma^t\mathbb{E}_{(s_i,a_i) \sim d'_{t,i}} r_i(s_i, a_i)\right]\right|\nonumber\\
&\leq \sum_{t=\kappa+1}^{\infty} \gamma^t \bar{r} \text{TV}(d_{t,i}, d'_{t,i}) \nonumber\\
&\leq \frac{\bar{r} \gamma^{\kappa+1}}{1-\gamma} = c\rho^{\kappa+1},
\end{align}
where $\text{TV}(d_{t,i}, d'_{t,i})$ is the total variation distance between $d_{t,i}$ and $d'_{t,i}$ upper bounded by one, i.e., $\text{TV}(d_{t,i}, d'_{t,i}) \leq 1$. We note $c = \frac{\bar{r}}{1-\gamma}$, and $\rho = \gamma$.

% \textbf{Step 3: Bound on Global Performance Loss}

% Since we can compute the global Q-function as:
% $$Q^\pi(s, a) = \frac{1}{n}\sum_{i=1}^n Q_i^\pi(s, a),$$
% the performance loss thus can be bounded by combining $Q_i$:
Therefore, we can compute the global performance drop:
\begin{align}
|J^\pi(s, a) - J^\pi(s', a')|&=|Q^\pi(s, a) - Q^\pi(s', a')|\nonumber\\
&\leq \frac{1}{n}\sum_{i=1}^n |Q_i^\pi(s, a) - Q_i^\pi(s', a')|\nonumber\\
&\leq \frac{1}{n}\sum_{i=1}^n c\rho^{\kappa+1} = c\rho^{\kappa+1}.
\end{align}

% Since $c = \frac{\bar{r}}{1-\gamma}$, we have:
% $$|J^\pi(s, a) - J^\pi(s', a')| \leq \frac{\bar{r}\rho^{\kappa+1}}{1-\gamma}.$$
This completes the proof.
\end{proof}

\begin{remark}
This result directly extends the exponential decay property of Lemma 2 in Qu et al.~\cite{qu2022scalable} to show that perturbations in both states and actions outside the $\kappa$-hop neighborhood have exponentially small effects on system performance.
\end{remark}

\section{Empirical Robustness}\label{sec:real_robot}
\begin{figure}[ht]
    \centering
    \includegraphics[width=.35\textwidth]{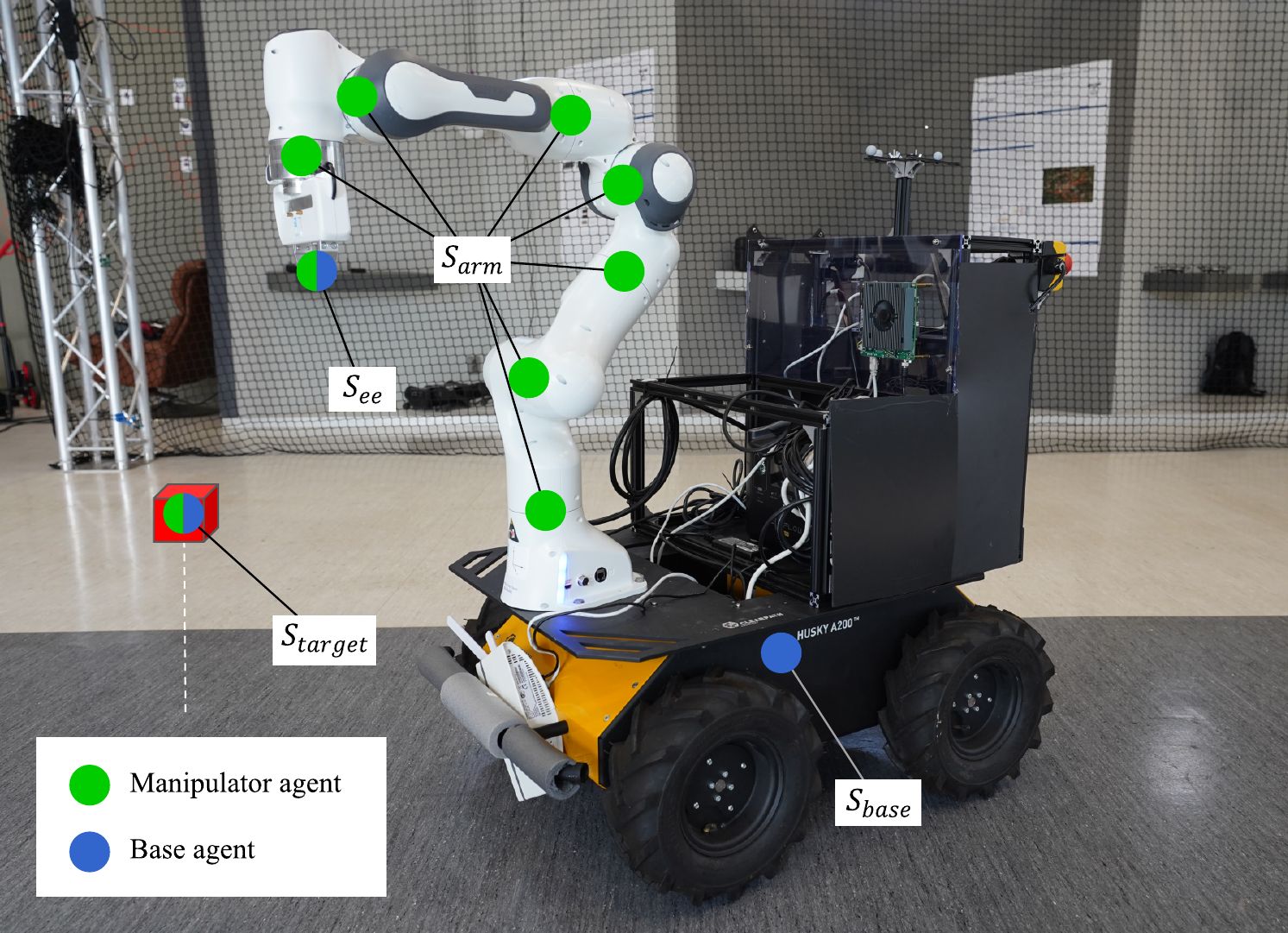}
    \caption{The mobile manipulator can be modeled either as a single agent or as two cooperative agents, corresponding to the mobile base and the manipulator. In the two-agent formulation, each agent may either share full observations of the entire system or operate using only locally available information.}
    \label{fig:real_robot}
\end{figure}

In addition to the theoretical guarantee of robustness through multi-agent reinforcement learning with local observations, we conduct a real-world robotic experiment to demonstrate the practical benefits of the proposed approach. Specifically, we consider a mobile manipulator robot, illustrated in Figure~\ref{fig:real_robot}, which comprises a Clearpath Husky mobile base and a 6-DoF Franka Panda manipulator. The task requires the robot to reach a randomly specified target point. Given the modular structure and high-dimensional action space of the system, this setup naturally raises the question of whether it should be modeled as a single agent or as two cooperating agents, i.e. the \textit{base agent} and \textit{manipulator agent}. 

\begin{figure*}[t]

    \centering
    \begin{subfigure}[t]{\textwidth}
        \centering
        \vskip 0.1in
        \setlength\figureheight{.2\textwidth}
        \setlength\figurewidth{.37\textwidth} 
        \footnotesize{\input{tex/realworld/e1p1}}
        \hfill
        \footnotesize{\input{tex/realworld/e1p2}}
        \hfill
        \footnotesize{\input{tex/realworld/e1p3}} \\[-1em]
        \caption{}
        % \footnotesize{\hspace{2.5em}(a)}
        \label{fig:exp_real_1}
    \end{subfigure}
    % \vspace{1em}
    \begin{subfigure}[t]{\textwidth}
        \centering
        \vskip 0.1in
        \setlength\figureheight{.2\textwidth}
        \setlength\figurewidth{.37\textwidth} 
        \footnotesize{\input{tex/realworld/e2p1}}
        \hfill
        \footnotesize{\input{tex/realworld/e2p2}}
        \hfill
        \footnotesize{\input{tex/realworld/e2p3}}\\[-1em]
        \caption{}
        % \footnotesize{\hspace{2.5em}(b)}
        \label{fig:exp_real_2}
    \end{subfigure}
    % \vspace{1em}
    \begin{subfigure}[t]{\textwidth}
        \centering
        \vskip 0.1in
        \setlength\figureheight{.2\textwidth}
        \setlength\figurewidth{.37\textwidth} 
        \footnotesize{\input{tex/realworld/e3p1}}
        \hfill
        \footnotesize{\input{tex/realworld/e3p2}}
        \hfill
        \footnotesize{\input{tex/realworld/e3p3}}\\[-.4em]
        \caption{}
        % \footnotesize{\hspace{2.5em}(c)}
        \label{fig:exp_real_3}
    \end{subfigure}
    
     \caption{Experimental results with a real-world mobile manipulator (a) under nominal conditions, (b) with the manipulator agent disabled, and (c) with the manipulator agent disabled, and also its initial position perturbed. Our experiments show all policies are able to complete the task under nominal conditions in (a). However, if one of the agents fails to move, as the manipulator in (b), the decentralized MARL approach that uses global observations for all agents fails. Finally, (c) shows that convergence is only accomplished with the partially observable MARL policy when a higher level of disturbance occurs. Therefore, our solution enables higher system robustness.}
    \label{fig:realrobotexperiments}
    % \vskip -0.15in

\end{figure*}

\subsection{Experimental Setup}

We evaluate the performance and robustness of three algorithmic configurations: (i) a single-agent policy with full observability, referred to as \textit{SARL}; (ii) a two-agent policy in which the base and manipulator are controlled separately, with both agents having access to global observations, denoted as \textit{MARL (global)}; and (iii) a two-agent policy where each agent operates based solely on local observations, hereafter referred to as \textit{MARL (partial)}. To assess both performance and robustness, we conduct experiments under three distinct conditions: (i) nominal operation; (ii) failure of the manipulation agent; and (iii) manipulation agent failure combined with a perturbation of its initial state. Performance is measured by recording the real-time Euclidean distance between the robot arm’s end effector and the target point. To ensure a diverse set of motion demands, we specify three target positions: $p_1 = (0.4,\ -0.6,\ 0.5)$, $p_2 = (0.3,\ 2.0,\ 0.7)$, and $p_3 = (-2.0,\ -1.0,\ 0.4)$. These locations are chosen to induce a combination of forward and backward translations, as well as rotational movements of the mobile base.

\subsection{Results}
We employ PPO~\cite{schulman2017proximal} as the RL policy and explore different observability conditions and policy configurations, as illustrated in Fig.~\ref{fig:pg}. After training in Isaac Sim~\cite{NVIDIA}, we conduct zero-shot \textit{Sim-to-Real} and \textit{Sim-to-Sim} evaluations to assess the robustness of policies learned by three methods. In \textit{Sim-to-Real}, we deploy the policies on a real mobile manipulator to test resilience against perturbations and agent failures. For \textit{Sim-to-Sim}, we perform extensive tests in the Gazebo simulator~\cite{koenig2004design}. 

\subsubsection{Real-world Experiments}

% In the real robot experiments, the trained model weights are exported and integrated into a ROS Noetic node. The base agent controller runs in a Jetson AGX Xavier onboard computer, while the manipulator agent controller runs in an Up Xtreme board with an Intel i7 processor. Ground truth and joint poses are obtained with an Optitrack Motion Capture system. Poses are only given to the base and the arm's end effector, in addition to the target. The rest of joints representing observable states are defined only by their orientation.

The real robot experiment results are summarized in Fig.~\ref{fig:realrobotexperiments}. In the first set of experiments with the nominal behavior of both agents as shown in Fig.~\ref{fig:exp_real_1}, the \textit{SARL} baseline achieves the best performance, reaching closest to the target point, while \textit{MARL (global)} shows only a slightly reduced margin. Notably, despite limited observation, \textit{MARL (partial)} attains near-optimal performance.

% In the second set of experiments shown in Fig.~\ref{fig:exp_real_2}, we disable the \textit{manipulator agent} entirely. The results show that \textit{MARL (partial)} outperforms the baselines by reaching the closest distances to the targets. Additionally, we observe that the \textit{MARL (global)} baseline diverges in all three experimental trials, whereas \textit{SARL} exhibits superior performance. We hypothesize that this disparity is due to \textit{MARL (global)} comprising two policy networks with full state dimensions, which likely requires more data for training to reach a performance level comparable to \textit{SARL}.

In the second set (Fig.~\ref{fig:exp_real_2}), the \textit{manipulator agent} is disabled. Here, \textit{MARL (partial)} outperforms both baselines by achieving the smallest distances to the targets. Meanwhile, \textit{MARL (global)} diverges in all three trials, whereas \textit{SARL} maintains relatively better stability. We hypothesize that this divergence arises because \textit{MARL (global)} involves two policy networks operating over full state dimensions, likely requiring more training data to achieve comparable robustness.

% The third experiment introduces an additional perturbation to the initial state of the manipulator, resulting in a different setup from the training process. In these experiments shown in Fig.~\ref{fig:exp_real_3}, \textit{MARL (partial)} consistently outperforms the other two baselines, with the \textit{SARL} policy not always being able to converge. This result suggests that introducing partial observability to the agents increases the robustness to perturbations to the system such as dysfunction of some agents. 

The third experiment introduces perturbations to the manipulator's initial state, deviating from the training setup (Fig.~\ref{fig:exp_real_3}). In this scenario, \textit{MARL (partial)} consistently outperforms the other baselines, with \textit{SARL} failing to converge in some trials. These results suggest that partial observability can enhance policy robustness against system perturbations and agent failures.

\begin{figure}[ht!]
    \centering
    % \vskip 0.1in
    \includegraphics[width=.4\textwidth]{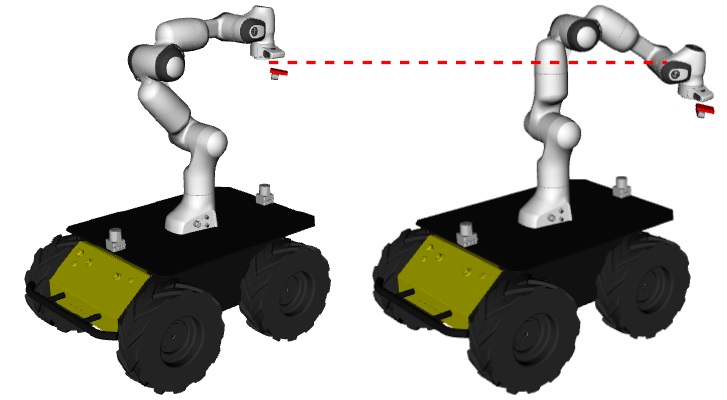}
    \caption{The left shows the mobile manipulator in the nominal position, while the right shows the robot with a perturbed initial position which has never been seen during training. The red dashed line highlights the changed position of the end-effector. }
    \label{fig:gazebo_robot}
\end{figure}

\subsubsection{Gazebo Simulation Experiments}

We further thoroughly evaluate the policies by deploying them on a simulated robot in Gazebo~\cite{koenig2004design}, as depicted in Fig.~\ref{fig:gazebo_robot}. Following the real-world experiments setup, we repeat each experiment 10 times using the Husky manipulation simulation package. In each repetition, we slightly randomize the target points, such that the target point is uniformly distributed within a $0.5 \times 0.5 \times 0.5\; \mathrm{m^3}$ box. The results are summarized in Fig.~\ref{fig:gazeboexperiments}.

In the first set of experiments with nominal agents shown in Fig.~\ref{fig:exp_gazebo_1}, all three methods achieve target points. In the second set, where the \textit{manipulator agent} is disabled, the results mirror the real-world results: \textit{MARL (global)} consistently diverge, while the \textit{SARL} policy demonstrates relative success across all three targets. The proposed \textit{MARL (partial)} shows the best performance, as shown in Fig.~\ref{fig:exp_gazebo_2}. In the third set of experiments shown in Fig.~\ref{fig:exp_gazebo_3}, with an additional perturbation of the initial state of the \textit{manipulator agent}, \textit{MARL (partial)} is the only method that consistently succeeds.

% The extensive repeated experiments in Gazebo emphasize the improved robustness of the tested mobile manipulator robot via partially observable MARL, showing that well-designed decentralized controllers with local observations, either by control or learned via MARL, can be a cheap way to achieve better robustness compared to centralized controller with full observation.

% Extensive experiments conducted in the Gazebo simulation environment underscore the enhanced robustness of the tested mobile manipulator robot when employing partially observable MARL. These empirical results are also consistent with the theoretical bound established in Section~\ref{sec:theory}. The findings demonstrate that well-designed decentralized controllers utilizing local observations can offer a cheap approach to achieving superior robustness compared to centralized controllers with full observation capabilities.

% \joni{Now, we are saying that MARL works but SARL does not but do not explain why much. Similar to the decentralized PI case, we should go deeper in the insights. Why do local observations work well here? and to what kind of applications could our approach transfer well, and, to what kind of applications not? This could be also discussed in more detail in the methods part and the MPE parts.}

\begin{figure*}[!htbp]

    \centering
    \begin{subfigure}[t]{\textwidth}
        \centering
        \vskip 0.1in
        \setlength\figureheight{.2\textwidth}
        \setlength\figurewidth{.37\textwidth} 
        \footnotesize{\input{tex/Gazebo/e1p1}}
        \hfill
        \footnotesize{\input{tex/Gazebo/e1p2}}
        \hfill
        \footnotesize{\input{tex/Gazebo/e1p3}} \\[-1em]
        \caption{}
        % \footnotesize{\hspace{2.5em}(a)}
        \label{fig:exp_gazebo_1}
    \end{subfigure}
    % \vspace{1em}
    \begin{subfigure}[t]{\textwidth}
        \centering
        \vskip 0.1in
        \setlength\figureheight{.2\textwidth}
        \setlength\figurewidth{.37\textwidth} 
        \footnotesize{\input{tex/Gazebo/e2p1}}
        \hfill
        \footnotesize{\input{tex/Gazebo/e2p2}}
        \hfill
        \footnotesize{\input{tex/Gazebo/e2p3}}\\[-1em]
        \caption{}
        % \footnotesize{\hspace{2.5em}(b)}
        \label{fig:exp_gazebo_2}
    \end{subfigure}
    % \vspace{1em}
    \begin{subfigure}[t]{\textwidth}
        \centering
        \vskip 0.1in
        \setlength\figureheight{.2\textwidth}
        \setlength\figurewidth{.37\textwidth} 
        \footnotesize{\input{tex/Gazebo/e3p1}}
        \hfill
        \footnotesize{\input{tex/Gazebo/e3p2}}
        \hfill
        \footnotesize{\input{tex/Gazebo/e3p3}}\\[-.4em]
        \caption{}
        % \footnotesize{\hspace{2.5em}(c)}
        \label{fig:exp_gazebo_3}
    \end{subfigure}
    
     \caption{\textit{Sim-to-Sim} results in Gazebo. Experiments on each target are repeated 10 times with slightly perturbed target points. The simulated robot is experimented with (a) nominal conditions, (b) the manipulator agent disabled, and (c) the manipulator agent disabled and its initial position perturbed. Under nominal conditions (a), all policies complete the task. In (b), the MARL approach with global observations fails. Only the partially observable MARL policy achieves convergence in (c).}
    \label{fig:gazeboexperiments}
    % \vskip -0.15in
    
\end{figure*}

\section{Conclusion And Limitations}
\label{sec:conclusion}

% Control policies learned with DRL are already empowering robots out of the lab, from quadrupeds to dexterous manipulation. As robots increase in complexity, a design approach is to decentralize control. 

% Multi-agent reinforcement learning has arisen as a potential paradigm for controlling not only multiple, but also individual complex robots that have been previously modeled as a single agent. In this study, we comprehensively explore the potential benefits and drawbacks of MARL and SARL control policies for robot learning. (i) We analytically examine the relationship between these two paradigms and show that the observability is the core factor to distinguish these two paradigms. (ii) We empirically show that in weakly coupled Dec-POMDP tasks, local observations are sufficient to achieve near-optimal decision making comparable with the full observability setting. (iii) We further prove the bounded performance degradation of MARL policies with local observations under perturbations. (iv) Our real-world mobile manipulator experiments confirm the improved robustness by exploiting local observations in the framework of MARL.  

Multi-agent reinforcement learning (MARL) has emerged as a promising paradigm for controlling not only multi-robot systems but also complex single robots traditionally modeled as single agents. In this study, we systematically examine the relative advantages and limitations of MARL and single-agent reinforcement learning for robotic control. In particular, we empirically and theoretically demonstrate that MARL with local observations can enhance the robustness of robot learning.
% \begin{itemize}
%     \item [i)] We analytically examine the relationship between these two paradigms and identify observability as the key factor distinguishing them.
%     \item [ii)] We empirically demonstrate that in two weakly coupled Dec-POMDP tasks, local observations are sufficient to achieve near-optimal decision-making performance comparable to that under full observability.
%     \item [iii)] We theoretically establish a bound on the performance degradation of MARL policies using local observations in the presence of perturbations.
%     \item [iv)] Finally, real-world experiments with a mobile manipulator confirm the enhanced robustness achieved by leveraging local observations within the MARL framework.
% \end{itemize}

However, our work has several limitations and offers opportunities for future improvement. First, our real-world experiments are limited in scope and could be extended to more complex tasks with critical robustness requirements, such as multiple-legged robots. Second, it would be beneficial to formally identify the specific robot systems to which our findings are applicable. Third, while we empirically demonstrate improved robustness through partial observation, there is still a need for a principled approach to designing observation spaces for decentralized controllers.

\bibliographystyle{IEEEtran}
\bibliography{bibliography}

\end{document}

%% file: tex/realworld/e2p1.tex
% This file was created with tikzplotlib v0.10.1.
\begin{tikzpicture}

% \definecolor{darkslategray38}{RGB}{38,38,38}
% \definecolor{lavender234234242}{RGB}{234,234,242}
% \definecolor{mediumseagreen85168104}{RGB}{85,168,104}
% \definecolor{peru22113282}{RGB}{221,132,82}
% \definecolor{steelblue76114176}{RGB}{76,114,176}

% \begin{axis}[
%     width=\figurewidth,
%     height=\figureheight,
%     axis background/.style={fill=lavender234234242},
%     axis line style={white},
%     tick align=outside,
%     x grid style={white},
%     major tick length=2.0,
%     xmajorgrids,
%     xmajorticks=true,
%     % xmin=-1.5005049348, xmax=33.1002847788,
%     xmin=-1, xmax=41,
%     xtick style={color=darkslategray38},
%     y grid style={white},
%     ylabel=\textcolor{darkslategray38}{Dist. to target (m)},
%     % yticklabels={},
%     % xlabel=\textcolor{darkslategray38}{Time (s)},
%     xticklabels={},
%     ymajorgrids,
%     ymajorticks=true,
%     % ymin=0.175531601110597, ymax=4.48597894290702,
%     ymin=-0.1, ymax=3.1,
%     ytick style={color=darkslategray38},
%     xtick distance=10,
%     ytick distance=1
% ]
\definecolor{darkslategray38}{RGB}{38,38,38}
\definecolor{lightgray}{RGB}{192,192,192}
\definecolor{mediumseagreen85168104}{RGB}{85,168,104}
\definecolor{peru22113282}{RGB}{221,132,82}
\definecolor{steelblue76114176}{RGB}{76,114,176}
\begin{axis}[
    width=\figurewidth,
    height=\figureheight,
    axis background/.style={fill=white},
    axis line style={color=lightgray, line width=0.5pt},
    tick align=outside,
    x grid style={color=lightgray, opacity=0.3},
    y grid style={color=lightgray, opacity=0.3},
    major tick length=2.0,
    xmajorgrids,
    xmajorticks=true,
    xmin=-1, xmax=41,
    xtick style={color=lightgray},
    xticklabels={},
    ylabel=\textcolor{darkslategray38}{Dist. to target (m)},
    ymajorgrids,
    ymajorticks=true,
    ymin=-0.1, ymax=3.1,
    ytick style={color=lightgray},
    xtick distance=10,
    ytick distance=1,
    % title=TARGET 1,
    % Scientific notation for y-axis if needed
    scaled y ticks=false,
    yticklabel style={
        /pgf/number format/fixed,
        /pgf/number format/precision=1
    },
]
\addplot [thick, steelblue76114176]
table {%
0.09250474 0.418801382077039
0.163940192 0.418572269782756
0.255251646 0.418751084249118
0.326305628 0.418464525931912
0.397476197 0.417695256561956
0.498876334 0.420202867697402
0.590162278 0.416793771485116
0.661240816 0.420210512113558
0.742403031 0.423703606666779
0.83376813 0.419571354545491
0.935195208 0.423419614655475
1.026726485 0.428061897505908
1.09782219 0.431474632449302
1.219786406 0.430012129145258
1.331327915 0.440036581500531
1.463115692 0.439564014475668
1.564584017 0.442117089233744
1.696352959 0.44666157620796
1.807969094 0.445515426922445
1.909441948 0.446674479837705
2.000802279 0.448248855890948
2.122510672 0.449650944923428
2.193617583 0.456193763126049
2.284997702 0.457680946615539
2.366241694 0.460993294478584
2.487862349 0.462833306152823
2.589267731 0.464870515835086
2.680645943 0.461785080328956
2.761850119 0.461890008025917
2.832982302 0.461945594039642
2.924291373 0.461129908479313
2.995361328 0.460199515358845
3.08678174 0.460155206708397
3.198265314 0.458863849668659
3.330027819 0.457335594819236
3.441509724 0.457085757872378
3.522857905 0.454088735476487
3.594108105 0.453100507317701
3.685631752 0.452283846450529
3.797243357 0.451089422379134
3.929163218 0.451204298545248
4.10124588 0.448898975552198
4.313900471 0.44653480985339
4.445611239 0.444414633876697
4.536956549 0.447836897873668
4.739275933 0.448050121334373
4.921463967 0.446873801721841
5.124212981 0.446505740734521
5.327070236 0.446276248777094
5.478993178 0.446576716512915
5.681494236 0.446124780281122
5.894230604 0.446812052673254
6.086878777 0.444748726073815
6.168064595 0.44539762204834
6.37054801 0.445164359092508
6.563112259 0.444997487438638
6.705000878 0.445067534756011
6.917821884 0.445034451884781
7.110368967 0.445000803020413
7.323172808 0.444128414231646
7.485586167 0.444090520948411
7.56680584 0.442735773844549
7.728897572 0.443384056136799
7.810038328 0.444088461234662
7.982586384 0.443478619936341
8.195161581 0.443962471329621
8.39761734 0.443673118673588
8.600094796 0.443820233163239
8.751982212 0.444033679250403
8.954515696 0.444638532213306
9.07633543 0.444300018966854
9.369727612 0.44378954013042
9.571942568 0.443938551966206
9.683565378 0.444130880485004
9.977035046 0.443730847613963
10.189804554 0.444199454166772
10.301288843 0.44402561399784
10.595568419 0.443947609580124
10.666796923 0.443489821498686
10.859315872 0.443958948278727
11.062064648 0.443783104378567
11.264414311 0.443429789762809
11.396034718 0.443763750935186
11.50750351 0.443540637688835
11.598783255 0.443488849225338
11.700218201 0.443302934943859
11.811644554 0.442129890154558
11.933239937 0.441315948467604
12.034729481 0.440785863733458
12.126038552 0.438869935603543
12.237662554 0.439373180097587
12.359256983 0.43884309599623
12.430381775 0.437693495106366
12.562155247 0.436715750943451
12.643394947 0.434922990781116
12.765100241 0.435181081063631
12.86661768 0.434368760678959
12.957873583 0.432558733531147
13.039026261 0.432259090958426
13.130301714 0.430598606276515
13.262112141 0.427987989502665
13.323127032 0.422808916428966
13.404210091 0.424648234512598
13.495471001 0.423657532335656
13.55640626 0.423104718646057
13.6477108 0.424893796705532
13.728859186 0.425996110498485
13.83031416 0.423343074023392
13.951911211 0.421228679133301
14.033199072 0.421592976940304
14.144652129 0.42188049755126
14.225793124 0.423203669912577
14.297086716 0.422401565472788
14.428799153 0.421741756750119
14.530296803 0.422484366692255
14.631744385 0.420400788434561
14.743179083 0.420366014730502
14.824346543 0.41942042835214
14.945932389 0.419101115137012
15.057528973 0.418495487859613
15.1487782 0.418543512485669
15.22996831 0.41695223486457
15.351686716 0.413722749529847
15.43292284 0.414300479225551
15.534305096 0.419649486563038
15.625598193 0.413471147496354
15.696880579 0.41018246540122
15.828632355 0.415827257604532
15.960434437 0.408472132954145
16.031580925 0.408369120410467
16.133126498 0.417506729197228
16.234486342 0.411529999771972
16.355988503 0.412671714580263
16.427055836 0.414986799531539
16.518376828 0.408171080939682
16.599606991 0.410427976512669
16.701022864 0.415801024166243
16.822535038 0.412439241480888
16.893602133 0.414609289689902
16.994947672 0.419227724570009
17.066084385 0.415558558632784
17.197723151 0.410924850661885
17.299069882 0.420844883233346
17.430784226 0.413954694425777
17.542302609 0.408418429975082
17.633820772 0.417307037315464
17.735202313 0.411686137305663
17.826707125 0.411787420961712
17.948357821 0.417656495176049
18.039777518 0.407761781808809
18.141279221 0.409751927068496
18.222634077 0.418075195991295
18.293773651 0.408508055641887
18.385210038 0.407410470712869
18.466770888 0.41618371216211
18.588275671 0.416966640814606
18.669589043 0.407600644226705
18.781056404 0.411658532430765
18.862235546 0.411044260576371
18.94345665 0.408016686926881
19.024600029 0.40963221950182
19.095686674 0.416078756682744
19.187045336 0.412061088389061
19.260252953 0.408004526991597
19.342592955 0.411836288468294
19.433882952 0.417332107725095
19.545318127 0.410499207125413
19.636643648 0.409736194019855
19.737942934 0.417291972364662
19.829193116 0.409989972026142
19.970947504 0.412974370625773
20.072439194 0.415964773674059
20.163925648 0.40992633316306
20.275474787 0.409156618396828
20.366814852 0.412359996201153
20.468169928 0.412468868875125
20.589821577 0.411855276731761
20.701357127 0.413192500691667
20.79285121 0.411345362404889
20.9045434 0.410797683999593
20.995945692 0.410654259453893
21.107535601 0.411056943609008
21.229075432 0.410283854494556
21.290050268 0.410066997204851
21.391376019 0.410272730918911
21.462469101 0.408980010815178
21.594170809 0.40947420855537
21.665356875 0.404165949997
21.767115355 0.406077246118478
21.86879921 0.410080544945533
21.980310202 0.402769207016561
22.071666241 0.40685437592135
22.162970305 0.409565859537892
22.274569989 0.403164446039188
22.365885496 0.407710327581077
22.477382183 0.409360758823361
22.568677903 0.404298268924052
22.670076371 0.407062844660978
22.761365414 0.409079608075279
22.822356463 0.404506509401348
22.923764468 0.406931513903936
23.025142432 0.410627090984471
23.157158613 0.404576042710597
23.22827816 0.405254910692597
23.299425841 0.412235574088348
23.390694857 0.404956921008584
23.461741448 0.402139599440085
23.553084135 0.405240588864637
23.62419653 0.410348918245124
23.705367327 0.404240660661386
23.796758175 0.402383321270671
23.908230305 0.410604588446967
23.999544859 0.404809150925387
24.11104393 0.403125614082357
24.192187548 0.409265793487072
24.263388872 0.408384086268586
24.405192137 0.403986960730299
24.506720782 0.408470791488832
24.59792924 0.406984696359551
24.759894848 0.407477797950727
24.871298313 0.412260305231912
24.982661009 0.40470536396217
25.06384635 0.407856697324558
25.18533969 0.411491496061817
25.256409645 0.403120444165839
25.388183594 0.407586398689668
25.459645748 0.408978681934837
25.541175366 0.402180437435237
25.622466564 0.405549736980757
25.693491936 0.411570670966597
25.795112849 0.406199103806833
25.866326094 0.402421278946585
25.998237133 0.40683673768962
26.110041619 0.40805275934635
26.19132185 0.405795656916537
26.262464047 0.407989862902724
26.394116879 0.409381904830212
26.475338221 0.405675635465346
26.566613913 0.411725025836101
26.668041468 0.411531859797197
26.789701462 0.401155234224197
26.870791674 0.410549534956898
26.962049961 0.409028828067134
27.023064614 0.402734647825712
27.114395619 0.405372842887451
27.195643425 0.410052580042154
27.266771555 0.405672396444327
27.368220806 0.404657814648072
27.469539881 0.407313996677083
27.590978861 0.40774888433297
27.662043095 0.406199138341841
27.753306389 0.410672011787084
27.824348927 0.411378331555209
27.90545249 0.404877044728639
28.007118225 0.403435754450479
28.108541727 0.409612559825315
28.199830056 0.405140957768659
28.31125021 0.40597452928516
28.392560005 0.411488508697891
28.463886738 0.405586342315823
28.595634938 0.404268204737243
28.666847706 0.409174696096132
28.758202792 0.407985247805005
28.829374075 0.403418117090064
28.961020232 0.409218544030646
29.032140494 0.409837334510744
29.163849831 0.403695785120881
29.275326252 0.409406851035351
29.366576433 0.408908107058128
29.478366614 0.404262939191432
29.559791327 0.409006598638383
29.67135644 0.408089431614435
29.762793541 0.405193277489384
29.874247074 0.409721923532952
29.986082554 0.406724867612569
30.067250729 0.40519878417401
30.198911429 0.40999342012796
30.310725689 0.407033887310495
30.392047406 0.411140298045112
30.473477364 0.411156035378018
30.564881802 0.405103961245553
30.666299105 0.402987857550943
30.76779604 0.408667161994701
30.870746851 0.408952098340965
};
\addplot [thick, peru22113282]
table {%
0.072258234 0.557603988273795
0.173755169 0.557663406467345
0.265066624 0.557783313659777
0.336300612 0.557651605385534
0.437692642 0.557613450221417
0.539129257 0.558499587871961
0.650568962 0.553227240517628
0.73179102 0.553415626010417
0.803028107 0.555231558643509
0.93479681 0.553972216481668
1.006043911 0.550128462388374
1.147950888 0.548497074872748
1.239265919 0.550174438736855
1.370927334 0.549753761217757
1.502593041 0.553220936726384
1.603785277 0.555100936328357
1.735441208 0.563744031553242
1.846903086 0.569307541496154
1.938157559 0.57964576251854
2.049492598 0.588853744448726
2.170958042 0.595844378462988
2.282470703 0.605423245467239
2.40391779 0.618505017027037
2.525469065 0.63243446588991
2.596440315 0.64925356680018
2.677579642 0.660010769748436
2.799129725 0.672995139825101
2.870125771 0.691212286427725
3.001636029 0.703036593171802
3.072611809 0.72656875304398
3.194098234 0.736796902211086
3.265113354 0.758783620889248
3.346412659 0.773505405129172
3.467844963 0.785685186290948
3.639775753 0.818316837766769
3.741040945 0.848020826007441
3.832239151 0.85283539830667
3.903239727 0.864148156578851
4.034794808 0.883691791322569
4.106040955 0.910335905307731
4.237790585 0.922860439935121
4.349244833 0.955175217184805
4.430305004 0.972430113741956
4.531536579 0.988225481106198
4.602577925 1.01159589410792
4.723994017 1.02140178611851
4.805092097 1.05325271951477
4.926723957 1.07840225540098
5.038221598 1.09428449492602
5.17010212 1.12046336028171
5.28159523 1.14874776640341
5.373135567 1.17066771518271
5.474510908 1.19400499061553
5.576079607 1.21433642713278
5.677525997 1.23767941926289
5.768904925 1.26162175189855
5.88034153 1.28407836859815
5.971586943 1.30938568245516
6.083054543 1.32900834640038
6.164246559 1.35555270481736
6.235399485 1.37480265411678
6.367132902 1.38943319895558
6.478605986 1.42379054352528
6.570023298 1.44931969831125
6.681585789 1.46777258197583
6.783078671 1.49412100340736
6.884455443 1.51340008714673
6.975837708 1.52781786546185
7.07742548 1.54895984569379
7.280065775 1.57551555366817
7.371351004 1.61769617966437
7.492967367 1.63824116836891
7.604493618 1.66668262712129
7.736216068 1.68663008335846
7.847661019 1.7132861368752
7.939089537 1.73557456267396
8.050518751 1.75542205369355
8.172173739 1.77743076418354
8.273626566 1.80328403814706
8.365190268 1.81868121415811
8.436280728 1.8267721162593
8.568075657 1.84360843735433
8.629021883 1.87280596772569
8.760724545 1.8861857301551
8.831844568 1.90530784443643
8.923192501 1.92049573407787
9.085355759 1.95336783152145
9.16663599 1.96387981848984
9.278161526 1.97397096808146
9.39981699 1.99569329383057
9.52138567 2.02327550810074
9.602562189 2.04487381756273
9.714013338 2.05603045244113
9.805422783 2.07675979653724
9.927042246 2.09464182001438
10.038499594 2.11855155296747
10.190481425 2.13385782535385
10.271654845 2.1643634137508
10.39353323 2.17933160937561
10.464628697 2.19768822038325
10.556029797 2.20914508140106
10.627077103 2.22893954459404
10.748689652 2.24132636189277
10.84000516 2.25872802328137
10.951560259 2.27378565027625
11.032793045 2.29400248066792
11.10404253 2.30814204573737
11.23615098 2.3198728061749
11.347682715 2.34060054292819
11.439028502 2.35937327287396
11.550813198 2.37668584648372
11.632014752 2.39761598915757
11.703116417 2.40835163118002
11.83480525 2.41947834475182
11.906043291 2.44010138451458
12.037756443 2.45725614240893
12.149302721 2.47483864566207
12.270931721 2.49257781772136
12.382733822 2.51370706749051
12.484257937 2.53136250185054
12.595997572 2.54446532060761
12.66716671 2.56228570627777
12.758622885 2.57353798735532
12.870144129 2.58959205830053
12.991750956 2.60238444335996
13.062871218 2.62176428966412
13.144103289 2.63355847818207
13.265911579 2.64557909678586
13.337089539 2.66232931361597
13.468921185 2.6733277574357
13.580370426 2.6934715100692
13.671633244 2.71059060307406
13.783187151 2.72227779833983
13.864342928 2.738288549769
13.945756912 2.75132998338277
14.03704691 2.76330566693371
14.14855218 2.77627519748368
14.229833842 2.79143820626317
14.311036825 2.80346331606886
14.432671786 2.81793967073039
14.503810168 2.83387271943715
14.605202198 2.84390085235551
14.706695557 2.85709545184564
14.798004389 2.8722042799178
14.909566641 2.8840869003848
15.021074295 2.89799244553699
15.112428665 2.91208867951402
15.223933935 2.92257657588979
15.315240383 2.94095653629627
15.436820269 2.95324742302241
15.568560601 2.97138560826017
15.669868708 2.98836770676709
15.801534653 3.00037559304714
15.872595549 3.01819275595628
16.004248381 3.02691572218048
16.105845928 3.04489085808591
16.237550974 3.0586887269633
16.349274397 3.07481098437562
16.440607786 3.09097750867696
16.552088022 3.10344867710661
16.633260012 3.117369673119
16.704523564 3.12764410285546
16.836636782 3.13670075795975
16.948155403 3.15524951184773
17.039432287 3.16929678807086
17.140834093 3.18047057558367
17.272461891 3.19322859083754
17.373909712 3.21023879873973
17.50592804 3.22299230531703
17.617417574 3.24046969300898
17.729142189 3.25225617933159
17.800246001 3.26532499650795
17.891577959 3.2741826678629
17.972866535 3.28569046409935
18.094500542 3.29511358069251
18.165624857 3.3106723092975
18.246884823 3.3188228118583
18.338172674 3.32842744185483
18.45981741 3.33874853451055
18.571370363 3.35591905674647
18.703159809 3.36903399721726
18.814754009 3.38249413178791
18.936317206 3.39524230520069
19.047653914 3.41084838993476
19.138849735 3.42451498923517
19.250213623 3.43512364294322
19.331339836 3.44897273070954
19.452833891 3.45680496049916
19.533961058 3.46972319130886
19.625276804 3.47960093730748
19.706492901 3.49126080519235
19.817982912 3.50073500984724
19.90924716 3.511290556556
20.03081131 3.52092936248893
20.102058172 3.53521407472434
20.223627568 3.54329007015884
20.304843903 3.55739012508776
20.436612129 3.5634248151987
20.558449269 3.57859254303392
20.639772892 3.59277964118315
20.741072178 3.60196272651128
20.832267285 3.61332083786047
20.903323889 3.62163744527388
21.034840346 3.62965569699661
21.105990172 3.64194403362259
21.217488528 3.65415569468078
21.309175492 3.66489667381694
21.420765639 3.67313056407682
21.512142658 3.6832999562315
21.674122095 3.69756023589334
21.826035977 3.71563757155547
21.988048792 3.72948757134312
22.109778881 3.73426418371188
22.211174727 3.73654673503748
22.342938662 3.74819951178472
22.454497338 3.76449683605773
22.545850754 3.76957890574194
22.64730525 3.78112308081097
22.738693953 3.79326236529786
22.840144634 3.80146041548836
22.971859694 3.80848446249801
23.073361397 3.82309024449423
23.205157519 3.83529973668211
23.316799402 3.84975327916137
23.438503027 3.85795675151296
23.54994154 3.87248994945703
23.671539784 3.88560685047185
23.793369293 3.8938373275253
23.874567986 3.90658144136973
24.026567221 3.91495074740863
24.138048172 3.93305203356188
24.259592295 3.94213873178676
24.330607414 3.95628647132803
24.401677609 3.96186619460413
24.533346653 3.97010786615124
24.695411682 3.98406325054803
24.766511679 3.99716773112428
24.888024569 4.00570612225905
24.96927023 4.01574208884246
25.090801478 4.02506773934343
25.172142506 4.03592957116049
25.313879729 4.0429451826524
25.415208578 4.05703462753302
25.536777258 4.06698292463306
25.648236513 4.07878005283953
25.739681959 4.08896122990468
25.851151467 4.09945301282314
25.932319641 4.11082244587522
26.003379584 4.11887420193356
26.135064602 4.12308664507574
26.246521473 4.1358364447074
26.337735415 4.14763272531977
26.449237347 4.1538548620898
26.540485859 4.16722045020254
26.652029991 4.17550965190775
26.743337155 4.18628971847146
26.844808102 4.19719418551407
26.946731329 4.20395646406911
27.048208952 4.20937804566835
27.200035334 4.2327421610728
27.291283131 4.23933380737346
27.36241889 4.2467394954878
27.443610668 4.25426926603303
27.534914494 4.26035276399311
27.646389246 4.27171317737897
27.73763752 4.28172603293964
27.869968176 4.29004951827991
};
\addplot [thick, mediumseagreen85168104]
table {%
0.092564821 0.575670944031955
0.194178343 0.575594109334304
0.295574665 0.575359161389327
0.386918545 0.575475270650159
0.498551369 0.574422549798289
0.620048046 0.572637867193991
0.741602183 0.571769879363093
0.822897911 0.568161124676006
0.924518108 0.564765774825432
1.026014567 0.561822846130298
1.127479792 0.559648547399839
1.218884945 0.551623401708137
1.330316544 0.55123031208293
1.421713352 0.546970470552516
1.533146858 0.545440821615586
1.654733896 0.549951242649441
1.766293526 0.550911899215031
1.887858629 0.548823651598091
1.999491215 0.549686631980698
2.090877771 0.545499131791324
2.20239377 0.541640272986428
2.293689728 0.538289508501286
2.395031929 0.533270121295169
2.48635149 0.529376563358086
2.557525635 0.52547626099997
2.689259052 0.521191523500538
2.800765514 0.515584157808851
2.892014742 0.512842323779471
2.993332625 0.509079803224269
3.094686031 0.505213828021415
3.196148872 0.500240397776442
3.307615996 0.496974900880283
3.408988714 0.492487789485847
3.490167379 0.489664938349364
3.601556301 0.48488487438237
3.692750692 0.482836432969093
3.79415679 0.479835429339705
3.915792942 0.475074441835751
3.986963987 0.470798131019391
4.118813992 0.46802517711371
4.230276346 0.462265487679401
4.35219574 0.45929426755401
4.423325539 0.45589989597104
4.575179339 0.453636785498071
4.666456699 0.45050126124033
4.777911663 0.447188563393049
4.869274616 0.444142308578333
4.950526238 0.443391692471348
5.021597862 0.442434246890006
5.153271437 0.439580216184801
5.254700899 0.436413902333784
5.356308937 0.434972771050753
5.457727671 0.433461464065051
5.589502573 0.429651958172525
5.701061487 0.426581815203343
5.78228879 0.425298880253485
5.853500366 0.423088371830121
5.985494375 0.421400026393222
6.096960783 0.42077922688524
6.188573361 0.419011046209413
6.310341358 0.418259954148881
6.391584396 0.414810390964429
6.513270378 0.413250670296154
6.584378719 0.411826271047021
6.655492783 0.411987428063457
6.787164688 0.411557155284547
6.888526678 0.409467272362518
7.020253897 0.40823930765077
7.121671438 0.405515530702949
7.253407478 0.403048413680163
7.35495615 0.403683960909203
7.486800671 0.403532263939353
7.598329067 0.400711219302006
7.720063686 0.400082022730883
7.831569672 0.39854315754768
7.922949076 0.399071199224406
8.034453154 0.398865816055582
8.125846148 0.396767262973425
8.23743844 0.394084286146126
8.338788271 0.393699544629342
8.440193892 0.394260436318303
8.521317005 0.3937685532307
8.642897368 0.392616253016298
8.724053145 0.390700778182665
8.845679999 0.390568728549988
8.936937809 0.390709737480265
9.038333416 0.389633675620653
9.119602919 0.388336345931045
9.190774441 0.387432661275128
9.332578898 0.387199807539444
9.434001684 0.387365098067403
9.56584239 0.386774296460845
9.65716219 0.386283574227261
9.788803101 0.384930123998565
9.900280714 0.383841070999508
10.021840572 0.384016143337273
10.133610249 0.383738740174257
10.224977493 0.383244988164546
10.346559048 0.383019718886152
10.458158255 0.384580544980327
10.58987999 0.379064208646098
10.701477766 0.382576051954506
10.792778254 0.383429242968915
10.894125939 0.380027136218912
10.985532999 0.381911879511652
11.056664705 0.383614131375929
11.178340435 0.38373422178199
11.299999952 0.381175986840305
11.421679974 0.381420861215612
11.543705225 0.381054221059903
11.625260592 0.380566091577149
11.737406969 0.380652163878652
11.819062472 0.379509492961803
11.890257597 0.378761042815308
12.021852017 0.378475810158646
12.123192072 0.378285179613288
12.254858732 0.377916554013124
12.35646224 0.376898025457674
12.488344669 0.376970429372577
12.599849463 0.377007121935558
12.721436501 0.376738300676705
12.832961559 0.376382723299192
12.924211025 0.376648112447401
13.035917521 0.37662719905158
13.127254486 0.376744115687628
13.228637934 0.376226740442459
13.360392332 0.376523960749829
13.461903572 0.376218231962934
13.563401938 0.373900972181254
13.665140391 0.380330081982646
13.75650692 0.381938376246525
13.858035565 0.375223335191787
13.989642859 0.375753741603904
14.101079941 0.377265366366743
14.192362309 0.376032062134927
14.293941975 0.37643748407958
14.385491371 0.375049605764442
14.466699362 0.376075017855296
14.588280439 0.375393345389949
14.699794769 0.378750009389082
14.780978203 0.375191817879397
14.852276325 0.376059185506912
14.94362545 0.378151406516854
15.024803162 0.376997594228559
15.146740914 0.376558396200036
15.21783042 0.376356999254587
15.33964181 0.376628862800197
15.420817614 0.376633705498394
15.532269716 0.376724399938185
15.6235466 0.377045576948381
15.734981299 0.376678777866199
15.826270819 0.376635851139734
15.937737703 0.376239441060658
16.018927813 0.376578058417182
16.130515099 0.378188973719775
16.252266407 0.375706220961659
16.323496819 0.371461025737707
16.455215454 0.376319629062096
16.556740284 0.377376322478705
16.658195496 0.376868117407631
16.759613276 0.377091501768493
16.891414642 0.377492026293058
17.002973557 0.376957838425796
17.084222317 0.376641344511253
17.155377865 0.376825082306204
17.287050486 0.376588819431751
17.439199925 0.376342395119742
17.520456076 0.376420696476671
17.682433128 0.376539559696554
17.773713827 0.377221017820625
17.8549335 0.37634172006459
17.966562986 0.376096702511367
18.058023214 0.376685863716543
18.159512758 0.37680921884578
18.250896931 0.376660640463371
18.321969032 0.376671879252446
18.453552485 0.376641429051447
18.565154791 0.376203405244478
18.686730385 0.376306172738319
18.798344374 0.378561945372813
18.919964552 0.376108747393962
19.031435251 0.372862818680954
19.153001547 0.375444449088529
19.224088192 0.376703554327337
19.386043787 0.37645246001657
19.457144976 0.376559652385133
19.599032164 0.376936025072298
19.700387716 0.37647102326262
19.79166317 0.377098384365116
19.893145323 0.376231251377879
19.994549036 0.376107752423759
20.105962753 0.376253331888152
20.19722867 0.376381698440239
20.298638106 0.376553709050988
20.389903784 0.37638747721285
20.501497984 0.376659868861378
20.592768192 0.374522148788785
20.704212189 0.3756563951544
20.785470963 0.378251529217032
20.856556416 0.375996464821436
20.988299847 0.374058437368576
21.099770784 0.375381176701557
21.191036701 0.377121551756915
21.30249691 0.377386659074106
21.393768311 0.376792686276163
21.495167494 0.376842857769557
21.5965693 0.376422373797934
21.698002577 0.376355801526655
21.819586277 0.376138509008124
21.890672922 0.376471263638173
22.022361994 0.376643224291653
22.143950939 0.376518689574306
22.225194454 0.376343200401443
22.326956034 0.376538159428852
22.41855669 0.37588110974432
22.489684343 0.379544909731309
22.621507406 0.377557196675774
22.723040581 0.373767550729552
22.854712248 0.376925731983271
22.966223002 0.377643364707319
23.057518244 0.37606061375856
23.169047117 0.376565763453589
23.290572643 0.376202215993051
23.41210103 0.376751690545123
23.493259192 0.376383989194058
23.59461999 0.376516374430034
23.686200857 0.376517147026263
23.75726676 0.378574060771894
23.889056683 0.377933249121604
24.000528813 0.375213890791971
24.122139931 0.376536922595106
24.233636379 0.377021541775484
24.36524272 0.376839982027707
24.45653677 0.376363900184046
24.588235617 0.376393949519901
24.699743748 0.376307316766684
24.791026831 0.37638352125203
24.902656794 0.376547761035117
25.02423811 0.376511622160936
25.135752201 0.376597927610941
25.257525921 0.376010828364487
25.389270544 0.374531920050993
25.490693569 0.377795425536593
25.622441054 0.373576123550422
25.723820686 0.375627739322689
25.855460644 0.376833624952887
25.95685029 0.376989293040603
26.088561297 0.37675898055913
26.189924955 0.377068995397438
26.321593523 0.37689393954505
26.443132401 0.376398846164457
26.524266005 0.376335340204199
26.635763884 0.376060704465371
26.716963053 0.375892135276737
26.78807497 0.376426397000532
26.919715166 0.37631294393493
27.021012306 0.374769152662755
27.152690172 0.374248767727292
27.223736525 0.377938087179645
27.355404377 0.377053578142222
27.467080355 0.375742799492255
27.558751106 0.377350950692026
27.660526514 0.377369563320251
27.752294541 0.377008063916648
27.823411703 0.375581420266214
27.955066443 0.376962752643923
28.066532612 0.376502667193432
28.157921314 0.376117324035172
28.269488812 0.376090709121613
28.411197186 0.376245081685713
28.5026927 0.376426871913217
28.593952894 0.376590938927744
28.69532609 0.376581099856749
28.796685696 0.376393621868342
28.898109675 0.379146712345432
29.01974082 0.377380258164247
29.090870619 0.374044150673822
29.222578049 0.375286818536548
29.334048748 0.377658356975852
29.455757856 0.376638191488895
29.577517748 0.376963510687105
29.689012766 0.377125395709068
29.810520172 0.376348764318871
29.891746283 0.376018664397908
30.003188372 0.375999098647568
30.124840021 0.376276698148579
30.236405134 0.37612025977636
30.31764245 0.374111081183607
30.449362517 0.376963319951021
30.520485163 0.376073447731128
30.642086983 0.376387335065197
30.723336458 0.377152462610867
30.844937563 0.376851668659186
30.916074753 0.376439327555904
30.997416258 0.376514832566443
31.118989468 0.376187842748037
31.230484724 0.375971274061552
31.32178545 0.375778937996726
31.443467379 0.376641546493179
31.52752161 0.374240349768909
};
\end{axis}

\end{tikzpicture}

%% file: tex/realworld/e2p2.tex
% This file was created with tikzplotlib v0.10.1.
\begin{tikzpicture}

% \definecolor{darkslategray38}{RGB}{38,38,38}
% \definecolor{lavender234234242}{RGB}{234,234,242}
% \definecolor{mediumseagreen85168104}{RGB}{85,168,104}
% \definecolor{peru22113282}{RGB}{221,132,82}
% \definecolor{steelblue76114176}{RGB}{76,114,176}

% \begin{axis}[
%     width=\figurewidth,
%     height=\figureheight,
% axis background/.style={fill=lavender234234242},
% axis line style={white},
% tick align=outside,
% x grid style={white},
% major tick length=2.0,
% xmajorgrids,
% xmajorticks=true,
% % xmin=-1.7600806002, xmax=38.7706077342,
% xmin=-1, xmax=41,
% xtick style={color=darkslategray38},
% y grid style={white},
% % ylabel=\textcolor{darkslategray38}{Dist. to target (m)},
% % xlabel=\textcolor{darkslategray38}{Time (s)},
% xticklabels={},
% yticklabels={},
% ymajorgrids,
% ymajorticks=true,
% % ymin=0.0432696279417267, ymax=2.82365774271066,
% ymin=-0.1, ymax=3.1,
% ytick style={color=darkslategray38},
% xtick distance=10,
% ytick distance=1
% ]
\definecolor{darkslategray38}{RGB}{38,38,38}
\definecolor{lightgray}{RGB}{192,192,192}
\definecolor{mediumseagreen85168104}{RGB}{85,168,104}
\definecolor{peru22113282}{RGB}{221,132,82}
\definecolor{steelblue76114176}{RGB}{76,114,176}
\begin{axis}[
    width=\figurewidth,
    height=\figureheight,
    axis background/.style={fill=white},
    axis line style={color=lightgray, line width=0.5pt},
    tick align=outside,
    x grid style={color=lightgray, opacity=0.3},
    y grid style={color=lightgray, opacity=0.3},
    major tick length=2.0,
    xmajorgrids,
    xmajorticks=true,
    xmin=-1, xmax=41,
    xtick style={color=lightgray},
    xticklabels={},
    % ylabel=\textcolor{darkslategray38}{Dist. to target (m)},
    ymajorgrids,
    ymajorticks=true,
    ymin=-0.1, ymax=3.1,
    ytick style={color=lightgray},
    xtick distance=10,
    ytick distance=1,
    % title=TARGET 1,
    % Scientific notation for y-axis if needed
    scaled y ticks=false,
    yticklabel style={
        /pgf/number format/fixed,
        /pgf/number format/precision=1
    },
]
\addplot [thick, steelblue76114176]
table {%
0.082223415 2.69689590107679
0.183786869 2.69688353419114
0.285087347 2.696879132542
0.376327992 2.69681330578329
0.447347879 2.69370718402875
0.579047203 2.6881164423092
0.680410385 2.6747954614701
0.812146425 2.66393354788614
0.883323908 2.64079774607745
1.004791021 2.62803401770165
1.086271048 2.59978990811072
1.197816372 2.57865838803395
1.279037476 2.55118076710148
1.400582075 2.53333530429657
1.481777668 2.50659300531733
1.593265534 2.48054709894578
1.714932203 2.46148842092178
1.826691389 2.43337167052204
1.948652268 2.40334320599242
2.060200453 2.37697435744683
2.181749821 2.34686615166029
2.303303003 2.32362525434745
2.384475708 2.29754311079502
2.516155958 2.27932469770954
2.617538691 2.24677691878009
2.74922204 2.22381052381847
3.103224516 2.13307054674599
3.184374332 2.10817971252407
3.305947781 2.09279432699586
3.41754961 2.06009960101001
3.559262276 2.03394861287113
3.650552273 2.00418238717467
3.782424688 1.98591318841755
3.894400597 1.95162589013284
3.985747099 1.92614511645932
4.087193966 1.90272436319446
4.21891737 1.8789744493948
4.330432415 1.84624602138297
4.421705961 1.81986311133621
4.543319464 1.80058117969558
4.644647837 1.76907003145424
4.736134052 1.74768673914775
4.847688913 1.72313266022532
4.979336023 1.69931993707286
5.050479889 1.66976383040818
5.182234526 1.65112848436599
5.293874264 1.62227854624502
5.44605875 1.59806069326434
5.577946425 1.55404082499703
5.719787598 1.53039960209164
5.831166744 1.50033759529751
5.912328959 1.47199861117327
5.983383417 1.45547507444193
6.115066767 1.43907538438503
6.226500034 1.40733305730784
6.348191261 1.38004008712142
6.459816694 1.35517452804193
6.561524153 1.32977742648075
6.673135281 1.30701140298787
6.744394302 1.2809737206541
6.855945587 1.2625369888471
6.977740526 1.23692336198228
7.089214802 1.20891077311104
7.180468798 1.18526489323491
7.302038193 1.16248941662076
7.383237839 1.12915305411495
7.494748831 1.11392154700953
7.61648798 1.0799956040556
7.748222351 1.05714109576839
7.85973978 1.02740486184148
7.981276274 1.00214930158228
8.092742205 0.979310371842477
8.184067011 0.953495167391477
8.295540094 0.933518869014861
8.386898756 0.907238642229819
8.508511543 0.886495763678578
8.620019913 0.859145960084448
8.751642704 0.833361829184219
8.863061905 0.805316111391573
8.954324007 0.780578709285443
9.055764914 0.759108143006716
9.18768692 0.737774309483688
9.299206734 0.708942053363267
9.390651703 0.686751971093186
9.492035627 0.668792491150236
9.583388805 0.645584701740313
9.694900751 0.631919326018171
9.786204577 0.610290565847108
9.897759438 0.592191590178638
9.989080191 0.567624324602371
10.090480328 0.552179800375136
10.212040186 0.533023939027845
10.283161163 0.51223699369877
10.414791584 0.498858034315379
10.516270399 0.476566437100749
10.617805958 0.455029143411625
10.729396105 0.441303956596127
10.851060867 0.431267746760746
10.96256876 0.412754735362131
11.084140301 0.397175845364617
11.195619822 0.389671241451844
11.317220926 0.378360350914252
11.428764343 0.372207271757124
11.550480127 0.365623916127296
11.661906242 0.362425411146857
11.753402948 0.356844866352106
11.864855766 0.35513893600397
11.946262121 0.345906716410015
12.057875872 0.355711095435131
12.149328709 0.340132869493239
12.260849238 0.347755807500029
12.352105379 0.336674392946558
12.463552713 0.349381760296405
12.564914942 0.349145032639303
12.666295052 0.343831629957763
12.747470379 0.339710954442077
12.8588202 0.347490363074024
12.950038195 0.339852588517591
13.061478138 0.334673145104249
13.152873039 0.329604354546649
13.264268875 0.328104769714821
13.345431089 0.319881296559911
13.456947565 0.318461596148968
13.558377981 0.309755892171521
13.659752369 0.306248800135372
13.751016378 0.30237092819621
13.862478018 0.298684464193259
13.984072685 0.292690477772934
14.095515251 0.284792190351791
14.186850548 0.279642824068801
14.298342943 0.277132573171367
14.379528523 0.270711715351817
14.450651407 0.267140763148807
14.582433701 0.26545188085701
14.69404912 0.259771559559505
14.8155725 0.253577534264594
14.92718029 0.248762747283114
15.018568277 0.245562597440179
15.241349459 0.236868162385731
15.312594891 0.233336862373844
15.40408659 0.224551137497145
15.485321999 0.21820353308911
15.607053995 0.214953852294866
15.678239346 0.218007673426829
15.799948216 0.213998483935223
15.881174088 0.209206071585103
16.003029108 0.20935688078345
16.084397554 0.206871922246962
16.206128836 0.20591477631887
16.337826967 0.202461647409821
16.479827643 0.202413406411052
16.550980091 0.199809804737747
16.68268919 0.197222467221886
16.794243336 0.196157909353108
16.916094542 0.194933876430275
17.027876616 0.193733415010481
17.14969039 0.191604739867632
17.261392593 0.189812454180539
17.383324623 0.190914659884708
17.495075226 0.181375041483922
17.596743107 0.178774320103407
17.688114643 0.18441004117929
17.819997788 0.180740524250508
17.931483746 0.17807468805062
18.012720585 0.183818431516506
18.083861589 0.183524561712641
18.215610266 0.182984815007422
18.317282677 0.183293478334064
18.448996782 0.184194290738931
18.59092164 0.185463333352287
18.68219018 0.18640170544652
18.813802242 0.187241080576042
18.915150404 0.188445511489435
19.046854496 0.187193751449748
19.148308992 0.183484800375405
19.280081749 0.189567518862844
19.381414414 0.183657076217535
19.513078451 0.181677055625263
19.59426713 0.189096432503476
19.715879917 0.187218555020993
19.827655554 0.186842423905525
19.919171333 0.193518160968779
20.02064085 0.193047714276698
20.152660847 0.192870749198286
20.264314413 0.196161334433958
20.345642328 0.197674718708884
20.457187176 0.199932012777217
20.548550606 0.20144901423856
20.659991741 0.202966731367024
20.75124526 0.203892217287126
20.862673283 0.205195718890923
20.984250307 0.207519195828402
21.095706225 0.209574891992327
21.186949253 0.212052657801833
21.298426867 0.214057649840667
21.37966609 0.212985690757738
21.491469145 0.210345989196545
21.593162775 0.208925119171422
21.694749832 0.21360307066963
21.796470165 0.212184409263086
21.988797188 0.214745484548515
22.090531588 0.217172659264465
22.181995869 0.218023829574096
22.303709507 0.21925670465779
22.385026217 0.222490372112433
22.506659746 0.223363486655905
22.699331522 0.224952570733148
22.780486584 0.229441325528637
22.912159681 0.230760905616681
22.983242273 0.234275141889982
23.114945412 0.235955223002474
23.226600647 0.239373300530308
23.317942619 0.23962713529768
23.429615974 0.237639776682658
23.520862341 0.234419432394055
23.642446995 0.237921307124045
23.723654747 0.238999964279151
23.845209837 0.239246544473786
23.956985235 0.241648808948782
24.078553438 0.242761906076748
24.190212488 0.244687411838409
24.291611195 0.246565945403344
24.392994642 0.248951999082944
24.484220266 0.250047381446452
24.595672846 0.25081933119327
24.68687582 0.252845002202352
24.798377991 0.254152421389154
24.87970829 0.256264497442192
24.950931311 0.257056109692724
25.082636118 0.257893117941973
25.194181204 0.259424112821359
25.285489321 0.257533591566706
25.396986723 0.256457934590981
25.478259564 0.259040511678612
25.600081682 0.259746081933195
25.6812675 0.260460820840516
25.802831411 0.262190298129959
25.884004116 0.263614430014741
25.995456457 0.264755745697113
26.086773634 0.2660947914559
26.198276281 0.268295304511575
26.289937973 0.269455378557335
26.391520262 0.270494033340549
26.483068705 0.272890562617469
26.594923496 0.274006467610547
26.676625013 0.276058706558413
26.798494577 0.277804096635272
26.920051813 0.280488683398865
27.031464815 0.283226654423064
27.153078318 0.28053525030126
27.254356861 0.277928975679379
27.345569134 0.280059381806559
27.416612148 0.278849492043181
27.548298359 0.27755687103591
27.649787664 0.279134941719263
27.781311512 0.279763560412988
27.903026104 0.282218411143844
27.98415947 0.283126873079642
28.095649719 0.283408357263002
28.186911345 0.285169470032873
28.298391581 0.286582308572137
28.379961491 0.287484303116544
28.501526117 0.288179988095074
28.582654715 0.290736451007085
28.694244862 0.291625778620113
28.785480261 0.29226552547828
28.896926403 0.291103514090755
28.978127241 0.28987452832994
29.089790821 0.29057681421663
29.18107152 0.292508971002338
29.292482615 0.29057282345661
29.383766174 0.288140306603996
29.48514843 0.288624336578528
29.626946211 0.289758988299814
29.728435278 0.289305074297183
29.84998703 0.2894177124589
29.971569061 0.287290000370807
30.052874804 0.289906510642469
30.174476862 0.289171764096101
30.245558024 0.288984579451101
30.326853275 0.290892567273216
30.448604584 0.29159790849657
30.560131311 0.290316596010115
30.661599636 0.290686111608575
30.763025284 0.291839545443204
30.884669304 0.293235401340525
30.996339083 0.289027027260371
31.118332863 0.29173916522349
31.230143309 0.29191143282531
31.351833344 0.288991718145858
31.463476181 0.289944821330693
31.554779053 0.28934250230585
31.67664671 0.288008472512264
31.747728586 0.289892337683669
31.859146595 0.289108078326457
31.980869293 0.288575754276205
32.102588654 0.288880664949808
32.183765411 0.287891723724437
32.295239687 0.289341924439827
32.386699438 0.290195071387245
32.4983356 0.287882522894929
32.579893827 0.289622576746194
32.691371441 0.288364752748066
32.782616139 0.288807873595592
32.904142618 0.29139985056857
33.015577555 0.291718362440021
33.147377968 0.29185943669147
33.248653174 0.291064600083971
33.38044405 0.292860108804649
33.492022514 0.294881056408146
33.583317757 0.291927838096693
33.714949131 0.29256866196121
33.826587677 0.291569995452844
33.948218823 0.28924415604554
34.062657118 0.293199770693703
};
\addplot [thick, peru22113282]
table {%
0.10250473 2.69727042799842
0.204057694 2.69727646476662
0.305381775 2.69726946402741
0.406722069 2.69696255862555
0.518092871 2.69142057515887
0.639760733 2.68289279441882
0.751154661 2.6681717572275
0.872625113 2.65581208155515
0.984027386 2.63527425196656
1.075459004 2.60649031937107
1.186970949 2.58733480489249
1.278200388 2.55986805641152
1.440206289 2.53779811751668
1.592057705 2.49436840763614
1.673197508 2.46605962527319
1.794741392 2.44692299904842
1.86580205 2.41713986513083
1.987557173 2.3979580110274
2.068836451 2.37076101500577
2.180288315 2.35113151658023
2.281718731 2.32730758444538
2.383163452 2.30374091922781
2.474473238 2.27884138794211
2.586036205 2.25949515049331
2.82890296 2.19723188375533
2.940451861 2.16901355197239
3.122537375 2.14339500615513
3.223904848 2.10179117956928
3.305132151 2.07926236015044
3.426835299 2.05857213662492
3.53831029 2.02772745744081
3.670249701 2.00118697904077
3.7818923 1.97430989616438
3.873194218 1.95042449594298
3.994739533 1.93097243383972
4.065919399 1.90181301170715
4.187427044 1.88388562906425
4.268607378 1.85523813324971
4.380105019 1.8371242074
4.612813473 1.81159650646862
4.704045534 1.76040994297282
4.825567722 1.73877499840393
4.906698704 1.71119500907262
5.018143892 1.68950267558018
5.109641314 1.66430763859733
5.221127272 1.64407749171743
5.302273035 1.62074219210567
5.413691282 1.60140253132642
5.5049088 1.57538621220552
5.616348505 1.55600247481777
5.707580567 1.52777577362561
5.819170952 1.50749466445254
5.910480738 1.48457487949454
6.032087088 1.46411218879029
6.10315752 1.43790260412302
6.224682093 1.42315554606663
6.305856228 1.39689921893221
6.427443266 1.37752206569643
6.539130449 1.35200715737941
6.670835734 1.32892875211645
6.782276631 1.30117122193877
6.873566866 1.27594342886845
6.985040188 1.2581223986999
7.076369047 1.23454199947459
7.181915045 1.21487909034303
7.304587841 1.19604571598144
7.426424503 1.17168058102078
7.53792429 1.14692488510434
7.669788599 1.12852459971287
7.781246901 1.10332739112677
7.872532368 1.0849767879111
7.983996868 1.06658975066459
8.075325251 1.05011562222772
8.176765442 1.03409198880114
8.298379183 1.01847629545713
8.420222044 0.998567873219029
8.511490107 0.980963816686298
8.63309288 0.96897404196114
8.704202891 0.951371105705722
8.826026201 0.944586115390845
8.937617064 0.925673937445918
9.069374561 0.912180362435757
9.180906058 0.897944611804735
9.272211313 0.887191078604724
9.383721113 0.87791860119676
9.505322218 0.868306752725089
9.616811991 0.857296810532665
9.718225479 0.847656766874385
9.860129357 0.83828244506762
9.941401959 0.832508305861221
10.063139677 0.828052023782197
10.134244681 0.823580431300397
10.215458155 0.82026037104318
10.347206593 0.820075551764742
10.448829651 0.815832540798314
10.540234089 0.815440653427903
10.661832571 0.817157972477134
10.753155947 0.814894142415456
10.854552984 0.81677719148776
10.935777664 0.821122480801746
11.047302246 0.822931406821618
11.138754606 0.825927990062941
11.2506001 0.830774940608048
11.342002392 0.837554941453476
11.46365881 0.844863948060613
11.534734011 0.85179817256993
11.666519404 0.857728640107139
11.77806592 0.871681754671756
11.909797907 0.879614978520608
12.021315098 0.893090675183586
12.112642527 0.906666649130006
12.214062214 0.916293422539102
12.335681677 0.928946843091096
12.447268725 0.943031842041764
12.538800717 0.95369425247909
12.660584927 0.964681872048891
12.772193432 0.980627713834558
12.893931151 0.992047763926853
12.965044975 1.0070586958799
13.036481381 1.01556451068357
13.168362856 1.02511963657134
13.279969692 1.0412875826204
13.371269226 1.05457823607603
13.482736588 1.06767779322366
13.573967934 1.08338490982184
13.695484638 1.09569292035345
13.766555071 1.11282167613544
13.898290157 1.12318700698482
13.969623804 1.1388644653501
14.101350784 1.14937636014312
14.202769756 1.16696153642139
14.334590197 1.18056482798222
14.446467638 1.19780027968256
14.568001747 1.21475071427661
14.689839125 1.231023713826
14.771083832 1.2468036094383
14.882827521 1.2575354882424
14.97415328 1.27323535595954
15.095739842 1.28466964145334
15.2072227 1.30249385772826
15.338968516 1.31986776744194
15.450589657 1.33722722093682
15.542018652 1.34946721619037
15.663644314 1.36141587176772
15.764975071 1.38117407707743
15.906793118 1.39194777732487
16.02840066 1.41151445633716
16.139939547 1.42758238554008
16.271684885 1.44170835440722
16.383324385 1.45761468823388
16.474657297 1.47199691475618
16.586255551 1.48396315291984
16.707865477 1.50292654364255
16.839613438 1.51444171839024
16.951100588 1.53144468296723
17.072714567 1.54393014410837
17.184216261 1.55930810776052
17.275574207 1.57233739894668
17.387167692 1.58545441945071
17.498856068 1.60101219260008
17.630765677 1.61252843110764
17.742302895 1.6287759415188
17.87403965 1.64135021804963
17.985719204 1.65585570960307
18.067018747 1.66936267331497
18.178547859 1.6786870677125
18.269922018 1.69153250593332
18.381507635 1.70308661689621
18.473069668 1.71321052912229
18.584692955 1.72292347746154
18.686155796 1.73614464037505
18.787518025 1.74763078651656
18.868866444 1.75900875343292
18.980458498 1.76802740059964
19.071743488 1.77828761052264
19.183320046 1.78763490000037
19.274664641 1.7987193973868
19.386237383 1.80927535245734
19.467463493 1.82197159958608
19.579159737 1.8292180902627
19.680501223 1.83958597479404
19.781876326 1.84867329824233
19.873296738 1.85850537421726
19.98487258 1.86765814018054
20.066123963 1.87896857383976
20.177581787 1.8875711962604
20.268937826 1.89778655967925
20.380483866 1.90674761524679
20.471811772 1.91781979347854
20.583317518 1.92560823534775
20.674683809 1.92492548752047
20.786224365 1.92206289829715
20.907831431 1.92531033332272
21.029516459 1.93641719449058
21.140957356 1.94619151271985
21.272746563 1.95551536259798
21.384280205 1.96651877563936
21.485771179 1.97716827579196
21.587170124 1.98665527390307
21.678492069 1.99487787051885
21.779997826 2.00187343943475
21.871433497 2.0082269961565
21.983003616 2.01526116561284
22.074374199 2.02619590828949
22.18606472 2.03689915148483
22.267362356 2.04183844399074
22.378936291 2.04911881265467
22.470400572 2.0601074825089
22.581907511 2.06683034305102
22.693373919 2.07553345055008
22.784618855 2.08457281665561
22.90627718 2.09477885740413
23.017765284 2.09963818183214
23.099143028 2.10693177188544
23.210919619 2.11363833080125
23.332480908 2.12236335084045
23.403753996 2.13195626731313
23.535502911 2.13774452987593
23.657229901 2.15269717532757
23.738478899 2.15644313417355
23.84996438 2.16162224386574
23.971522093 2.16870636631152
24.083062172 2.17666000362183
24.174417257 2.18178987150453
24.275983095 2.19026353133772
24.407778502 2.19294516570596
24.519418001 2.20507345814876
24.620852709 2.20769676560516
24.722284794 2.21750509050093
24.803621292 2.21995368055194
24.915139914 2.22898345765097
25.036952734 2.23570980580946
25.148556948 2.24528292097411
25.23990345 2.25359980180863
25.351410389 2.25832419069454
25.442979336 2.25986303731147
25.554503441 2.26427225447546
25.635697365 2.27037340828967
25.747430325 2.2781206029717
25.849032402 2.28063935177952
25.95061326 2.285149870077
26.193421364 2.29972428854204
26.274731875 2.30945827363233
26.40660286 2.31333940164673
26.518163681 2.31804527190037
26.599469662 2.32421716299904
26.731154919 2.32716396169789
26.802242279 2.33195800420414
26.923867703 2.33806042721418
27.005087376 2.34097879416936
27.207643509 2.35064940091035
27.319206476 2.35861012340382
27.410559893 2.36381060516207
27.512186766 2.36551773921799
27.633847237 2.37676177716898
27.795973063 2.37810091144801
27.907682419 2.37859475815729
28.039463043 2.38620490949583
28.150976181 2.39590538525479
28.272561789 2.39914631061813
28.384041548 2.40351811324948
28.475398541 2.40596666199447
28.587109804 2.41138203705607
28.688827753 2.41664872915286
28.780253172 2.41858902853291
28.91190052 2.43453457285375
29.023342133 2.43257569864944
29.104531765 2.43450590352636
29.2160182 2.44410201841385
29.337733984 2.44117071136346
29.449287653 2.44513021069908
29.540667772 2.45685424547375
29.65217042 2.45293678936407
29.773980618 2.46025297003663
29.895656586 2.46117069175045
30.239592791 2.46476272754834
30.34103775 2.47240497293775
30.493075371 2.47825708217117
30.574287176 2.47899899700071
30.716126204 2.48295089688172
30.807476759 2.49038181390161
30.939353705 2.48849382107316
31.040915966 2.4914697230776
31.142350197 2.49373720803656
31.243845701 2.48539699172627
31.365475893 2.500828946455
31.497258186 2.50987766208369
31.598890305 2.51213109170678
31.740715981 2.51171211003248
31.84202981 2.52921471303904
31.97371316 2.52813575124764
32.095202685 2.5309700444465
32.206542015 2.53848792321612
32.328295231 2.53323988619766
32.450027704 2.53350161024173
32.571498871 2.53875412962765
32.713250876 2.53870413561225
32.915480375 2.54106179445452
33.006723404 2.55550423100956
33.168758631 2.55664091765404
33.330765486 2.55259707277389
33.482553244 2.54871833505647
33.695100069 2.54943575807598
33.80659318 2.55177331948396
34.019050837 2.55505715949369
34.201137066 2.55742710706948
};
\addplot [thick, mediumseagreen85168104]
table {%
0.092749357 2.69532584466161
0.194571495 2.69536407978516
0.295927286 2.69535342319832
0.397295237 2.69503066640363
0.498687268 2.69155820026554
0.630420208 2.68280955187144
0.731807232 2.67082446698213
0.873546601 2.65774056813526
1.015236616 2.62913162193392
1.096526623 2.60397096482484
1.208120108 2.58205695622821
1.319832802 2.5535348347617
1.441441775 2.52771417687071
1.532889843 2.50114730320973
1.634280205 2.48021999465762
1.756043911 2.45669302490502
1.877689839 2.43075414836158
1.999400616 2.40004708814408
2.110909939 2.37547937885298
2.222436667 2.34555337627448
2.344038487 2.32043425576651
2.4556036 2.29169269919292
2.577157021 2.26846242454615
2.658332348 2.24091551547226
2.779933214 2.22290105336252
2.901623488 2.19263098160874
3.01310587 2.16518947940389
3.09430337 2.14301276879339
3.195670843 2.12390869101885
3.327274323 2.09979239650224
3.438867331 2.06767388086515
3.560672045 2.04171964189925
3.672322274 2.01521373671091
3.794044972 1.98855386315139
3.9056108 1.96227503339527
3.98681283 1.93520191099972
4.098224163 1.9160299585857
4.189531327 1.89083351037992
4.30099988 1.87099320240177
4.392407656 1.84672606671703
4.504116774 1.8236805061695
4.595497132 1.8001493051288
4.696820021 1.7788611952115
4.828576565 1.7532353946456
4.940070153 1.72249681257046
5.051758528 1.69366730592729
5.173372984 1.66703157243003
5.254634381 1.64267131349054
5.366331578 1.62205401804575
5.457675696 1.59756679654842
5.569262743 1.57673967321887
5.660528898 1.54791442985261
5.772022963 1.52828363675284
5.853181124 1.50510125105947
6.015392304 1.47826971422043
6.086459399 1.45077913471587
6.167662382 1.43578883317253
6.299464703 1.41663590927824
6.400852204 1.38604438012615
6.522502661 1.36119893604629
6.633999586 1.33545619565306
6.755537987 1.30942837065009
6.87707901 1.28350746021837
6.988657952 1.25030679808691
7.120493889 1.22869098305139
7.191880226 1.20094308535197
7.313582421 1.18228360449236
7.394774437 1.15851928343589
7.516396284 1.13985985253294
7.627945185 1.10996377598531
7.759761334 1.08236066227044
7.871265888 1.05322415304249
7.962668181 1.02777142096543
8.064099312 1.00757965149844
8.165546417 0.987654494031627
8.267122269 0.964623799253411
8.359038592 0.942516369117268
8.470552206 0.920913507660293
8.592604161 0.895446498799083
8.724296332 0.869209102292539
8.825909853 0.843269513797388
8.957550526 0.823445407829723
9.069173336 0.789646847437078
9.160444737 0.769010324771086
9.272014141 0.74803872043561
9.393602133 0.725219898314556
9.505135775 0.702230286952916
9.596476794 0.680275605205985
9.707962513 0.662221384901737
9.799405098 0.63981463603654
9.921073914 0.623149397045248
10.02270627 0.603326300611571
10.154556513 0.587564290757014
10.266175509 0.565431112076245
10.387788773 0.54700974913568
10.509338856 0.531314365984393
10.62077117 0.514923981179044
10.752539873 0.504388448718393
10.8539114 0.491113142695254
10.955278635 0.481793209499084
11.076848984 0.472512267658594
11.17824769 0.461166402828284
11.269640446 0.452437021964009
11.360925436 0.442895790211094
11.472461224 0.436846422038217
11.553653956 0.428876533756885
11.624711037 0.423692764188005
11.756547451 0.419258922643361
11.867986679 0.408468163178653
11.959932804 0.40552658450246
12.071689129 0.400432946590906
12.163217306 0.393011061178839
12.264731169 0.392064987312683
12.356029511 0.398226276666296
12.467617035 0.396132834278704
12.569068909 0.396017479123561
12.670520306 0.396741360740888
12.761967182 0.392856984072477
12.873691559 0.387355803702335
12.955041647 0.385369878278223
13.076720953 0.385044357655343
13.168129206 0.378640948444457
13.269598723 0.371899342065372
13.36093545 0.369087228311688
13.47247839 0.364728161587984
13.563925267 0.361933236728472
13.665263176 0.35667109429691
13.786963702 0.354743843437608
13.858056546 0.347595226376017
13.959726572 0.34982883857135
14.071274758 0.339194930646242
14.162765026 0.328788987399747
14.264199496 0.333436808188341
14.39594841 0.328449956716765
14.497368336 0.319004763450574
14.62910986 0.321511420039292
14.740737915 0.316201660650677
14.862351418 0.312607415014537
14.973830462 0.313452561566098
15.055043221 0.306705706327125
15.166751862 0.304357054842231
15.25812316 0.300652901488833
15.370038986 0.298650411791345
15.461699963 0.294191670627483
15.573328495 0.284036390126822
15.654643297 0.286210513422999
15.766266346 0.284968304431318
15.897960186 0.275852450407374
16.049854994 0.278212691711227
16.161319733 0.2683934664448
16.293277025 0.271652948109044
16.404815913 0.267787958070133
16.496121168 0.263532684280454
16.60761714 0.260744454676725
16.688771487 0.259099366230657
16.810431481 0.256170328488587
16.891777277 0.246956201117727
17.003225327 0.245408131700846
17.094508887 0.248322615352587
17.216152191 0.246251127171464
17.327637673 0.24286293764496
17.459357977 0.242067801650158
17.570872545 0.232933699351332
17.66214943 0.230915818084389
17.773618222 0.23399459988162
17.86491108 0.228171042379749
17.976423025 0.225106083305062
18.057627201 0.22368054168137
18.179302931 0.230438930408344
18.290838719 0.222232956437779
18.412426472 0.225840893525914
18.523897648 0.225259360441671
18.655683995 0.218349165379137
18.767121792 0.223251805252616
18.858399153 0.215848276006257
18.980016947 0.211868187511587
19.101689577 0.215342519153953
19.213320256 0.212986295844496
19.324842692 0.22094562731982
19.456649542 0.208845784694461
19.558016062 0.206084335690645
19.740279675 0.206307365591646
19.851831913 0.206067912648437
19.973475456 0.202376401584939
20.054690123 0.20124511703513
20.176310063 0.200639257108746
20.257515908 0.199923722183275
20.369133234 0.205161672058484
20.460415602 0.201049737119404
20.572008848 0.195585243480797
20.663651705 0.202817377177041
20.765105009 0.200301124553073
20.85641861 0.195859299531183
20.978082419 0.194889928786337
21.089539767 0.196847807459557
21.221156121 0.196987687582296
21.32273674 0.191234198157512
21.454642058 0.190285397209092
21.556151867 0.190003275260089
21.687977553 0.189838385023164
21.789466858 0.189026094389237
21.921172858 0.188510582467473
22.042912483 0.191990439801067
22.124085427 0.189409258061954
22.245729685 0.191204915855626
22.326945305 0.189377126372402
22.438534737 0.188147052480036
22.560176373 0.186331440735318
22.671762944 0.18709119969212
22.763144255 0.185404405647022
22.864528418 0.184739640979283
22.965890646 0.187493313926835
23.07733655 0.184421758743822
23.158542156 0.188298742570247
23.270016194 0.184164533959042
23.361242771 0.188337949638962
23.482842207 0.184938986974915
23.594377518 0.192089775007322
23.726264239 0.190476065165419
23.837829352 0.19031149751078
23.929333687 0.192539372389051
24.040923119 0.191433742296222
24.132299662 0.189636706608021
24.243967295 0.189899467970845
24.35546279 0.189613387764391
24.487127066 0.18853003334128
24.598737955 0.187859845100382
24.730718613 0.187565251175354
24.832204104 0.182101256699812
24.964029551 0.181111787381373
25.065548182 0.18316862857011
25.156856776 0.179778365236969
25.268502474 0.178289164758158
25.359835148 0.179505512766326
25.501724482 0.186255954621531
25.60338831 0.182222218803343
25.694659233 0.182656863780929
25.796010494 0.187973146503673
25.927680016 0.185564659324897
26.089750052 0.186303281768399
26.191209555 0.181146466284741
26.322964669 0.184821758520554
26.435714722 0.184448755077142
26.557341337 0.184451708398824
26.679023743 0.183313273599998
26.790708542 0.180355870775224
26.922486544 0.178007775888402
27.023913384 0.178063579933812
27.155667305 0.180114540250052
27.287353516 0.179089441440278
27.388815642 0.180080467725461
27.520552874 0.179456715350459
27.632161141 0.188974938817497
27.753757715 0.188702933253311
27.824869871 0.1768489060768
27.96668005 0.178164725139226
28.06804061 0.184993389835638
28.159334898 0.180228165357584
28.270829916 0.169650905885769
28.392413616 0.173969642943112
28.503904343 0.173859748958
28.625821352 0.177562485533305
28.737323523 0.178740893547527
28.828621149 0.180104226932228
29.010771036 0.177407221672012
29.1325593 0.179679413882269
29.22394538 0.179989887370351
29.335382939 0.179759997649412
29.426649332 0.188630661764224
29.538116455 0.18197068332408
29.659627915 0.178809097981752
29.771115303 0.184279635448453
29.862697602 0.177218614380124
29.974448204 0.18596541843768
30.055954695 0.181657004012463
30.167499542 0.178597303091541
30.259178162 0.178510587159471
30.370756865 0.177312371032933
30.462192059 0.180013055937904
30.563672304 0.178925150326652
30.655120611 0.172712453535506
30.766945362 0.171978530238456
30.888732434 0.177255687084525
31.01035142 0.177660696312781
31.101880312 0.179936432390718
31.21347189 0.179895511850691
31.32514143 0.172758732913851
31.456985235 0.172871800304334
31.558483601 0.177794983128592
31.690218687 0.186419474020136
31.802013874 0.185162048694917
31.92359066 0.185545268408468
32.035135746 0.185064558430714
32.12645483 0.183352812032044
32.237980843 0.180904045855786
32.329270602 0.183156778958893
32.43067956 0.181105533395249
32.552684307 0.188203139537381
32.674434901 0.182189648799804
32.755828619 0.18222244791948
32.867239714 0.178989558809361
32.958488703 0.178005469184465
33.069995642 0.178477927620492
33.161302328 0.178316429497671
33.272889376 0.178202880063775
33.364168883 0.171807201391317
33.465551377 0.174445137121403
33.587146521 0.176697348754618
33.65824008 0.177709077561753
33.789969921 0.177652221373025
33.90163827 0.177261498268722
34.003165961 0.17740505991285
34.104572296 0.176279381896132
34.226234675 0.172032159601746
34.398252011 0.176950567982259
34.499633074 0.177334030001254
34.590940953 0.17799130281724
34.712563038 0.177832488282394
34.82423997 0.179830996153833
34.956071854 0.177954409833072
35.067591429 0.178672830658506
35.159157753 0.178721931707702
35.270722628 0.17920059681157
35.392382145 0.17855486324415
35.504040003 0.178345539022677
35.625867605 0.177360465355445
35.737408638 0.181026391454501
35.828811407 0.182507833909865
35.940318585 0.178012300918798
36.031830788 0.182439870659073
36.133165598 0.181835007183505
36.25473857 0.17888924134922
36.366215945 0.177684994999094
36.457577229 0.177595943904018
36.569316864 0.178386116815901
36.691051722 0.176440360270184
36.802678108 0.172622101642726
36.928303719 0.172435190120339
};
\end{axis}

\end{tikzpicture}

%% file: tex/realworld/e2p3.tex
% This file was created with tikzplotlib v0.10.1.
\begin{tikzpicture}

% \definecolor{darkslategray38}{RGB}{38,38,38}
% \definecolor{lavender234234242}{RGB}{234,234,242}
% \definecolor{mediumseagreen85168104}{RGB}{85,168,104}
% \definecolor{peru22113282}{RGB}{221,132,82}
% \definecolor{steelblue76114176}{RGB}{76,114,176}

% \begin{axis}[
%     width=\figurewidth,
%     height=\figureheight,
% legend style={at={(0.99,0.99)},anchor=north east, nodes={scale=0.8, transform shape}},
% axis background/.style={fill=lavender234234242},
% axis line style={white},
% tick align=outside,
% x grid style={white},
% major tick length=2.0,
% xmajorgrids,
% xmajorticks=true,
% % xmin=-1.93085986455, xmax=42.58847480955,
% xmin=-1, xmax=41,
% xtick style={color=darkslategray38},
% y grid style={white},
% % ylabel=\textcolor{darkslategray38}{Dist. to target (m)},
% % xlabel=\textcolor{darkslategray38}{Time (s)},
% xticklabels={},
% yticklabels={},
% ymajorgrids,
% ymajorticks=true,
% % ymin=0.327272642945214, ymax=4.06223502142334,
% ymin=-0.1, ymax=3.1,
% ytick style={color=darkslategray38},
% xtick distance=10,
% ytick distance=1
% ]
\definecolor{darkslategray38}{RGB}{38,38,38}
\definecolor{lightgray}{RGB}{192,192,192}
\definecolor{mediumseagreen85168104}{RGB}{85,168,104}
\definecolor{peru22113282}{RGB}{221,132,82}
\definecolor{steelblue76114176}{RGB}{76,114,176}
\begin{axis}[
    width=\figurewidth,
    height=\figureheight,
    axis background/.style={fill=white},
    axis line style={color=lightgray, line width=0.5pt},
    tick align=outside,
    x grid style={color=lightgray, opacity=0.3},
    y grid style={color=lightgray, opacity=0.3},
    major tick length=2.0,
    xmajorgrids,
    xmajorticks=true,
    xmin=-1, xmax=41,
    xtick style={color=lightgray},
    xticklabels={},
    % ylabel=\textcolor{darkslategray38}{Dist. to target (m)},
    ymajorgrids,
    ymajorticks=true,
    ymin=-0.1, ymax=3.1,
    ytick style={color=lightgray},
    xtick distance=10,
    ytick distance=1,
    % title=TARGET 1,
    % Scientific notation for y-axis if needed
    scaled y ticks=false,
    yticklabel style={
        /pgf/number format/fixed,
        /pgf/number format/precision=1
    },
]
%\addlegendentry{SARL}
\addplot [thick, steelblue76114176]
table {%
0.092746257 2.03272605059964
0.194439411 2.03290260567056
0.295846224 2.03280047506789
0.387152195 2.03245836242528
0.498736858 2.03277284190224
0.630597591 2.02941188005186
0.74241662 2.02450759494504
0.823600053 2.02032445194879
0.935168266 2.01703155500444
1.056748867 2.01240735447061
1.168245792 2.00556209174606
1.259631395 2.00137907301259
1.371222257 1.99473763334341
1.493196487 1.98807785292752
1.604615926 1.98200740191929
1.685986757 1.9751950940836
1.797436952 1.97220992906502
1.919038534 1.96620174930001
2.030517578 1.96009902419255
2.12193036 1.95488330178538
2.233628273 1.95029821090109
2.345168829 1.94329662580436
2.477068901 1.93530336080283
2.588628292 1.92909270248363
2.720574379 1.92209452189485
2.822016954 1.9142347304212
2.953767061 1.91037714756283
3.05513072 1.89916245313716
3.18675971 1.89705279325771
3.298205137 1.88677556933582
3.389438867 1.88180936982747
3.501092672 1.87597997232132
3.592688799 1.86962308688829
3.704322576 1.86410046772024
3.795946836 1.85659612377974
3.927654743 1.84985946695564
4.039196729 1.84492238860898
4.120390176 1.83495694013817
4.231881857 1.83202662356198
4.353398084 1.82651240205282
4.474923849 1.81919993838816
4.556126356 1.81213437530586
4.66758275 1.80600443391437
4.789116382 1.80022535359782
4.900512695 1.79163903369354
5.021984577 1.78616924461329
5.133519172 1.77979097069198
5.255088567 1.77355453668849
5.366514682 1.76648615281205
5.488030672 1.75962308677269
5.599403858 1.75118739165623
5.721029997 1.74536394515719
5.832426309 1.73686493569349
5.953957796 1.73010927496639
6.065304994 1.72290517397611
6.156524658 1.71641872595255
6.267874241 1.710436794883
6.389498949 1.70203323325114
6.511061907 1.69676778983196
6.622539282 1.68875384791848
6.754064798 1.68209131766155
6.865514517 1.67320904479074
6.987002372 1.66670832807877
7.098371744 1.6579546902949
7.189575195 1.65125382151353
7.30112791 1.64618991885473
7.392642975 1.63958516505225
7.494033813 1.63433059975895
7.625893831 1.62800408786558
7.737589836 1.62027726345856
7.85384345 1.61337303482135
7.975666046 1.61026629575655
8.056814432 1.6007628117341
8.168312788 1.59705824862478
8.310379266 1.59062808638725
8.462262869 1.58125723629879
8.563645363 1.57432640168584
8.654989958 1.56551174080923
8.776506424 1.56089551975556
8.888012409 1.55237063888283
9.009916544 1.54770331981828
9.121572971 1.54029695616756
9.243291139 1.53536070301343
9.395493746 1.52565771434159
9.517211199 1.51876633294646
9.628702879 1.51276743543129
9.750303268 1.50709006684194
9.871905803 1.50013339753741
9.993513345 1.4933800207059
10.115123748 1.48781085282044
10.186248779 1.48139877333622
10.327983856 1.47759100054624
10.41925025 1.46934054615626
10.550930977 1.46368644369634
10.672684669 1.45792697214252
10.784212351 1.45034603583055
10.90583682 1.4459877061796
10.987088203 1.43871291163583
11.108779907 1.43404919885079
11.22023344 1.42872996159938
11.352041483 1.42337356897491
11.473692178 1.4151402766356
11.585169792 1.4091284882197
11.706635952 1.40486604625063
11.787773609 1.39798131935755
11.899363279 1.39394993803592
11.990725279 1.38822734474312
12.092239141 1.38291743946321
12.223934412 1.38981531597712
12.345514774 1.38998480067291
12.4571352 1.38579547072704
12.578714132 1.38156233031706
12.690252304 1.37550278441547
12.821906089 1.37094348647821
12.923442602 1.36348663790953
13.055068493 1.35855942385987
13.156489134 1.35261811417559
13.288090229 1.34564273971852
13.399765968 1.33976079577309
13.521525144 1.33383117107353
13.633259296 1.32628863195951
13.754992485 1.3225560391839
13.866553306 1.31610462959356
13.957896709 1.30889049853356
14.069360733 1.30368519504531
14.191091537 1.30016069417467
14.322907448 1.29292197098231
14.424307584 1.28544694837903
14.556223154 1.28186137940175
14.66776371 1.2763663885719
14.759126901 1.27082095005001
14.870677232 1.26813142667011
14.961975097 1.2613118703807
15.083566904 1.25665625232209
15.184995174 1.25254228957701
15.326819658 1.24871036965593
15.418194294 1.24288329234762
15.550132036 1.2381770061947
15.661795616 1.23210575151587
15.783568859 1.2284846523514
15.895141363 1.22218897341318
15.986377239 1.21783623374341
16.097892046 1.21439402899711
16.189217567 1.21014275725041
16.300673484 1.20669115289964
16.422339201 1.19989742765607
16.543881178 1.19707704927796
16.66543746 1.19044393290319
16.787001848 1.18659022542977
16.908584594 1.17888323571269
17.020030975 1.17517148260328
17.151806116 1.17073016827582
17.263355016 1.16462266274308
17.394977569 1.16009430031531
17.536801099 1.15436396209542
17.658377647 1.1483681618026
17.790016174 1.14453555593719
17.901561975 1.13957879703914
17.992813348 1.13367075980834
18.094392538 1.13159565050148
18.226169109 1.12656941293342
18.347820997 1.11917846014266
18.459360838 1.1153224219234
18.591066837 1.11193166138312
18.702700376 1.10660786846658
18.794224977 1.10206489921653
18.915958881 1.09706109332773
19.027486801 1.09463992529915
19.149032592 1.09140269952488
19.250521183 1.08755169565772
19.392622947 1.08238167915434
19.504172325 1.07938557263055
19.585419178 1.07555461786482
19.697027445 1.07285426031958
19.788396358 1.0688694972814
19.900036096 1.06616851999912
19.991410494 1.06066814114965
20.103143453 1.05943138913139
20.184346437 1.05552896699712
20.295769691 1.05489468221204
20.427416801 1.04893736550233
20.589470625 1.04456747888477
20.701218128 1.03858013049992
20.822947263 1.03496313708499
20.934466362 1.03145938027931
21.02578187 1.02818563824531
21.137245893 1.02499308569067
21.218410253 1.02275669221553
21.350200891 1.02183948124854
21.461687326 1.01685735904623
21.583280801 1.01320570563909
21.664448261 1.0107723110798
21.796101093 1.00717659321171
21.897429228 1.00296058065897
21.988693714 1.00112697705342
22.100158453 0.997421295605873
22.201573848 0.993417435483974
22.363496303 0.99042298211209
22.454742431 0.984283326825135
22.586374759 0.981558057414051
22.697935104 0.978806059946136
22.819472313 0.973520207200982
22.941046953 0.969721207504322
23.052533388 0.966757715910555
23.184133768 0.962392858660293
23.295634269 0.957768885420656
23.427267074 0.953772517909998
23.579361438 0.951765673367491
23.701023817 0.948394995209711
23.86328578 0.943938436140893
23.984818697 0.941380983587488
24.09628582 0.938863934787856
24.227977752 0.934616570401629
24.33948493 0.932078460884682
24.420611143 0.929374429873852
24.532224178 0.927028090452313
24.65388298 0.924574691075414
24.886533737 0.917986613858576
24.998262167 0.915497273275747
25.089728832 0.913144713649993
25.201389789 0.910462648430115
25.302944898 0.907275766999157
25.404438018 0.905435111688508
25.485922336 0.90314252172526
25.597362756 0.900764287503134
25.688667059 0.899724869638817
25.800404548 0.89731335700063
25.921995639 0.895320820827708
26.033652067 0.891411434883091
26.155250787 0.889363368844142
26.297091961 0.885855819489982
26.408564567 0.882401701362998
26.520056963 0.88053353203985
26.641675949 0.87685879058501
26.753169298 0.875111830362145
26.884890556 0.872071915254577
26.996363401 0.869991011352374
27.117922306 0.866243072383749
27.229438066 0.866189742138995
27.351204157 0.863389530800626
27.462805509 0.860824406417834
27.584470749 0.858786632316141
27.696113825 0.85697410657261
27.817788124 0.85450145976246
27.929408073 0.852490060410109
28.051193952 0.851767551121327
28.172848463 0.848588770903884
28.254068613 0.847342084089601
28.385902881 0.8450450327419
28.487395763 0.843218493636426
28.588900327 0.841497093423757
28.700481176 0.839698032360047
28.822185754 0.83716142542306
28.953937292 0.835752780316741
29.065478563 0.83290273769327
29.166903495 0.829793451775156
29.278414011 0.82927178271453
29.389914751 0.827205958678261
29.521631241 0.824528486747109
29.633152246 0.821061341210034
29.754767894 0.818614749882224
29.866287946 0.816086525214262
29.957645893 0.814842944163104
30.069292545 0.811905395071614
30.160559892 0.811119294732717
30.272122144 0.808700182373076
30.353387356 0.805796277953507
30.515521288 0.804355545138278
30.627018928 0.80209593891924
30.75886774 0.800919087378371
30.870446443 0.801318233568207
30.961958646 0.800413060111809
31.063568115 0.799314678476478
31.154802322 0.798845435805703
31.276325464 0.797046451544892
31.397878885 0.797264201134544
31.519534349 0.795336424510455
31.62099123 0.794722513689754
31.752613067 0.793631792323558
31.864070654 0.793043244273122
31.985633135 0.791391797997074
32.107186794 0.789698678122774
32.218639612 0.789945879346638
32.350319624 0.788781881128937
32.451779842 0.788857713655786
32.583614826 0.788948218344913
32.705128431 0.787901169301762
32.826719522 0.787699350362199
32.948393106 0.787169209898034
33.059862137 0.780998168678452
33.191622019 0.784517426545948
33.303122043 0.782452351608974
33.404462099 0.782468045492984
33.515959263 0.786387689894825
33.627421856 0.784676029663024
33.759194374 0.784696277950067
33.870717287 0.787231392674718
33.962098837 0.784144507585889
34.073630809 0.786588184533318
34.155019521 0.787116522286418
34.266516447 0.784464233033959
34.35781312 0.788299590758225
34.469369888 0.787953217830191
34.560652256 0.78544020659536
34.672158956 0.789099415741549
34.753326177 0.78882527742079
34.874918222 0.785097165812333
34.956156492 0.789040529678636
35.077689409 0.788856110787457
35.20939517 0.785768348708628
35.351329565 0.789086863000919
35.523330211 0.785416290050323
35.634782552 0.788907938764097
35.726133108 0.7867506800928
35.837611437 0.788372816403022
35.959354639 0.789041188269421
36.081166744 0.784874832752973
36.192754745 0.788626071839212
36.324522733 0.788119574836723
36.446252346 0.785986143258708
36.557761192 0.789064411312792
36.689530849 0.784671507674477
36.811074972 0.788880406525372
36.942989349 0.786719647082424
37.024198293 0.786673653421375
37.145798921 0.785154997755533
37.257217169 0.786995122119789
37.398953438 0.787005883339753
37.500347376 0.786788921536195
37.591635942 0.787182715576761
37.703105926 0.788711624149958
37.794431209 0.786681306433338
37.90598607 0.786278532341916
37.987194299 0.790352836663192
38.109165191 0.789254257242546
38.220582485 0.786281640951348
38.342488289 0.790030909038694
38.453971624 0.787205276663915
38.585791826 0.787398946706184
38.707520246 0.788491757814523
38.819048643 0.786534058230369
38.950711012 0.788413086645774
39.05210638 0.786909673969188
39.193918705 0.78865757681718
39.305466652 0.787642651305788
39.386763334 0.787151054269912
39.498209476 0.789908412713233
39.619963884 0.788035194488545
39.731460809 0.78866389187316
39.853155851 0.789438634036703
39.964612245 0.785526227203096
40.096265077 0.788299708994128
40.197633266 0.787651157545949
40.319283485 0.786987008319197
40.45113945 0.789115811353887
40.564868688 0.785869389306824
};
%\addlegendentry{MARL (global)}
\addplot [thick, peru22113282]
table {%
0.112878799 1.97621427334967
0.214924812 1.97626711806532
0.316340685 1.97618159741813
0.438024282 1.97661583332489
0.569836616 1.97724498542921
0.671271324 1.98472413594646
0.813064336 1.98722630654425
0.92452383 1.9993825491645
1.036002874 2.01276996562351
1.167828798 2.02215112765885
1.279314279 2.04387303546334
1.410994291 2.05500340048968
1.522503376 2.07079317449149
1.613772153 2.08574357840568
1.725290537 2.09395477058401
1.836997985 2.10880016743329
1.958567381 2.12066266149451
2.070230007 2.13302477418452
2.202110529 2.1464911854085
2.313668966 2.16084391343083
2.445377588 2.17397730369966
2.546778202 2.19175860935229
2.678448677 2.21003082749963
2.789943933 2.21505029522622
2.871131181 2.22715413311476
2.982661247 2.23766477368158
3.124457597 2.24957654923298
3.215801239 2.2603971760259
3.337396383 2.27638223160279
3.448896169 2.29232975437455
3.57051444 2.30390840340849
3.682114362 2.3175118547966
3.783906936 2.33067240929075
3.885326147 2.34224323297583
3.976834774 2.35368022280844
4.08831954 2.36376387416861
4.210041523 2.38061380557897
4.331680536 2.39122698101385
4.443166971 2.40637349857171
4.574876785 2.41692144990865
4.686525344 2.43339670866552
4.808006525 2.4478002456289
4.919521093 2.45865107354639
5.010836601 2.4690486359344
5.132483482 2.48121405293955
5.243885755 2.49822718814568
5.375420332 2.51051388473829
5.487003326 2.52867455131605
5.608556032 2.54111176250236
5.72007823 2.55357463284119
5.84173727 2.56797601978198
5.973479986 2.58338681450622
6.074849844 2.5984387887508
6.206581354 2.61548099637183
6.328219175 2.62542718176539
6.450045347 2.63957239827937
6.571694135 2.65274588686154
6.6833117 2.6651117410694
6.805025339 2.67712637047849
6.916631221 2.69144373586968
7.007945776 2.70200875097795
7.13975048 2.70964969025525
7.241183996 2.72569007574697
7.372745752 2.73528508071162
7.474050521 2.74759738841074
7.605576515 2.76228788947429
7.706870317 2.77619039969485
7.838460445 2.78819188473053
7.949878692 2.80292152541067
8.071373224 2.81794443812873
8.233346223 2.83413894788322
8.344834804 2.84849754653139
8.476555347 2.8543339295728
8.588061094 2.86477533651456
8.67929244 2.87586668578012
8.790631055 2.88872134793171
8.881777286 2.90245359508826
8.983267545 2.90955828723589
9.074509859 2.92233706765471
9.186006546 2.93464658702033
9.277286529 2.9449863939449
9.388738393 2.95722196673729
9.469896078 2.96836751219405
9.591390609 2.97776496338376
9.672594308 2.99173110958067
9.784068346 2.99936934972319
9.875371933 3.01187217159124
9.987012386 3.02296127187668
10.079393863 3.02983815135416
10.191002369 3.04317533258414
10.302704095 3.05682026802909
10.434437513 3.06915530837529
10.54596281 3.08267128173499
10.677663803 3.09836044763428
10.789100408 3.11249563550143
10.880452633 3.12138079748355
10.991990566 3.12953744105979
11.103625536 3.13962963595483
11.245594501 3.14418390929562
11.347211599 3.15630124211592
11.468841552 3.17146973586316
11.610739469 3.1820635613563
11.712146043 3.19690598950985
11.843868017 3.21001075535737
11.955340623 3.22187167852173
12.076915025 3.23692178934966
12.198527813 3.24736721342247
12.279834985 3.25930536508333
12.381293773 3.27042944683554
12.513154983 3.28134187457585
12.614608764 3.29919549208759
12.736382007 3.30694365690848
12.858050346 3.32324982982167
12.939308881 3.33428955137051
13.050855159 3.34639921178876
13.152396917 3.35884434275981
13.25387001 3.3687204682901
13.375457525 3.38019098558416
13.487045049 3.39263799990459
13.608677625 3.40524038670697
13.730267524 3.41495826883207
13.811461925 3.42793515563223
13.922861576 3.43584535099336
14.014179706 3.45164216234846
14.176310777 3.46045846483089
14.287825107 3.47761911137835
14.379195451 3.48794402241272
14.490801334 3.49776547518755
14.58223176 3.5102627135002
14.683734178 3.51880197006126
14.815418243 3.53253286115711
14.937059164 3.54558408497661
15.048570633 3.55766185285297
15.190378904 3.56917807776755
15.332069158 3.58662801711685
15.453895807 3.59899629890903
15.545155763 3.61063537017765
15.646557808 3.62020916871332
15.778283834 3.63078804834319
15.889838218 3.64328923122087
16.183249473 3.66907662240516
16.284584522 3.68679258659184
16.406445741 3.69784606489059
16.528026819 3.70946289084394
16.639600992 3.72255308282191
16.771320581 3.73420605242109
16.872684002 3.74738047111355
17.004320621 3.75881222379417
17.105748891 3.75617180055852
17.237502336 3.76425630751276
17.359247207 3.77744739108431
17.470730066 3.78700762953162
17.602469921 3.79494848479961
17.714009761 3.80314541224794
17.815429926 3.8117076335252
17.967395782 3.83116300400419
18.079002142 3.84292600793289
18.240962267 3.85683937362834
18.352503299 3.88434765346732
18.494283676 3.89246400421979
};
%\addlegendentry{MARL (partial)}
\addplot [thick, mediumseagreen85168104]
table {%
0.102184534 2.02737929216644
0.213835001 2.02744048996924
0.305140972 2.02807096413874
0.406527996 2.03602655280086
0.51809454 2.02475628718594
0.619523764 2.03249478877163
0.710754394 2.02490707294081
0.842639446 2.01879255985553
0.954053163 1.99518823882732
1.075801849 1.99066947436812
1.197393894 1.96975679223414
1.309116125 1.95697602227847
1.431080103 1.95215478247408
1.532797098 1.93517601655676
1.644406795 1.9201864077092
1.755860567 1.90561722998951
1.867279768 1.89701505025314
1.988897562 1.88129115435184
2.100409746 1.87190893083514
2.232454061 1.85578751209253
2.334009886 1.84613953680804
2.475940943 1.83341729096497
2.567225695 1.82110607260531
2.709121227 1.80425805986138
2.810518265 1.79640800055766
2.932108641 1.78492154014625
3.053903341 1.77016856311511
3.135202169 1.76281103778112
3.246805429 1.75285669017284
3.368474483 1.73993309128734
3.490278006 1.73064219258346
3.612017155 1.71607051804919
3.723601103 1.70754877189412
3.835143328 1.69495933314601
3.96689415 1.68201820631557
4.078384399 1.67153071696221
4.169744015 1.65750510887177
4.281166315 1.6470833705232
4.402868032 1.64082761813469
4.514329433 1.62227168274857
4.615871668 1.6132396215488
4.757580519 1.60594136142975
4.869273663 1.58838708181789
4.990851879 1.58243130102257
5.112486362 1.56302958235241
5.234079123 1.55285715111374
5.345631361 1.53776779376964
5.437021732 1.52757719988099
5.548541784 1.5197817922042
5.670097113 1.50430264390266
5.791688919 1.49414506810748
5.903214693 1.48411744057192
6.034986734 1.46800416474742
6.146525383 1.45916435207192
6.268128395 1.43613886306085
6.37960887 1.43602679086949
6.470905304 1.42321929082468
6.582540035 1.40923417784523
6.673838615 1.39941959501624
6.785425186 1.39137107180545
6.876669884 1.3844342680552
6.978136539 1.36970550489093
7.069410801 1.35755561037445
7.190936565 1.34934330420168
7.302551985 1.33144114129448
7.424249887 1.32415043683035
7.505503893 1.30227223345837
7.637428522 1.29306451692382
7.738950014 1.28616878500868
7.870642185 1.26590511595629
7.982140064 1.25484071492524
8.103881597 1.2474684123554
8.215378284 1.23844544920127
8.337021589 1.21907803701062
8.448535204 1.21138487033998
8.570157289 1.19786432780808
8.691752434 1.17966957631634
8.803401232 1.1706784534385
8.965462208 1.15386062964666
9.097173691 1.14126413719503
9.208756924 1.12376979895549
9.330355883 1.10741579000801
9.442072868 1.10198340235328
9.573884249 1.08218600398641
9.695471048 1.06590467955674
9.776703119 1.05159582199962
9.888262033 1.04096679338375
9.999767303 1.03547641019241
10.131473541 1.01752051852891
10.232907534 0.997290316792842
10.364583969 0.988102734943475
10.465961695 0.97197632429898
10.607738018 0.955996877914213
10.709172487 0.948542645122586
10.840851068 0.935196945315915
10.952470779 0.914000942882141
11.033629656 0.903250923731212
11.155171394 0.902331818975694
11.236355782 0.884929806258418
11.357972145 0.86650800366762
11.469509125 0.856859399208852
11.601293325 0.84065934445708
11.702831268 0.827278253031865
11.834609032 0.816899702400356
11.946046829 0.804377699126028
12.067797422 0.798206024131537
12.179426432 0.779416723356841
12.311124086 0.767906528703543
12.432849407 0.762552957450582
12.534374952 0.74455703989527
12.676245451 0.729305975225055
12.787714243 0.72198755726231
12.909292698 0.722655857296341
13.010756493 0.702022668468009
13.142610073 0.701360133828737
13.25407815 0.683561150929837
13.33528471 0.671495918580157
13.45711112 0.667843396816169
13.568648338 0.656465093651217
13.700322628 0.653302837922013
13.801737547 0.643361480800652
13.933481693 0.645939435828569
14.034798861 0.638403732851731
14.166400433 0.621344270839008
14.277824163 0.615853461873468
14.399502277 0.611962111235905
14.511102676 0.610354588090129
14.643081903 0.600505053644187
14.744415045 0.591586974473245
14.83569479 0.591334494459006
14.947163582 0.588548201039968
15.038433552 0.586436181815164
15.149892569 0.576577746261499
15.241194248 0.578826624740311
15.352746725 0.566957055145164
15.444088697 0.56772739769022
15.555588245 0.558899926936898
15.667104244 0.557839098458114
15.798747539 0.56297165599005
15.900259972 0.552733389274371
16.031998873 0.566234512728496
16.143711567 0.557312144452944
16.275501728 0.55437980077228
16.366759539 0.548751299998393
16.508559227 0.551474799335061
16.609934092 0.567636816440134
16.761815071 0.542013224538155
16.903833151 0.532491225063384
17.005206108 0.534460704959257
17.137061596 0.533142835226265
17.248550653 0.531039225153019
17.339997053 0.528725977233294
17.451570034 0.5320911853556
17.543304205 0.53853711971218
17.675096989 0.53805440488513
17.776489735 0.5361716847977
17.898066521 0.535664680852438
18.009713411 0.527071722422026
18.141473532 0.527427712824816
18.253419638 0.526784864496324
18.334750175 0.527180596923143
18.446319103 0.520882256629608
18.537587166 0.564233914263164
18.649101257 0.564506372100146
18.770682812 0.526992875995321
18.892567635 0.526994569969067
19.004157305 0.527230192105846
19.135878801 0.526659936332783
19.247303247 0.523024657003433
19.338539839 0.524441297050767
19.450068235 0.522577616648697
19.541437626 0.520418081482664
19.643148422 0.520536199460582
19.744495392 0.517111653752782
19.856120825 0.517738224856814
19.967686892 0.518159617942412
20.099467039 0.517632267571081
20.200841904 0.51457176159657
20.332621336 0.515946930549886
20.444252014 0.514417500028307
20.565928221 0.515079124689374
20.687688112 0.515692712965583
20.769005299 0.515419621347992
20.890725136 0.548317437335428
21.002378702 0.513495565701881
21.124169826 0.53511588902294
21.205353975 0.513668637378124
21.327120542 0.512619690168027
21.438779592 0.512482597894992
21.570604324 0.512209407041939
21.682156324 0.512268559360812
21.773504257 0.512347100125739
21.885037422 0.512246353138066
21.966288805 0.511558707238671
22.087849855 0.511319956865932
22.16900444 0.510717207888661
22.28046155 0.510610164032066
22.381863117 0.509245276179033
22.483333111 0.505830681493974
22.574625969 0.511456501034707
22.716529131 0.509128335765536
22.807836771 0.510075869537094
22.939494371 0.509884375436737
23.051062107 0.508556776316166
23.142693043 0.522628315566598
23.254245043 0.51461613707737
23.335480452 0.506688608819848
23.446971893 0.508603577385711
23.538542032 0.50928780442886
23.650600672 0.50777458029683
23.772362709 0.506960991783622
23.893951654 0.535663729270933
23.975325346 0.508187592005468
24.096958399 0.509393214346404
24.208457947 0.513920893741291
24.370424509 0.512424630844182
24.522582769 0.51259140067821
24.674461842 0.510500332497514
24.826427221 0.511428761741647
24.937967777 0.51307005928763
25.059698582 0.51094552436542
25.171281099 0.510936110608217
25.303167343 0.513257625453774
25.404632091 0.512776530122448
25.536642075 0.507846944315748
25.648453712 0.507138065263737
25.770160913 0.506704687040656
25.891814947 0.506508247713061
26.003300905 0.505622082454459
26.135010719 0.504223188420429
26.246543646 0.542692310121634
26.337875605 0.510397948519534
26.449429512 0.520243233062102
26.571026802 0.514653608621317
26.692673683 0.504019574710407
26.804148912 0.507700851532411
26.925827026 0.508155752956406
27.037363291 0.50815482276499
27.169051409 0.507433586588633
27.280567884 0.506896802880309
27.371875763 0.505949754037365
27.483665943 0.506112964994913
27.575078487 0.505842458662663
27.727113485 0.505491255037998
27.848711491 0.520246179185258
27.940028667 0.531118977497108
28.061636686 0.529874577072669
28.173243523 0.510081014189323
28.294883728 0.513737289407572
28.406508207 0.510773435575023
28.53822279 0.509848063156535
28.649705172 0.513349806964105
28.771289587 0.505876713010347
28.893132448 0.506358393910494
29.00476861 0.50354820434851
29.136762857 0.507800931032329
29.248262644 0.505804439661683
29.339570284 0.505802324333359
29.451102495 0.507095196703613
29.542678595 0.506861488448947
29.654274225 0.505959003921278
29.775872469 0.505676011236637
29.897426128 0.507618159779516
29.999090672 0.507456480187296
30.140952349 0.505769325409844
30.252477169 0.506609614589651
30.374123096 0.505273087778998
30.485636473 0.502279777773326
30.566848278 0.503985933322535
30.688414097 0.503514457738753
30.799963474 0.503784061339351
30.931628227 0.502943416568514
31.04317832 0.505396155472666
31.164794922 0.514452964503869
31.235903501 0.506533910276139
31.357514858 0.497043660148765
31.468996763 0.504150962527672
31.590594053 0.509695007268412
31.712187052 0.509451537039129
31.83384347 0.505782876946177
31.945369482 0.506749736050786
32.036802769 0.505230646063442
32.148352623 0.505942310167455
32.239797592 0.506208605369829
32.351330996 0.504801596447871
32.44268918 0.504443172005489
32.554266214 0.502453631397994
32.635429144 0.535189016629041
32.767113686 0.50777403091312
32.868566751 0.501487699210967
33.010349035 0.502127853265114
33.111683369 0.508265607878219
33.24332428 0.504321240078846
33.354777098 0.53840879886518
33.446046352 0.528262230933488
33.547432661 0.505620525003009
33.638735533 0.538067293136794
33.770609379 0.506106619761746
33.882196665 0.5010440436634
34.003874302 0.505949227413328
34.115363836 0.509404516314315
34.237282276 0.538366900929914
34.348884344 0.514984136480436
34.440211534 0.510291758501115
34.551745415 0.5044925258901
34.643258572 0.503913605868877
34.754737139 0.504163936291149
34.84605217 0.51622149689083
34.947411299 0.504119459764327
35.038681984 0.504379506718537
35.150151253 0.507613407146156
35.241416931 0.504673667378252
35.353068829 0.504933794063933
35.444331408 0.504553999075144
35.545800686 0.503986622771035
35.637112141 0.504533152466781
35.748610973 0.532122663812889
35.840135336 0.524478484973936
35.951642513 0.500671347479943
36.04330492 0.532748232236429
36.14467454 0.511548206811537
36.235966682 0.505349253318219
36.347437143 0.507476433633124
36.438831568 0.512802179640089
36.560446501 0.532422532446036
36.671969175 0.503745772895379
36.803840637 0.516667650322486
36.905201912 0.505342217126061
37.037024736 0.504225985313543
37.148559809 0.505407185267511
37.239951611 0.519179516170546
37.3515172 0.509926194488459
37.442930698 0.503611489695341
37.554457664 0.53319115578246
37.686286449 0.505929517386722
37.807969093 0.506659712131354
37.949814558 0.507697199461121
38.041119575 0.504038036353491
38.172908068 0.52493933124057
38.274245262 0.505508345857066
38.406039 0.504851401340064
38.507479191 0.504926359835766
38.63930726 0.506767888708453
38.771051884 0.504006033727501
38.872597456 0.504966585349193
39.004346848 0.50373100834995
39.105897903 0.527347790593895
39.23764658 0.530561899719094
39.349120855 0.50210951230578
39.440476894 0.523113592549158
39.551990986 0.511071678048164
39.643442631 0.513272346297153
39.76515913 0.506916788316393
39.866687775 0.508279179190356
39.998607397 0.508008325276807
40.110356092 0.505223402486442
40.242121458 0.521615719163261
40.35366416 0.504071970587193
40.458749056 0.502021273417828
};
\end{axis}

\end{tikzpicture}

%% file: tex/realworld/e3p1.tex
% This file was created with tikzplotlib v0.10.1.
\begin{tikzpicture}

% \definecolor{darkslategray38}{RGB}{38,38,38}
% \definecolor{lavender234234242}{RGB}{234,234,242}
% \definecolor{mediumseagreen85168104}{RGB}{85,168,104}
% \definecolor{peru22113282}{RGB}{221,132,82}
% \definecolor{steelblue76114176}{RGB}{76,114,176}

% \begin{axis}[
%     width=\figurewidth,
%     height=\figureheight,
% axis background/.style={fill=lavender234234242},
% axis line style={white},
% tick align=outside,
% x grid style={white},
% major tick length=2.0,
% xmajorgrids,
% xmajorticks=true,
% % xmin=-1.85868511195, xmax=40.82899198495,
% xmin=-1, xmax=41,
% xtick style={color=darkslategray38},
% y grid style={white},
% ylabel=\textcolor{darkslategray38}{Dist. to target (m)},
% xlabel=\textcolor{darkslategray38}{Time (s)},
% ymajorgrids,
% ymajorticks=true,
% ymin=-0.1, ymax=3.1,
% ytick style={color=darkslategray38},
% xtick distance=10,
% ytick distance=1
% ]
\definecolor{darkslategray38}{RGB}{38,38,38}
\definecolor{lightgray}{RGB}{192,192,192}
\definecolor{mediumseagreen85168104}{RGB}{85,168,104}
\definecolor{peru22113282}{RGB}{221,132,82}
\definecolor{steelblue76114176}{RGB}{76,114,176}
\begin{axis}[
    width=\figurewidth,
    height=\figureheight,
    axis background/.style={fill=white},
    axis line style={color=lightgray, line width=0.5pt},
    tick align=outside,
    x grid style={color=lightgray, opacity=0.3},
    y grid style={color=lightgray, opacity=0.3},
    major tick length=2.0,
    xmajorgrids,
    xmajorticks=true,
    xmin=-1, xmax=41,
    xtick style={color=lightgray},
    ylabel=\textcolor{darkslategray38}{Dist. to target (m)},
    xlabel=\textcolor{darkslategray38}{Time (s)},
    ymajorgrids,
    ymajorticks=true,
    ymin=-0.1, ymax=3.1,
    ytick style={color=lightgray},
    xtick distance=10,
    ytick distance=1,
    % title=TARGET 1,
    % Scientific notation for y-axis if needed
    scaled y ticks=false,
    yticklabel style={
        /pgf/number format/fixed,
        /pgf/number format/precision=1
    },
]

\addplot [thick, steelblue76114176]
table {%
0.092444182 0.766706348474454
0.184252501 0.766570492827065
0.255344868 0.766815666919978
0.316370488 0.767435159214142
0.387472392 0.76625594997286
0.458733798 0.76527352455418
0.59065342 0.762919647674793
0.692143918 0.762054014008127
0.823871851 0.759307898959179
0.925225974 0.756618464291852
1.057037831 0.753844691565031
1.168576003 0.752609554028237
1.259999037 0.749618857744723
1.361508608 0.747476996511347
1.463040352 0.748055161230154
1.564382077 0.748791247050527
1.68603158 0.749189669324895
1.7571342 0.754583722717317
1.88886714 0.756435976434048
1.99032116 0.756635311544275
2.122100592 0.762281057657442
2.193236351 0.767327533098175
2.315051794 0.767364289019911
2.386134625 0.76844291619198
2.467307806 0.772094247181493
2.558573246 0.773578926036536
2.660155297 0.776662183974789
2.791789294 0.777278834825345
2.903266669 0.779266940186822
3.024807692 0.781775650848596
3.146373749 0.783082306549177
3.227578164 0.784744639647033
3.328971386 0.784759324236703
3.420320511 0.78583649606071
3.531940222 0.787070058262912
3.62361002 0.787796660828806
3.745280028 0.78807836808266
3.826479197 0.789119425186145
3.937951327 0.789576661839024
4.01923585 0.790499591910012
4.141100646 0.791601194629224
4.22240448 0.791648833452363
4.333936453 0.790343704062551
4.455461264 0.790852674445113
4.526561499 0.791112007432213
4.648155928 0.791592615282751
4.719338179 0.792241278347346
4.840998412 0.792648600187733
4.922147751 0.792964981042836
4.99317813 0.792919165070597
5.124771119 0.79296188705361
5.226214648 0.793667013435669
5.347734929 0.793553202944587
5.429053307 0.793905240492589
5.540598393 0.793799458078393
5.631941796 0.794027038946337
5.733352423 0.793888665780361
5.864970446 0.793980797933313
5.966289282 0.793729640380028
6.087857485 0.793843188462909
6.158925295 0.793936776040613
6.29057026 0.794075951496686
6.39196682 0.793890420585684
6.523655177 0.793894275077171
6.625001669 0.793827747030743
6.756647826 0.793863763358568
6.868096114 0.793913929989677
6.989973784 0.793844016073194
7.111626864 0.793974367305499
7.19285822 0.79401681375905
7.3042655 0.793978724076084
7.425911665 0.793895888582759
7.537316561 0.794027777650442
7.628587246 0.794038314360916
7.730089665 0.793968063568362
7.821367026 0.793925862089625
7.90257597 0.793948947748274
7.993847847 0.793987310995026
8.095209837 0.794022294537873
8.186803818 0.79397200974698
8.257940531 0.793912092361733
8.400368691 0.794175031187199
8.501865626 0.793890626217447
8.623724223 0.79392882817248
8.735404492 0.793963215662124
8.857087136 0.793928167788826
8.968747139 0.794029309626753
9.060091258 0.793972755269302
9.161581994 0.793816716667574
9.263164044 0.793908179396855
9.364588023 0.794029246700166
9.486283541 0.793805595149099
9.597748042 0.793974038419527
9.7295115 0.793796944994473
9.830948353 0.793935672200942
9.95265317 0.793965239128806
10.023783684 0.794007593563947
10.15547514 0.794033127533886
10.216507197 0.794109526453378
10.297677994 0.793961521327796
10.419228554 0.793950784597063
10.530755997 0.793805092663894
10.65272832 0.793837995205061
10.76420188 0.794129785813476
10.885801793 0.793928420790805
10.956980229 0.79387326778873
11.088639975 0.793874886937985
11.159919501 0.793948875717651
11.291794777 0.793888794985374
11.403352976 0.793945026798314
11.524897099 0.793974757592306
11.636320591 0.793930880972127
11.757857562 0.793941322785245
11.86943841 0.793907329478065
11.991102934 0.793982205939972
12.102556944 0.793917978255849
12.224115849 0.793973713405377
12.335861445 0.794007631968331
12.457418204 0.793966631545323
12.569184065 0.794000070439543
12.700976372 0.794122467978571
12.802423716 0.794141783859731
12.893639565 0.793968572517029
13.005100489 0.793924485287421
13.086463452 0.793832639631699
13.157570839 0.794011474880717
13.289372445 0.793981356408595
13.390850544 0.793901502670334
13.522620917 0.794050597429609
13.6342659 0.793999318526471
13.755938054 0.794004321131931
13.867421151 0.793903558577195
13.989055396 0.793996297232425
14.100535393 0.794015478083882
14.222269297 0.794131147755069
14.33399725 0.794008231031887
14.425381184 0.793971122852866
14.536908627 0.794013851573402
14.658563614 0.794003659769187
14.770323754 0.794094578969801
14.891956091 0.794044766244553
15.003522158 0.793908270370016
15.125068427 0.793974824616975
15.236513615 0.793990684631744
15.35807848 0.793938878243075
15.459457398 0.79402074097429
15.591222525 0.794087903487106
15.702745438 0.793911926084627
15.82434845 0.793935169366133
15.93586588 0.794005970339794
16.027143717 0.794016680434947
16.128492117 0.793916806980849
16.219905854 0.79392580521504
16.35167408 0.794042809163559
16.422849656 0.793930651757174
16.544656754 0.793918007917448
16.625894309 0.793953309256596
16.747455121 0.793970650229955
16.818823815 0.794009921165131
16.900031329 0.793929604129633
17.021678448 0.793911517852888
17.09276867 0.794001706970345
17.224524022 0.79404389655132
17.325987339 0.793908872373305
17.447601796 0.793924044040188
17.518798352 0.794004498847452
17.600200415 0.793936356120856
17.721802474 0.794076185248798
17.793123723 0.793973700235668
17.924836398 0.793986145708342
18.026295901 0.79401192744074
18.147967101 0.794009586466959
18.229264021 0.794097571457783
18.341104269 0.79391370135477
18.422415972 0.793991224683215
18.534056426 0.793963723731466
18.625385762 0.79407963374849
18.737200499 0.794046444448828
18.818638802 0.793965796672546
18.889924765 0.794098859134324
19.021754027 0.7940205135092
19.092885733 0.793857219720374
19.20436287 0.7940059306388
19.285807133 0.794007617107126
19.356985093 0.794141786659204
19.488702536 0.794165229354862
19.559942723 0.793934706944608
19.691712141 0.793918234710626
19.803234816 0.793935929661822
19.894517661 0.793982904894807
19.995902539 0.793976218233846
20.117560864 0.794112564025721
20.188714028 0.793970760221308
20.28005004 0.794081156979922
20.391530991 0.793941041074342
20.513076783 0.793952241111103
20.594180823 0.793920614772139
20.695522309 0.79402363727149
20.817201138 0.794042043453452
20.888328076 0.79408217121074
20.979747296 0.794002085836829
21.091287375 0.794111598362197
21.223118306 0.793994766878508
21.344962836 0.79398624363096
21.426218749 0.794076022784332
21.547767401 0.793935190317048
21.62902236 0.793876678333497
21.740551234 0.793992378883778
21.831854582 0.793988245846564
21.933372498 0.793972609533238
22.055007935 0.793966185218227
22.126328707 0.793998832232078
22.247922183 0.794015140484175
22.319246531 0.793825648447062
22.40049243 0.794025657603656
22.532124043 0.793882210151474
22.633514643 0.794039749430858
22.755102158 0.793978382853315
22.82627368 0.79396381649087
22.957998753 0.794055892306802
23.059573889 0.794071606370379
23.191500903 0.794069766911801
23.292888403 0.794129486256296
23.424528361 0.793946249618114
23.52595377 0.794002122947039
23.657754899 0.793950981270641
23.759284974 0.793999643562127
23.891047478 0.794070774512748
23.992369891 0.793962284875639
24.124011279 0.794009608175115
24.225417853 0.793946446777481
24.357266427 0.793934783210645
24.458919764 0.794024742344062
24.590704442 0.794028777509621
24.692116738 0.793964072656858
24.823957444 0.793921828813163
24.935746432 0.79403483511212
25.057378531 0.793992758359334
25.168986321 0.794000155565141
25.290611268 0.793979955448462
25.402727366 0.793882745024689
25.484074116 0.793967121265674
25.555118561 0.793958173839396
25.646571637 0.793873824926906
25.727720499 0.793925933841799
25.829062701 0.794094161818001
25.960681677 0.793878421577391
26.062048197 0.794117318379405
26.15328908 0.793968821992888
26.224432707 0.793965595348793
26.356054783 0.794056276351631
26.467516423 0.793855196636815
26.589045763 0.79401569585743
26.700559617 0.794203776401936
26.832165957 0.794027319237181
26.933557988 0.79386707785666
27.055356503 0.79385615133098
27.126465083 0.793947755601144
27.248060704 0.79398426438081
27.319173575 0.793798408702599
27.40039301 0.79388565394943
27.521860362 0.793848886881103
27.592927695 0.794040990264637
27.724574566 0.794057948823747
27.826198817 0.793928636377347
27.957880021 0.793890705891438
28.069447995 0.79415255870229
28.160746813 0.794130037949441
28.262099505 0.794045573827573
28.39393735 0.794083183524976
28.495306492 0.794108856309647
28.627040863 0.793990390467589
28.728378058 0.794111043616647
28.860163928 0.794086671248096
28.961703539 0.794016462481084
29.09359932 0.793934293459701
29.195035458 0.794005091038425
29.326820612 0.793963139546971
29.428208113 0.793927682086972
29.600620509 0.794007111000291
29.823279381 0.794094130012014
30.025688887 0.793960679452903
30.238555193 0.794026708645448
30.441114188 0.793989339035941
30.633949995 0.793886241758449
30.725319624 0.794025703118919
30.857051611 0.793933239388634
30.968491316 0.793956179745173
31.059983731 0.793960290030333
31.161430121 0.793952695518932
31.293067456 0.793841826740255
31.404502869 0.794109584162911
31.485867262 0.794001611011888
31.556961775 0.794060978165704
31.688680172 0.79389681284292
31.759954215 0.794070855791394
31.871440411 0.794087980173433
31.96274829 0.793998549557065
32.064164639 0.793955923245947
32.155466319 0.79397987561577
32.226540328 0.793990419582809
32.358250142 0.79402479152844
32.459643603 0.793990089570057
32.591341734 0.793968661285396
32.698976756 0.793972364687743
};
\addplot [thick, peru22113282]
table {%
0.101952315 0.684053968205348
0.213447094 0.684019523100138
0.314730644 0.683972671164154
0.405979157 0.682851041847773
0.557744503 0.679933510862774
0.679555416 0.676864758124005
0.790926218 0.670651229058298
0.872049093 0.664472006285508
0.983428478 0.659786023477305
1.10480237 0.653316508045525
1.226388216 0.648048402890775
1.348046541 0.641998102854858
1.540369272 0.637174916471713
1.651736498 0.635059953662022
1.742921829 0.633975704991383
1.854310274 0.634839756128479
1.975747109 0.633885732689387
2.087200642 0.633498072300233
2.178506375 0.634052964517203
2.279862642 0.635435731199291
2.411485434 0.637256207893878
2.522880554 0.636270005134423
2.604038 0.640403464075734
2.725707531 0.641231677315326
2.837139368 0.646053824333172
2.958860159 0.646748213295599
3.07029891 0.651676131088337
3.19182992 0.654425322911111
3.303182602 0.657973298927377
3.424694777 0.665918593045226
3.546261072 0.667462902878563
3.657644034 0.673204263046912
3.769156456 0.676057660595887
3.89084816 0.682369757059253
3.972077608 0.688729141548738
4.083470345 0.690716620587243
4.205035687 0.69591598652913
4.316534281 0.702790931940554
4.407878399 0.707243098411612
4.519314528 0.713737623433828
4.640799523 0.719165212725808
4.752158165 0.725118009714359
4.873608589 0.73030236973855
4.98494339 0.736528543683519
5.076248884 0.743621429217379
5.187665463 0.749383131371521
5.309334517 0.757409532939134
5.410673857 0.762897044956238
5.542256832 0.767372411809713
5.653637886 0.775677845816448
5.744893551 0.783531754188282
5.846354008 0.78804068230069
5.937595844 0.793978374897896
6.049191475 0.801340439203221
6.140560389 0.807606617473192
6.252058268 0.814247319946706
6.373514652 0.819502814086517
6.485013962 0.828748748629885
6.606626034 0.834582243319056
6.718142033 0.842516585366444
6.809596777 0.851565323977131
6.910939694 0.856179703357755
7.04274416 0.861400218073252
7.154142857 0.872815881945511
7.245391369 0.877573341505759
7.346708298 0.884089693463953
7.468267679 0.891035993859791
7.579763651 0.89702663709886
7.701187849 0.901868418285732
7.812522888 0.911163606105876
7.903736592 0.915092826516135
8.015126467 0.922643857036356
8.10641408 0.92810123129781
8.217954159 0.935254707413561
8.339756489 0.942579214241189
8.45108819 0.95005303486832
8.572600603 0.956377820149278
8.683979988 0.966398646703681
8.775252104 0.971160418336823
8.886629343 0.978974636252549
8.977809429 0.986871165977988
9.089183569 0.991760353960334
9.210597754 0.999928130722298
9.311884642 1.00601521588832
9.44341898 1.01192428591228
9.554780245 1.02127419386185
9.646026135 1.02840249541854
9.747326613 1.03685155239687
9.868829727 1.04031773768953
9.980262757 1.047517550206
10.111845494 1.05587499557456
10.213099003 1.06243123838879
10.334564448 1.06839240024722
10.445906639 1.07883517692313
10.53707695 1.08610140415905
10.648485422 1.0933660069618
10.770169974 1.1013078123302
10.881709576 1.10784892909509
11.003253699 1.11206789246465
11.114842892 1.12435750612532
11.236521006 1.13108123356139
11.34815836 1.13943034066068
11.469840527 1.14426397975992
11.591517687 1.15476038581085
11.672723294 1.16042823615599
11.784096003 1.16700341977897
11.895643711 1.17391082065296
11.966652632 1.18025628504216
12.09819746 1.18570892238653
12.199534655 1.19356847566261
12.331106425 1.20043817199848
12.442758084 1.20917660314258
12.574366331 1.21635138529313
12.675750494 1.22245961350678
12.807257176 1.23089373027908
12.909023285 1.23872025366
13.030603647 1.24691276197352
13.142432928 1.25419417378887
13.27424717 1.2601324171159
13.385863543 1.27013546727077
13.477169752 1.28067742235049
13.689889908 1.28421507341692
13.811441422 1.30152304284946
13.933116198 1.3045187583946
14.004421234 1.31352078289418
14.136358023 1.31862526753063
14.237789631 1.32694726957632
14.369729758 1.33346598179712
14.471298456 1.34335289191232
14.603109837 1.34958569218242
14.704748869 1.35492289189247
14.836394549 1.36313666218374
14.937725544 1.3726413197006
15.06947732 1.37810602461564
15.170913458 1.38443986521224
15.302587986 1.3927224466477
15.40411973 1.40043600777341
15.535998583 1.40801572882069
15.637495518 1.41988238267649
15.769412279 1.42468299270364
15.8709054 1.43140840813612
16.00262022 1.44081806793384
16.104081869 1.44722551989223
16.235773325 1.45639008442821
16.347337008 1.46623554024996
16.469065905 1.47363976308701
16.580586434 1.47954387381101
16.671894312 1.48676226054553
16.783386708 1.49396383136724
16.904957295 1.49893009831646
17.016447544 1.50905094283157
17.138023615 1.51742049355763
17.24954176 1.5256226483796
17.371130228 1.53185909995984
17.4827137 1.53995187546362
17.573998451 1.54604085392707
17.685446978 1.55410328750925
17.807010651 1.5630285970435
17.928589344 1.56939754594217
18.040145397 1.57682253920287
18.161836624 1.58254705773193
18.273336411 1.59313056201965
18.405070543 1.59713125816694
18.516708374 1.60700261937914
18.608309746 1.61511209810119
18.71999979 1.61895233273216
18.841751337 1.62644864299031
18.953335047 1.63486596355797
19.044627667 1.64091968545886
19.145985365 1.64739145607013
19.26757288 1.65494610941857
19.379103661 1.66276598077612
19.520947933 1.67014575777497
19.61226511 1.67916655034017
19.74400878 1.68519163655328
19.845428944 1.6922996523011
19.936691523 1.7008539812552
20.04818058 1.70633742089894
20.139576435 1.71400260497237
20.251100302 1.72047471378639
20.372824669 1.72731530694265
20.48428607 1.73332110915626
20.57577014 1.7421999076834
20.687187195 1.74933733169302
20.768323422 1.75748399344104
20.879840613 1.76294593310909
20.971203327 1.76936419575781
21.082726002 1.77473626860986
21.174021244 1.78090064999511
21.285482407 1.78533683847443
21.376917839 1.79426213157953
21.488381148 1.79997159075609
21.60005498 1.80739142216104
21.671182156 1.81371946848419
21.802777529 1.81985604147615
21.904110909 1.82712883431788
22.035850525 1.83471290148813
22.137230158 1.84319820512835
22.26903224 1.84954408977223
22.370383978 1.85843836318266
22.502019167 1.86354172931295
22.603406668 1.87193958351563
22.735101938 1.87943254305798
22.836460114 1.88652722064836
22.968154669 1.89435011724887
23.069686175 1.90248908665872
23.20140028 1.90789999864437
23.30283308 1.91492271981419
23.434602738 1.92177231129671
23.54617858 1.93140296051926
23.637454272 1.93722121223276
23.748968601 1.94410514047356
23.870520115 1.95093879026606
23.992402554 1.95835158720946
24.073690891 1.96570516823839
24.185410261 1.97078713036673
24.307180643 1.97765125722581
24.418693781 1.98729877249423
24.540293932 1.99291242582816
24.651860476 1.99929563316011
24.775224686 2.01026257606701
24.886772156 2.01608436758929
24.978144169 2.02361014600329
25.079782248 2.03011917434605
25.17102623 2.03782302711564
25.282702446 2.04433370684467
25.373985291 2.06538993566561
25.485694647 2.05563668281204
25.576942444 2.06205135471562
25.739073277 2.08266548227815
25.84039259 2.08754163545729
25.941898108 2.09255073802388
26.04322958 2.10543712741455
26.164761543 2.11513342286421
26.235819102 2.1188356328136
26.367445708 2.1100697305751
26.448679686 2.11505467470076
26.570359469 2.12126450272171
26.671755076 2.15406046237083
26.803768635 2.15728678078862
26.905168772 2.17262931374701
27.036974192 2.17425722669723
27.138389826 2.15434872202759
27.270028353 2.1607687906474
27.38153553 2.18477982780373
27.513284445 2.19043880868728
27.614676953 2.19198530745818
27.705968857 2.2171746357878
27.817543984 2.22121068390079
27.939400912 2.22710031225737
28.050885201 2.21208088391331
28.172543526 2.23907658822457
28.284034252 2.24639761883492
28.375751734 2.23540949858346
28.487300158 2.25917057108543
28.609140396 2.26402074422172
28.710489989 2.25080494094344
28.84228611 2.25352177433962
28.953811884 2.26244911764999
29.0450418 2.26859964454509
29.146391869 2.27311427594419
29.268020392 2.27665650664285
29.379608393 2.2845176196963
29.471033097 2.29244714733226
29.582754612 2.29542410253114
29.704324722 2.30095601327712
29.815984726 2.30786019787875
29.937606096 2.31257763601017
30.049113751 2.31750510207628
30.140425682 2.32510786378326
30.251888275 2.32904291217397
30.373578787 2.33509487261685
30.485229016 2.34213529868779
30.576562643 2.34608213557268
30.738556147 2.35508428228416
30.83989811 2.36131375879062
30.971638918 2.36495960790713
31.089544773 2.37159806834082
};
\addplot [thick, mediumseagreen85168104]
table {%
0.081663847 0.777512823821161
0.173087835 0.777656364889375
0.244186162 0.777750211956536
0.345599889 0.777403715794038
0.43692851 0.774719466304116
0.528249502 0.771972797083271
0.609479665 0.769219402009934
0.720936536 0.763944796554625
0.812233209 0.757912373864216
0.913655519 0.752903782171058
1.004987239 0.745639311840685
1.086476087 0.736615930184401
1.177764177 0.73153384786913
1.289258718 0.726016817410791
1.410892486 0.715371081642994
1.532540559 0.707211553812314
1.603648424 0.699923483732677
1.695010662 0.693009382085132
1.776218652 0.68527056977791
1.887716054 0.677839640081029
2.009349107 0.671684755105219
2.120883226 0.664079184073454
2.212255954 0.658164382615785
2.324351549 0.662058563025957
2.405726432 0.663159108063165
2.517614126 0.655436496201347
2.609025716 0.645794507345276
2.710519552 0.645187847377055
2.842246294 0.638912708360135
2.953950405 0.629757435189584
3.045557022 0.623539433289383
3.157339096 0.615424802069039
3.279261589 0.608826399199908
3.380799293 0.602449825322273
3.5129354 0.5963478166235
3.64499402 0.588368928858713
3.746546745 0.580748029358696
3.878315448 0.575315042430598
3.96984148 0.567113270276267
4.061161756 0.559687001182012
4.152590274 0.55524659446475
4.264062166 0.548873290982808
4.345287323 0.543089517117221
4.45685029 0.53946096430149
4.548120021 0.533287476266421
4.659742116 0.528737976059635
4.741057157 0.525515066560013
4.812146186 0.520621264889484
4.933914661 0.517189923789861
5.005008697 0.511127062122485
5.086436748 0.508266643176673
5.207985639 0.504520371188485
5.279183864 0.499817789515739
5.400971412 0.497966254527546
5.482303381 0.493692357233382
5.594343185 0.490745513441802
5.69617176 0.486632209904384
5.787445545 0.482232512915944
5.878823757 0.480618517496099
5.990443944 0.475963049538758
6.071626186 0.474414175227568
6.142779827 0.472652149134734
6.274559736 0.469581078113544
6.345655918 0.466492796271169
6.477408647 0.465668791499252
6.578847169 0.462361911377566
6.710520505 0.460677274100873
6.821966409 0.456802706377491
6.913397073 0.454501302398762
7.014725446 0.453654079491018
7.146404266 0.451999714455259
7.257858991 0.448890646710752
7.369673967 0.446913570852358
7.440853834 0.447118477385114
7.532307148 0.445298421592358
7.613726854 0.444165847106886
7.71518445 0.443596508888854
7.836822986 0.442163646762237
7.948349237 0.440318078399592
8.070041894 0.438365143597042
8.141175985 0.437886657363049
8.232462406 0.437570200739946
8.303633928 0.43913165105137
8.374769926 0.436798408410473
8.466172695 0.435331381710277
8.577749252 0.435632386099861
8.699307203 0.434433827513515
8.780442476 0.433126263670177
8.902002096 0.4318145364991
8.973088502 0.431163498455179
9.054242372 0.430549490569236
9.145479917 0.429433884028678
9.257465839 0.429312587153194
9.338609933 0.42842719119202
9.409708261 0.428849116286
9.541328191 0.427469056072379
9.612385034 0.42668507733713
9.733939647 0.428011517743829
9.805083274 0.428169106790765
9.886420488 0.426218127231601
10.008035659 0.423635130063648
10.079107046 0.425747433461652
10.200747013 0.425848955551298
10.271861791 0.428331879032337
10.353062629 0.428235113793785
10.484696626 0.422474961242477
10.586510181 0.420397068782926
10.678186416 0.426741729335975
10.78979969 0.424288050117634
10.911387443 0.425967148395737
11.022881984 0.426747858434352
11.114194154 0.423907083675709
11.215582132 0.423716392709401
11.337193489 0.424214023132046
11.408247947 0.423237860635205
11.539863586 0.423841552130532
11.611192941 0.425445793530184
11.732945919 0.425013211147417
11.804131746 0.423467961627322
11.875226736 0.423257503050268
12.007091283 0.423045413176559
12.108451604 0.422148111773875
12.240246772 0.422692433204891
12.311374187 0.421264674967879
12.433039426 0.421789590864232
12.514198541 0.423569698499053
12.615614891 0.423249943791173
12.706956625 0.422486406216551
12.818466901 0.420447568619371
12.940029382 0.418740314132024
13.011140346 0.422909268971414
13.142867088 0.421314525223996
13.254360914 0.419797583756888
13.345685005 0.420941944419567
13.447069406 0.42080842862364
13.538327217 0.422822139099273
13.609386682 0.421814063177829
13.740981578 0.421775224929132
13.81211543 0.421153741961334
13.933830738 0.421497069975676
14.015113115 0.421324904855111
14.136859893 0.421403258405948
14.238295793 0.421946745703467
14.329644441 0.421854252017222
14.410839319 0.420913987006919
14.54275155 0.420689295431903
14.654246807 0.421781256000937
14.776110887 0.421889618735122
14.887564182 0.42109679860726
14.978893518 0.420729227845074
15.090397834 0.420365935460715
15.171568155 0.419505509809843
15.283143997 0.418003889427911
15.374453306 0.422887836044307
15.486297369 0.419278709766082
15.607971906 0.417670772855948
15.679084539 0.419524378914139
15.810775756 0.42203226492689
15.922182321 0.419960973835992
16.013883829 0.419046293169013
16.11537671 0.4201377226717
16.236940383 0.420565915813864
16.308013677 0.421361875781815
16.399306058 0.422020083877331
16.510751485 0.422051697594532
16.642391443 0.422437369311599
16.743811845 0.420463885179233
16.87558031 0.420778995668162
16.97704935 0.420396736319203
17.108750343 0.420198891791446
17.220205307 0.419483429016376
17.311585903 0.418351551284671
17.413015604 0.418618697556194
17.544755935 0.421514320697581
17.64769721 0.418168604634968
17.749071359 0.41654434980566
17.850570678 0.420115261908815
17.941783905 0.422126111478645
18.053303957 0.420782717746065
18.174815893 0.419886050066548
18.286334037 0.419636265154324
18.38767004 0.419652939077367
18.53946805 0.41923619445881
18.610459089 0.421916777876667
18.74198842 0.422362648034034
18.883701562 0.421132557137177
18.985045433 0.420654518711992
19.076482772 0.421567452426007
19.187861681 0.421935647639108
19.279083967 0.421496931479937
19.390497684 0.420735606512484
19.511979579 0.420400104352715
19.613331556 0.420449140799874
19.714646339 0.420366289751162
19.815959215 0.419476201942375
19.93743658 0.418401567164633
20.008459091 0.418848846507438
20.140137195 0.423491079495009
20.211255073 0.417829990732141
20.332882642 0.416788603937323
20.403898716 0.420049877136719
20.485040664 0.42126935559966
20.606550693 0.42123744020161
20.677499532 0.41943321294104
20.809051275 0.419291708240681
20.920516967 0.420268608042391
21.011676311 0.419642929856516
21.12304449 0.421074949450522
21.204129219 0.422348773036933
21.275251627 0.42174341237751
21.407036066 0.420771336851889
21.478111028 0.420384902088023
21.609766483 0.419860963617611
21.721271276 0.41981883227158
21.842766761 0.419867971512753
21.954199075 0.419276598047182
22.075871229 0.418410889585257
22.187363147 0.420575096567401
22.278601884 0.422746521207832
22.390072822 0.417867970628597
22.511660337 0.419165807101553
22.623050212 0.420330218374619
22.704107999 0.420816430772061
22.775115966 0.419514826598585
22.907323837 0.418886217739943
22.978577613 0.419665965996319
23.090260267 0.419869108950207
23.171508073 0.420767807375224
23.252949714 0.421806100055989
23.34425807 0.421971390959393
23.44567275 0.421194219648454
23.547261953 0.420219860721209
23.648715257 0.420513539653249
23.740028381 0.420847230683266
23.811126947 0.420126866738273
23.942831039 0.419258884936989
24.044205188 0.419211587837224
24.175924062 0.417901194922463
24.297589302 0.419791301084527
24.378769159 0.422898698503825
24.490298509 0.41666386076246
24.571501731 0.41869183233768
24.642576932 0.419738788931962
24.774234294 0.421256906076287
24.84533596 0.419856320747569
24.977146625 0.420096678613646
25.088767051 0.420083813723353
25.169993639 0.419266992974396
25.251247167 0.41950531827021
25.342578172 0.419934220982428
25.454084157 0.421155733936138
25.545354843 0.422016935131178
25.657043218 0.421736562249517
25.738306283 0.421071279365634
25.809441089 0.42023200780749
25.940976858 0.420533637813639
26.00191164 0.420533603777547
26.143800735 0.419728115285519
26.24523735 0.419343971972783
26.346842527 0.418265723377814
26.448229312 0.417766796255704
26.569825649 0.422065354276541
26.640983343 0.417436810819277
26.732337474 0.416045761161812
26.844130992 0.419060648857429
26.975961446 0.421831401202389
27.087461948 0.419471077650094
27.178813219 0.419171418125052
27.290315628 0.42026839791054
27.371504783 0.419566089092597
27.452922582 0.420780658364332
27.574498653 0.421827237273275
27.645671606 0.421311409585369
27.747145652 0.420683354200463
27.848711967 0.420316262185668
27.941245317 0.420478537617685
28.012465715 0.420558630560669
28.144241809 0.420276847692948
28.255879402 0.419376139670187
28.34841752 0.418210418305947
28.46002841 0.420191452067054
28.541233778 0.421531416964188
28.612294673 0.422214545239235
28.733858585 0.416471693763814
28.845358848 0.419127711220611
28.977134943 0.420988602657554
29.088538885 0.41938432154652
29.179793834 0.418725524018607
29.291265964 0.419913386582332
29.372389555 0.419921148096184
29.443581104 0.420479132616304
29.575277328 0.421503753603119
29.686877727 0.421804016213453
29.778238773 0.420799189300585
29.899867296 0.420637643896519
29.981110096 0.420860834121883
30.102875471 0.420752056897552
30.173953533 0.419652242547525
30.265245676 0.419209619590795
30.346430063 0.41801701501229
30.457888603 0.417949529585716
30.549142599 0.421603500781297
30.650566816 0.418177893290588
30.741891145 0.417158332813851
30.853472232 0.420647181941896
30.944886922 0.420674130836328
31.056631088 0.420283818563817
31.137984275 0.419897214216882
31.249533414 0.419213298118121
31.371124506 0.41941939690362
31.442279815 0.419958394543199
31.543643713 0.420801017768064
31.655242681 0.421284392883716
31.746577501 0.421970142796926
31.848031759 0.420986188148111
31.939414501 0.420649216993922
32.010578393 0.420431131665979
32.142321586 0.42088812983952
32.253942489 0.419711584669195
32.34525156 0.419374166284635
32.4567914 0.418086456966646
32.578382015 0.42158952161301
32.730268478 0.422317781279489
32.811552047 0.41693871724741
32.973752737 0.419494187811877
33.055013179 0.418753923854195
33.176716566 0.419223865733793
33.288173675 0.419970915233225
33.379625797 0.419034120397596
33.48101449 0.419274124023578
33.572322607 0.420051530988023
33.694116354 0.420926186036906
33.775288581 0.421799977444314
33.897004604 0.421394545897988
33.978224992 0.420964632823512
34.089728355 0.420310153313739
34.170989275 0.420641908224771
34.252179384 0.42040032737094
34.343450784 0.419367225409208
34.444796085 0.419067379354694
34.536279439 0.418108981570281
34.607371568 0.419077496790703
34.698619842 0.421280390515356
34.810149908 0.418406083511442
34.951926708 0.416557995554393
35.093849659 0.4198739813118
35.23562932 0.420317698197779
35.316878795 0.420028592320724
35.408160448 0.419354722538183
35.469218015 0.419637898410543
35.570542335 0.419364497402161
35.712313175 0.420686047199164
35.854065895 0.421650075574428
35.975645065 0.422470002345181
36.117443561 0.420860287540644
36.249145507 0.421812469430267
36.370728015 0.42080998105054
36.441803216 0.420545090806685
36.543138742 0.421020060388652
36.654726743 0.420558234674439
36.776477575 0.41939482150113
36.908194541 0.418000582846735
37.009634256 0.417569403477607
37.14129424 0.418905685283118
37.28327012 0.422619510058979
37.384804248 0.417576915430433
37.506679296 0.416718691959642
37.608131408 0.421498691091416
37.750157117 0.420633944007266
37.841473341 0.420221842594215
37.942889213 0.420191513793806
38.054419279 0.418827778296899
38.146031618 0.419079620829365
38.267792463 0.421017947193449
38.36967349 0.421090842575972
38.51155281 0.420477755285474
38.623064279 0.420634413844013
38.744787931 0.42072059379821
38.888643026 0.42016019650553
};
\end{axis}

\end{tikzpicture}

%% file: tex/realworld/e3p2.tex
% This file was created with tikzplotlib v0.10.1.
\begin{tikzpicture}

% \definecolor{darkslategray38}{RGB}{38,38,38}
% \definecolor{lavender234234242}{RGB}{234,234,242}
% \definecolor{mediumseagreen85168104}{RGB}{85,168,104}
% \definecolor{peru22113282}{RGB}{221,132,82}
% \definecolor{steelblue76114176}{RGB}{76,114,176}

% \begin{axis}[
%     width=\figurewidth,
%     height=\figureheight,
% axis background/.style={fill=lavender234234242},
% axis line style={white},
% tick align=outside,
% x grid style={white},
% major tick length=2.0,
% xmajorgrids,
% xmajorticks=true,
% % xmin=-2.24838079205, xmax=49.03273762505,
% xmin=-1, xmax=41,
% xtick style={color=darkslategray38},
% y grid style={white},
% % ylabel=\textcolor{darkslategray38}{Dist. to target (m)},
% xlabel=\textcolor{darkslategray38}{Time (s)},
% yticklabels={},
% ymajorgrids,
% ymajorticks=true,
% % ymin=0.0298281469811177, ymax=3.06259763603752,
% ymin=-0.1, ymax=3.1,
% ytick style={color=darkslategray38},
% xtick distance=10,
% ytick distance=1
% ]
\definecolor{darkslategray38}{RGB}{38,38,38}
\definecolor{lightgray}{RGB}{192,192,192}
\definecolor{mediumseagreen85168104}{RGB}{85,168,104}
\definecolor{peru22113282}{RGB}{221,132,82}
\definecolor{steelblue76114176}{RGB}{76,114,176}
\begin{axis}[
    width=\figurewidth,
    height=\figureheight,
    axis background/.style={fill=white},
    axis line style={color=lightgray, line width=0.5pt},
    tick align=outside,
    x grid style={color=lightgray, opacity=0.3},
    y grid style={color=lightgray, opacity=0.3},
    major tick length=2.0,
    xmajorgrids,
    xmajorticks=true,
    xmin=-1, xmax=41,
    xtick style={color=lightgray},
    % xticklabels={},
    % ylabel=\textcolor{darkslategray38}{Dist. to target (m)},
    xlabel=\textcolor{darkslategray38}{Time (s)},
    ymajorgrids,
    ymajorticks=true,
    ymin=-0.1, ymax=3.1,
    ytick style={color=lightgray},
    xtick distance=10,
    ytick distance=1,
    % title=TARGET 1,
    % Scientific notation for y-axis if needed
    scaled y ticks=false,
    yticklabel style={
        /pgf/number format/fixed,
        /pgf/number format/precision=1
    },
]
\addplot [thick, steelblue76114176]
table {%
0.102745771 2.85159668811436
0.19490385 2.85156851848396
0.26617527 2.85157480179257
0.367589235 2.85143225269382
0.469255686 2.84819219051138
0.601083755 2.84158524719126
0.69260025 2.82721527872895
0.794291973 2.81545915816612
0.896020889 2.79599333498889
1.027839422 2.77544084094007
1.099061966 2.74769225120687
1.230915069 2.7307658772333
1.373281002 2.68725952817103
1.474683046 2.65972485668857
1.586467981 2.63357584568101
1.667723894 2.60741893755353
1.779306412 2.58436893133726
1.86051178 2.55751411558729
1.931577921 2.53708942985475
2.063210249 2.51858049027461
2.174723625 2.48593323503745
2.296346903 2.45557665959841
2.407807588 2.42614897075745
2.499067068 2.39677498090336
2.610692501 2.3723636609306
2.701928615 2.34421669352579
2.803369999 2.31777506765025
2.925058126 2.29130081795029
2.996139765 2.26230051124555
3.127801895 2.24029080833318
3.188750982 2.20620030474982
3.290106535 2.18741341174344
3.361191034 2.1636069867259
3.493016004 2.13768003054674
3.564055681 2.10818231998034
3.685863018 2.08750549711822
3.797312736 2.05134204193715
3.928919792 2.02275043995039
3.989875078 1.98938163741969
4.081120491 1.97472284539744
4.16242671 1.94982919521432
4.274133444 1.92724501125264
4.396007061 1.89564503699179
4.507416248 1.86674310543457
4.629024982 1.83200415684706
4.750603437 1.80114764006009
4.831755638 1.76851125254113
4.933336019 1.74648529443556
5.06498456 1.71952968159528
5.176439523 1.68199785429105
5.267787695 1.6519916929219
5.379323721 1.62674692686718
5.460472822 1.59883752578842
5.53162384 1.57386283464059
5.653317928 1.55316265173182
5.744682312 1.52278871723866
5.846049785 1.49678091794063
5.92718625 1.46701804425567
5.998384237 1.44419863374868
6.130166769 1.42459873124981
6.231550216 1.3870105620516
6.363390922 1.3604127011692
6.4749825 1.3119398409403
6.566339493 1.28681732528298
6.677953958 1.26281932634424
6.799631595 1.22970540736223
6.901062488 1.19424328316751
7.002699137 1.16546823686967
7.104147196 1.13372499126137
7.22582221 1.09943355486831
7.296828747 1.07393901495659
7.42845726 1.05079309686009
7.499602556 1.01489155476139
7.621220112 0.995013047595006
7.692292213 0.963059494557754
7.773533106 0.941179129270019
7.864804268 0.918217162849705
7.976327657 0.893904112099632
8.067643881 0.86176208901202
8.179279089 0.834277655690651
8.260458231 0.804439292983413
8.331570864 0.785183806844332
8.463287115 0.762940623489798
8.58484149 0.726793869272327
8.666311502 0.695021248500086
8.77779293 0.678884595651799
8.869231939 0.652907105915554
8.98076415 0.630762553536027
9.062021971 0.607214455521534
9.133130073 0.589417597264341
9.264813423 0.57228853761322
9.376323223 0.545393343690157
9.46754241 0.525143739627214
9.579034328 0.506515171327505
9.660137653 0.487461830302552
9.73126769 0.472093365918721
9.862946987 0.460118103175796
9.974483013 0.440511291814671
10.065807581 0.425127288942903
10.1772933 0.413090302879491
10.258576631 0.404516856911938
10.329664945 0.398403865863512
10.461303711 0.393034675094505
10.532381773 0.388769275904389
10.66405034 0.386147498752018
10.775635242 0.383228042179056
10.866980076 0.369182326156143
10.96841979 0.358492640777255
11.090052604 0.371135204156441
11.161134243 0.370291374232188
11.242309093 0.355161173115627
11.33356142 0.351414959185185
11.43495655 0.369545147110695
11.55666995 0.369654151973568
11.627847194 0.361866403850664
11.759845972 0.365407193874354
11.831135273 0.373631788837476
11.952924013 0.377345474909962
12.054479122 0.382796790433471
12.196504831 0.390087097991815
12.308006286 0.433167191454796
12.429591656 0.37074434203664
12.54111743 0.399665880183681
12.66271472 0.41915357777841
12.76422596 0.380809072354628
12.895997286 0.423992685419971
12.997369051 0.404310151147844
13.129064321 0.428902687232612
13.331472158 0.432257410740927
13.432850361 0.434134524302339
13.564620971 0.425307711177922
13.676320553 0.393785985275514
13.797876358 0.436752838557035
13.909413576 0.385532209564141
14.031074047 0.418008985325589
14.142601967 0.427997280238074
14.233957052 0.387051011682689
14.385998249 0.389012919933754
14.497719288 0.381177422902963
14.619355202 0.441782040135905
14.730751514 0.390896879727506
14.862404585 0.364908679843897
14.963737964 0.376114974392287
15.095319986 0.3810100398317
15.166389227 0.363911079190307
15.277829408 0.367698207731138
15.359233618 0.402745159872579
15.430271387 0.377264765668956
15.561891556 0.34414272894197
15.632966518 0.352229710091078
15.76468277 0.41416467960922
15.866147518 0.360930701603916
15.997812986 0.366784864923157
16.109371424 0.3649267590003
16.231002092 0.351582838797837
16.33240366 0.353889522153356
16.46402502 0.363351459566329
16.575473785 0.357211615426365
16.707148075 0.357730693322652
16.808502435 0.404502613329579
16.899786234 0.403229614357484
17.011264801 0.358828240802475
17.092525005 0.361615265918649
17.16372323 0.361470805769773
17.295591116 0.360283242487785
17.407133817 0.363826759565107
17.528674602 0.353948737822496
17.589610815 0.342567452133272
17.731349945 0.343114766123697
17.832760334 0.383151422198658
17.964627266 0.343935240355615
18.07635498 0.34723787214906
18.167640447 0.347830535580459
18.26916933 0.389828165087304
18.360689163 0.348127852549568
18.43174839 0.348932195789431
18.563552141 0.350316624483822
18.675071478 0.351244021721279
18.766410827 0.351490926891756
18.87787342 0.352145994208127
18.959061861 0.353295707214009
19.030155182 0.353302788297124
19.161821127 0.353184535025554
19.232932567 0.355289737705827
19.364737987 0.355972898184833
19.476225137 0.356613730879821
19.567471742 0.356868774067874
19.679044008 0.357017429116256
19.760219097 0.357913647930837
19.831257105 0.358533283185902
19.962836981 0.359595775463095
20.074306965 0.359849953943019
20.195993423 0.360184332199475
20.307408094 0.361931780959148
20.398686409 0.364390949512875
20.510139465 0.364442829426189
20.601433515 0.364903688500012
20.702780247 0.367182072294054
20.80414772 0.368045922292401
20.905473947 0.369565601418056
20.996737242 0.371980611473826
21.108206272 0.3739351735227
21.209622144 0.373394756400633
21.310923815 0.375030548800911
21.402272224 0.377521648523454
21.503650427 0.379470580432554
21.625313759 0.381353949149672
21.696642399 0.383836355519884
21.828480482 0.383945888758303
21.88947773 0.387728357334799
22.021118641 0.389475882238527
22.092168569 0.39110917660415
22.173393011 0.392728218137101
22.264601946 0.395351692234883
22.37632966 0.396725754498428
22.457550049 0.398658911163319
22.528654098 0.40079377558937
22.619952917 0.403655302650056
22.691092968 0.404515923686628
22.772285223 0.40499905745892
22.863582134 0.40694562167786
22.975011587 0.409994947159267
23.066312074 0.409657954884056
23.178055048 0.412285455739206
23.269271374 0.416116934018254
23.370634079 0.417353627154113
23.461847782 0.41824968710049
23.532976389 0.41974480930457
23.664668083 0.421432341817449
23.776323557 0.423725061217072
23.857400179 0.426487966684219
23.928416967 0.428448438020818
24.019748688 0.430356660444202
24.090846061 0.430717129573162
24.171997547 0.43099333182392
24.293562889 0.434007014896948
24.405014992 0.435616575025545
24.496301174 0.437358777635471
24.608023405 0.440374142541791
24.729839563 0.442353317492653
24.841385841 0.444920668900802
24.96304965 0.447998446859324
25.074500799 0.450949316651896
25.196050167 0.452991087901198
25.307532549 0.455380380331131
25.429274797 0.458582084486976
25.540712833 0.460687746318529
25.662282705 0.463668287992486
25.773728609 0.463978316368404
25.864985227 0.468282158436295
25.986500978 0.46994143658813
26.097901344 0.473807910016023
26.229585647 0.475997862482697
26.330980062 0.478647981889403
26.462713957 0.479129691402271
26.574246406 0.482148738265904
26.665705204 0.48505357297109
26.777209043 0.486607519015853
26.868472099 0.487699430594855
26.969912767 0.489422402543675
27.061241388 0.492360948282466
27.132371425 0.492538153651825
27.264237165 0.494955489891647
27.375724792 0.497138268808943
27.467069149 0.535533924793734
27.568529367 0.526511531315634
27.659767151 0.501232672026111
27.730894327 0.502648521097587
27.86288476 0.505844605619518
27.974463224 0.50783550301735
28.065962314 0.505298149508976
28.177614689 0.510245539726613
28.299327135 0.505322311312433
28.400709629 0.504068494233886
28.491954803 0.509159094703959
28.563047886 0.509980440009389
28.694783449 0.510630277274571
28.806476831 0.511805058312458
28.928232908 0.514256906220148
28.999311209 0.514243458775938
29.131119728 0.513775915993262
29.242784023 0.51610593053141
29.334271431 0.515896547001394
29.4457376 0.517915485175679
29.526989221 0.520049650793095
29.598104 0.521748401999964
29.729819536 0.522787884551627
29.841309786 0.52695727865144
29.932637691 0.52848916328821
30.04420805 0.529370402482879
30.135881662 0.529028896473365
30.237236738 0.529280821024806
30.35874629 0.534623888573921
30.429846287 0.534872013173032
30.561509847 0.531767861270463
30.632615328 0.535381932970541
30.764439821 0.53771123757249
30.87595129 0.536326861800681
30.967375278 0.538552577486055
31.089046955 0.539732589536803
31.160146475 0.541696482132234
31.261569023 0.542724957965596
31.36302638 0.545102225913988
31.494740486 0.546232984192163
31.606338739 0.574750984888759
31.697613716 0.548943540705337
31.809112072 0.562651489804815
31.900495529 0.542476409708472
32.002014398 0.542380466141137
32.133822918 0.547075929644279
32.235210657 0.542640188765797
32.356757879 0.547873935421969
32.427806377 0.547879957225507
32.519305229 0.545007507676187
32.590417146 0.547430829009436
32.681679964 0.550599764759748
32.762864828 0.548925809017936
32.884572506 0.549995357370052
32.965838194 0.553574792072093
33.087469578 0.551673844648179
33.158572912 0.55348806670642
33.239832163 0.556174633245531
33.36148119 0.555274843847615
33.432590723 0.553238151798599
33.564382553 0.55662705844668
33.676024437 0.552511833309488
33.767347574 0.551945570909064
33.868909597 0.55707395453982
33.960151672 0.551364956396766
34.031244039 0.550744093538345
34.163060665 0.550043268909491
34.274610042 0.549043857322836
34.396190166 0.550945130065071
34.507656574 0.553111241757217
34.599081039 0.553291952828672
34.700471401 0.553700823201448
34.832212448 0.555475807757631
34.933770895 0.556252466024895
35.0353055 0.557621455485201
35.136766195 0.558620146024372
35.238192558 0.558835517151423
35.339676142 0.559543031263891
35.43099451 0.561688301524853
35.542518616 0.563311527609064
35.633911848 0.597630926157765
35.745501518 0.599388480840155
35.826680898 0.56754822560502
35.897792578 0.574021513969898
36.029418707 0.594469559580978
36.140930652 0.55901576901192
36.232214212 0.594531388022432
36.333721638 0.561646371617899
36.465441227 0.587957577507679
36.576962471 0.55334471758214
36.668273687 0.560512449721886
36.769703865 0.558672415691412
36.86097455 0.560750610703396
36.932049989 0.562289982280587
37.06391263 0.561086840858642
37.175423383 0.564114104166295
37.25669527 0.565118487151579
37.327784061 0.563654759843009
37.419234752 0.563938168395494
37.490329742 0.566458590644568
37.571498632 0.567496922876556
37.672895431 0.568080517278244
37.774443626 0.568624161931525
37.865743875 0.5711249057661
37.977243423 0.571979520447252
38.09886527 0.57287968681632
38.210367679 0.574644108936419
38.321862697 0.575853934692001
38.39330244 0.57765650433179
38.484731912 0.57841355742825
38.566300869 0.577062861705381
38.677925825 0.574651870193885
38.759286642 0.612577074977906
38.83039093 0.579206697438329
38.962198019 0.57760156697385
39.033294439 0.61094039059185
39.165040016 0.578020485175137
39.276539802 0.576062093514551
39.367815017 0.578969967415424
39.479278803 0.609158540330618
39.560448646 0.610055694227053
39.631561756 0.613105793343512
39.763389587 0.57417012340848
39.874848842 0.603870617627177
39.966180086 0.576063571125555
40.077670336 0.613318762630588
40.199238062 0.603065988926834
40.300605535 0.570282275998824
40.391889572 0.57800183748843
40.463139057 0.604576761366013
40.59495902 0.573950086795394
40.666192531 0.574876509801582
40.797886848 0.575434580204177
40.919727564 0.577317214063874
41.001052141 0.565305951932381
41.112670898 0.566863749214618
41.193935871 0.572478752954286
41.285223007 0.56988240470785
41.396708727 0.567833635796318
41.528360128 0.568850026452183
41.589417457 0.583057385393175
41.690871715 0.572382982751589
41.762017965 0.602398844123632
41.83313179 0.568483572617178
41.964869737 0.568286140044748
42.076448202 0.573257693904879
42.191873789 0.605019696550814
};
\addplot [thick, peru22113282]
table {%
0.092980385 2.92319990021594
0.184582234 2.92474447744404
0.255721093 2.92353392408191
0.357144356 2.92455149093734
0.45851779 2.9158214354142
0.590339184 2.91037947419493
0.701810837 2.8983396462712
0.783243895 2.8860623137867
0.854324103 2.87461448149013
0.976183892 2.86354217763252
1.107966185 2.83493261277714
1.219624996 2.79737402805097
1.351437331 2.78856760881199
1.422768116 2.75614086345419
1.554582835 2.73911511485796
1.666085482 2.70634030905966
1.787890196 2.67922799907729
1.899576903 2.65135976173963
1.991168499 2.63330920955772
2.092963219 2.60865406794562
2.184338332 2.58524014276464
2.255503178 2.56205201543152
2.387427569 2.54444617915856
2.488772393 2.51535725992149
2.620414734 2.49097217356623
2.721782685 2.45739262541922
2.853424073 2.43477094062898
2.964891434 2.40368752696866
3.056119919 2.37723952114772
3.157527685 2.35469631475428
3.289213181 2.32940478129092
3.40069151 2.29993810678436
3.522246123 2.27615430588955
3.623779774 2.24501031212116
3.755394936 2.22447102739403
3.866967917 2.18665449654296
3.948162794 2.16104984666762
4.019289494 2.14312085000501
4.150938511 2.12604464836912
4.222046137 2.09656142849718
4.343750954 2.07912441339207
4.414847613 2.04903011545042
4.506184578 2.03458999597311
4.587599039 2.01035899579932
4.699173212 1.98950888750294
4.790501833 1.96577916202585
4.902067185 1.94271021833697
5.023645163 1.91494071826599
5.12501359 1.88585855573664
5.21629858 1.86293549187228
5.327783108 1.83793529677004
5.419099093 1.80851865143305
5.490190268 1.78845601796984
5.621868134 1.77194168856712
5.723225355 1.73908881615858
5.844790459 1.71449892692435
5.926194907 1.68236184184996
6.047819138 1.66102546215125
6.149126292 1.62859957566344
6.2403934 1.60576194167991
6.321521044 1.58052826548506
6.432967425 1.56175612254848
6.524249077 1.53237200611271
6.625635147 1.51028402077152
6.727002621 1.47998993002937
6.828351021 1.45375394989628
6.919603348 1.42892496829618
7.031032563 1.40484900965436
7.152556897 1.37748402463733
7.264131785 1.34420713704506
7.38587451 1.31800794561815
7.497483015 1.28137780929042
7.619114399 1.25386754868329
7.730582714 1.2254293169358
7.822084427 1.19463650105506
7.933683157 1.17359226580327
8.055265189 1.14475999270673
8.156734228 1.11442173643795
8.288363457 1.08640670033324
8.399836302 1.05372329253467
8.521376372 1.02385931405623
8.632822275 0.991233910884
8.724080324 0.966297410474727
8.835532189 0.943047756429393
8.916790247 0.914577325792556
9.028280974 0.891849515570676
9.119594336 0.861948745028073
9.231062651 0.837176488502492
9.322307349 0.809656497905736
9.434053898 0.784778272196511
9.525358677 0.758431206532453
9.636888266 0.74196907323743
9.72812438 0.710871939743284
9.840060711 0.69170917411319
9.921624899 0.659771583723985
10.033501149 0.641046979537482
10.124877692 0.610430539374749
10.236393214 0.586593939238865
10.317549706 0.558409553223933
10.429004669 0.538346799641983
10.530395508 0.510980451665317
10.631762028 0.487006877388059
10.723166228 0.459612074916332
10.834633828 0.438206962700712
10.926115275 0.412647858889867
11.027528048 0.387662771828154
11.118815422 0.363721001514399
11.230366707 0.342750048197716
11.321713686 0.320307584259654
11.423221827 0.298303531196859
11.524534226 0.280572482142303
11.656226635 0.257346051285734
11.767710448 0.231959103695127
11.848891259 0.212334941228213
11.920040608 0.203056801274975
12.051787615 0.19297027530739
12.122900248 0.180864060624993
12.254633665 0.175604868247653
12.356024027 0.169675377057931
12.487710238 0.168114743104802
12.599168063 0.16768130557459
12.69051075 0.172107916428289
12.802178145 0.177491114660675
12.893543959 0.187207497170919
13.00525117 0.191751069365867
13.096529961 0.20096942944309
13.19786191 0.21390300201079
13.319533825 0.225360577789285
13.430957079 0.235661058249247
13.52219677 0.250093209079869
13.633793593 0.26310364299823
13.755383492 0.2735419990955
13.867011547 0.288686990462449
13.988622427 0.300995463074354
14.100058079 0.31639224022742
14.221598864 0.32801782469213
14.33331728 0.340081416110218
14.424627781 0.350018162196414
14.576688767 0.35903457082353
14.698261261 0.376573032618297
14.820001602 0.382029248565575
14.951646328 0.395545717013393
15.022753954 0.405079297172359
15.144314766 0.414568449952571
15.225564242 0.423182804510687
15.326937676 0.432245900033558
15.458617449 0.438077860627916
15.559975386 0.453106760601364
15.691643954 0.462284644932691
15.793040514 0.471318197253764
15.914600134 0.473839279529439
15.985707999 0.483904981503877
16.117412806 0.505656822918311
16.228889466 0.502877637559643
16.320162058 0.510227770727127
16.431672097 0.513489350804458
16.553370714 0.525041245582141
16.664859295 0.526768297413686
16.786462069 0.536790471682749
16.897998572 0.542584760745451
16.989372969 0.550205701853877
17.100826979 0.554107951811255
17.19205904 0.561168425423385
17.303506136 0.568753678844976
17.384643555 0.571254577989117
17.45579505 0.578243303700933
17.587697745 0.580176523569905
17.689103604 0.591696016977202
17.8208282 0.593292737832308
17.932394743 0.598602831891934
18.023975134 0.603008072702835
18.135627508 0.610925081106648
18.217203617 0.615286423037446
18.328743697 0.620911405176509
18.450330258 0.623859589216634
18.521413088 0.627759219665539
18.653150797 0.631026034827346
18.754521132 0.636978160191943
18.886254311 0.641423476110647
18.98776269 0.6476107390289
19.11962533 0.654112656430582
19.231220484 0.655813927689829
19.322528839 0.658185070960132
19.42395401 0.663274436109965
19.555587054 0.66580051451115
19.667343617 0.668453479951521
19.758739472 0.677469800549554
19.86012125 0.679816970331987
19.981778384 0.681527430690423
20.052989722 0.6861168203012
20.184804678 0.688386370866052
20.255993367 0.695619552862266
20.387769461 0.698172030424931
20.499267817 0.700141114268725
20.620841742 0.703540614599153
20.732377529 0.710559665694816
20.854128838 0.711160662480359
20.965592146 0.711863315917089
21.087323666 0.716177947771796
21.198890686 0.725002926798884
21.280107022 0.722682701230623
21.351223708 0.72768191460636
21.442833662 0.726944227728476
21.524024248 0.732752701071905
21.645677805 0.731803922534359
21.71682477 0.738733878232646
21.838438273 0.736200005645477
21.919657231 0.744391015013717
22.031157255 0.743379687858451
22.152757407 0.748688992621792
22.264282465 0.753844156300323
22.385865927 0.753378755224152
22.497381449 0.759416218478733
22.588676215 0.765247784295561
22.700349808 0.763952811152093
22.781519175 0.76940954980415
22.85298729 0.769674068824764
22.984699488 0.775674243730481
23.055814505 0.772955282220725
23.187498093 0.773590632686124
23.288887501 0.780512161296584
23.420838118 0.777365892402169
23.522384644 0.783096167890285
23.654268265 0.783651606550528
23.755675078 0.782513945263677
23.85719204 0.788095855849013
23.958621979 0.786739515333577
24.080402375 0.78977952932644
24.151475907 0.792560496848252
24.242915392 0.790834748391534
24.314079285 0.792592472041014
24.405466557 0.791863236322913
24.486778021 0.791668492495133
24.598263741 0.794999746374176
24.689575911 0.798369635903713
24.80114317 0.797344008149916
24.892855406 0.796414416078538
24.994248152 0.795736242248646
25.085549355 0.799348685239408
25.197096825 0.801579302941199
25.288446427 0.799616677919163
25.420177222 0.799663864948967
25.521564961 0.802600830372746
25.653339386 0.805630997747524
25.754742861 0.803013182194938
25.886587382 0.80608100337648
25.997995377 0.804939238691365
26.129667044 0.80815684685365
26.231049776 0.805974017565162
26.352712155 0.807131717278859
26.464304686 0.808975295331796
26.585994959 0.811154481139582
26.69749713 0.810795019422032
26.829236269 0.811954549247889
27.021466732 0.815156143366179
27.13298297 0.813386809630312
27.224303961 0.81391093329973
27.325833798 0.815075560638538
27.457520724 0.815355202714357
27.55892992 0.814462469730383
27.690735817 0.814536522055375
27.802288533 0.814455303753304
27.883549452 0.815454655209986
27.954729319 0.814836174366674
28.086863756 0.814576207529728
28.198464632 0.814641646530761
28.289779663 0.81534667818399
28.401305437 0.816616408156074
28.522928715 0.815862621986372
28.624377251 0.815392306681555
28.756306649 0.81705440538676
28.857915402 0.814994573848115
28.989781142 0.815856756843296
29.101348162 0.815598350780963
29.18258214 0.81571841269176
29.253675461 0.815566914606685
29.385344506 0.815869609483647
29.456650734 0.815419792066279
29.588241816 0.816694156192373
29.699874401 0.816799915628369
29.821448088 0.816374257832596
29.932970524 0.817480787195104
30.054582596 0.8168187998076
30.155980349 0.817199043490652
30.257587195 0.816762354566578
30.359164 0.816799065090618
30.470701218 0.816613952002979
30.572103739 0.816849935133664
30.65339756 0.816875105895672
30.765050888 0.816257060015219
30.886800766 0.816853062866772
30.998453141 0.817302831035292
31.120253802 0.81707295930291
31.231733561 0.816913876969957
31.333200455 0.816613963624304
31.434588194 0.817189979424368
31.525892973 0.816935541461176
31.627254725 0.816491221245714
31.748882056 0.816607348554821
31.820094347 0.816999776482444
31.951931954 0.817015405557587
32.023407698 0.816496435598576
32.155258656 0.816747734017115
32.256910086 0.8168515275233
32.388540984 0.816807816869598
32.489937783 0.816659256797765
32.621602536 0.817335950050287
32.72301364 0.81675539073587
32.844679356 0.816785155536473
32.915868998 0.817115425498962
33.007253647 0.817201721644835
33.088535548 0.816136614560785
33.200035572 0.816511534193078
33.291319609 0.816458827595946
33.393056155 0.816741418560691
33.51476717 0.816908814133737
33.585844755 0.81682064735881
33.717684508 0.816247651819158
33.829170227 0.816420426873523
33.920437575 0.817116127326788
34.03207469 0.817343150382118
34.153713703 0.816418103356093
34.265300513 0.81663472944796
34.386981488 0.8170156889699
34.498446942 0.816364510287226
34.620110274 0.816405580180279
34.731571913 0.816607387920989
34.822860003 0.816790370217992
34.924239636 0.816262065579485
35.055936337 0.815132848673961
35.157316447 0.816682909450708
35.248773098 0.816506534458688
35.319945813 0.816791221172264
35.451581955 0.816656806463587
35.522756577 0.816813941989279
35.644361735 0.816480531510939
35.725580454 0.816914389941096
35.827027798 0.817318304514389
35.938493491 0.817327962357973
36.019670248 0.816647366643393
36.131095887 0.816909284022979
36.262788296 0.81676814952555
36.36418891 0.817110259585389
36.485861779 0.81658139645269
36.5974257 0.816737900576766
36.688783408 0.816550222419924
36.800354243 0.816255649351533
36.892103672 0.816480921349978
36.993531466 0.816792277421578
37.115366459 0.81720856222394
37.18653965 0.816456288306243
37.318295002 0.816670564149104
37.429856539 0.816239050743274
37.551425934 0.816211966601934
37.663055182 0.816906333880113
37.78464365 0.816553320235348
37.855813265 0.816857459553091
37.987558842 0.816707419200106
38.088958979 0.816599313063336
38.220804453 0.816423536360901
38.32226944 0.818620007501152
38.454061032 0.81832769595287
38.555472374 0.818520688460103
38.68727088 0.818776442797666
38.788809777 0.818906895856626
38.920501232 0.818689485677771
39.021869183 0.818594032597458
39.153549433 0.819346196573214
39.25497365 0.819636074981513
39.387076617 0.819483892395495
39.49866271 0.819874117515429
39.620331526 0.818718211392529
39.731841564 0.819191042380155
39.853572846 0.819378651411635
39.965126038 0.819269140274657
40.056710959 0.819158142277537
40.158099413 0.819566580879824
40.289806366 0.819283593495251
40.401316405 0.8194662301187
40.482461453 0.818904012887726
40.553683758 0.81888348816676
40.685523272 0.819380037743092
40.756875754 0.818587054318858
40.888866425 0.818814630067204
41.0005157 0.819572901118009
41.101890564 0.819153778298667
41.193570853 0.820371555544935
41.284931422 0.819356262690947
41.356181622 0.818598933694535
41.488030672 0.81952236999398
41.599632025 0.817285610763568
41.690902472 0.818803303862111
41.79229784 0.819309703517264
41.893743515 0.818465888984511
42.005286694 0.819211129660618
42.086449385 0.819728977866346
42.198046685 0.818649084237429
42.289358139 0.818955568014373
42.400859118 0.819276505813295
42.492126942 0.81663491681557
42.593539477 0.81903975987438
42.715147734 0.818909396860578
42.786238432 0.818858411388857
42.917981386 0.818546083076459
43.019579888 0.818966978243469
43.151390076 0.819043858194671
43.22249341 0.818944110974834
43.354173899 0.818728969671166
43.465646982 0.818691643513296
43.58738184 0.81890793115153
43.698863983 0.818818786336976
43.820543289 0.81840496254865
43.932158232 0.81691975266153
44.053765774 0.81911175360312
44.165224314 0.820015758422097
44.28679967 0.819578335938616
44.398262501 0.818520013588985
44.489607334 0.819416121617448
44.601113081 0.819908833581325
44.682392359 0.818661248104634
44.753826618 0.818257802910192
44.885591507 0.816986444855373
44.956993819 0.819086846480827
45.088776827 0.819642054881534
45.19020176 0.817966703458313
45.311818123 0.817663375267036
45.38295126 0.818347566752708
45.464116812 0.81874120509317
45.555416346 0.819252463538142
45.667158127 0.81870804936998
45.75845933 0.819006447961818
45.859905958 0.819207491743515
45.951186419 0.818005796219111
46.022260428 0.818870410454506
46.153996945 0.818777553802781
46.265490294 0.81912866316257
46.35678339 0.819925881435692
46.458206654 0.818720133018171
46.590033531 0.818789480232525
46.701777697 0.818737317374413
};
\addplot [thick, mediumseagreen85168104]
table {%
0.082579136 2.85825365029932
0.153891563 2.85828146887487
0.24521184 2.85829418660637
0.316357374 2.85831503049333
0.377455711 2.85602422097804
0.448756456 2.85444769305252
0.530088425 2.8498749839961
0.62146306 2.84104685956251
0.723076343 2.83141861491686
0.844710588 2.8169215478717
0.966576099 2.7962496231865
1.047815084 2.77427052276173
1.118911266 2.75447849037984
1.250695229 2.73859737677152
1.321840763 2.70655560027879
1.453631163 2.69128786867293
1.555050135 2.66159590565264
1.686800718 2.64039094301248
1.798269272 2.60967709802364
1.920058012 2.58102392842691
2.021403313 2.55616156203533
2.153257608 2.5337851987311
2.264719486 2.50310606764525
2.3459692 2.47829491459593
2.417075396 2.45764426456288
2.508293629 2.44061470063614
2.579389572 2.42155040282279
2.660559416 2.40581628775077
2.762145519 2.3848705602781
2.863561392 2.36047614678323
2.985126495 2.33165914073867
3.096805096 2.30528520303996
3.218446493 2.28137350027055
3.330331802 2.25403595271685
3.421738148 2.22647541442546
3.533337593 2.20817154777221
3.614516973 2.18021760092457
3.685624361 2.16504590467746
3.817281008 2.14502699100368
3.918634891 2.11650394360392
4.050364971 2.09424465566877
4.161873341 2.06311100009487
4.253281593 2.03844775582905
4.36483407 2.01477839098524
4.446028948 1.99117122601535
4.517163277 1.97329006701916
4.608458996 1.95616241821644
4.679579496 1.93582198721673
4.760752439 1.91980465425226
4.85213542 1.89912072272097
4.963600635 1.88012621361204
5.054846525 1.85170585932648
5.16642499 1.8322589369445
5.247760534 1.8071128388916
5.329017639 1.7882140193917
5.420328617 1.76666617787489
5.531921863 1.74911980915808
5.623192787 1.71958242966502
5.724529743 1.70096935337862
5.846363068 1.68029613772155
5.957939625 1.6508888570513
6.07966733 1.62625602754211
6.150838375 1.60130045413401
6.242168188 1.58535717987289
6.323423386 1.56356463923645
6.424813747 1.54649590606858
6.546490669 1.5225166423489
6.617595673 1.49761759915333
6.749245882 1.478907025621
6.820348263 1.45358661031671
6.95201993 1.43726981723195
7.053479671 1.40885419712095
7.18519187 1.38692390891067
7.286701918 1.35805671311353
7.418508053 1.3364179749744
7.530154705 1.3085667187925
7.651850939 1.2822026226517
7.753314733 1.25889118885125
7.885097742 1.24071979291685
7.996702433 1.20911788742177
8.077875853 1.18245222144377
8.148979664 1.16932912993232
8.301585913 1.14857515985124
8.392873049 1.12110773872371
8.48409152 1.10157894031322
8.545062303 1.08279065108526
8.646581173 1.06815737726853
8.717984676 1.04634555448215
8.839753151 1.0309411493572
8.920996904 1.00739376754737
9.042586088 0.989334190705455
9.113651037 0.965612868405248
9.18474412 0.94899159868002
9.31657815 0.932795370232116
9.387801647 0.906252266666265
9.519532442 0.892381344013321
9.631092787 0.86527144963695
9.722439766 0.839193437638994
9.833943129 0.822260761804833
9.925226927 0.802044879159049
10.026756287 0.783294844329021
10.148423672 0.761221350869594
10.2195611 0.742179973095045
10.341196775 0.726177294045457
10.412349462 0.709929695969407
10.494220734 0.694000806602213
10.585577249 0.717821056677868
10.697242022 0.673466663122067
10.78853941 0.607735772092916
10.889998674 0.608070449473738
10.981294393 0.601629315706933
11.052793503 0.598629682955873
11.184473515 0.562727796373354
11.296005726 0.576418524949493
11.387246609 0.559158345249424
11.498640776 0.600236442518617
11.620265961 0.543393124155637
11.721668243 0.529840352292537
11.853437185 0.527880604482549
11.964982271 0.521364737500009
12.046270132 0.51778529281458
12.11727047 0.498631159429549
12.208445787 0.490599321749851
12.279573202 0.485175426734989
12.360671759 0.492559388687686
12.451939821 0.478935049601901
12.553268433 0.482575055363149
12.684840202 0.461235544180915
12.796245336 0.472546401935824
12.877310991 0.463581779110223
12.948378324 0.460801308494788
13.04975152 0.462769716832595
13.120771885 0.457527662096176
13.252478361 0.451334054685492
13.353759527 0.44570999646554
13.455091 0.446132458274061
13.55638051 0.432211054808401
13.677841902 0.428169123251486
13.789191961 0.417139340614015
13.880373478 0.409999549890843
13.951372623 0.406546667801741
14.083115578 0.401055389070883
14.15425849 0.388174859727216
14.235450506 0.395261065299882
14.316647053 0.391950694652173
14.387726307 0.388276669230241
14.519536018 0.384012533354836
14.63107276 0.379236269740411
14.752770662 0.375197000392596
14.854135275 0.367050845349638
14.985824585 0.365804263088993
15.097286701 0.361457207219027
15.218904495 0.352412799822454
15.330482244 0.344739167504284
15.421858311 0.340262399080834
15.523193836 0.341391892180076
15.63459301 0.335246708732105
15.725758552 0.329852447312955
15.816953659 0.326310324411798
15.877851486 0.320850017260356
15.969124317 0.317572685515826
16.05022788 0.31713701771385
16.111180067 0.316004054159698
16.202512741 0.315976186587473
16.293721676 0.310523800454121
16.395031214 0.309632667949457
16.48651576 0.305231405650223
16.597968578 0.300293244155132
16.689209938 0.296664782733948
16.790503263 0.292968406591381
16.891830206 0.291813095762712
16.99319005 0.292776714638616
17.084349394 0.288490266926942
17.185632944 0.290041294171807
17.317210197 0.277892485410956
17.378134251 0.274327437808613
17.45923996 0.272467617548719
17.550427675 0.276451128043012
17.611376047 0.277261757412063
17.712850332 0.27565576561092
17.783919573 0.273600596395267
17.865059853 0.268836317982617
17.946249962 0.265592532134459
18.017333269 0.259829205516525
18.098502636 0.263150942293624
18.179601192 0.26275002126816
18.250669241 0.262475700359637
18.341963291 0.262504109885857
18.413060188 0.260174019738118
18.494227886 0.256518338699605
18.585452795 0.254734544903912
18.686832428 0.250839777395551
18.818451166 0.249163977266731
18.930111647 0.246769918468949
19.051884174 0.251275104644129
19.153431177 0.249481127291902
19.285073996 0.24871072851125
19.396565437 0.245148987093517
19.487853527 0.240297424313087
19.589291334 0.234360908547135
19.720858812 0.236111346847945
19.832264185 0.232667668612131
19.913613081 0.231891706697367
19.984646082 0.231999376198589
20.116232872 0.232349367689041
20.227627754 0.234402350716414
20.318846703 0.232216117593791
20.450598478 0.231013065118883
20.521626711 0.229907563895128
20.643409252 0.228164866774727
20.714415789 0.223645894592375
20.805842876 0.222863065240439
20.886970043 0.222640268121422
20.988320351 0.222491980660281
21.089704037 0.21842416354942
21.191090584 0.221314855278139
21.312678337 0.217682920124619
21.383785248 0.214851984537151
21.515404224 0.215527952782538
21.586484432 0.214487594178488
21.697928667 0.214418723934721
21.779391289 0.214573109588916
21.901272535 0.212171393715913
21.982545614 0.215285477339425
22.063844442 0.219789981284364
22.185524702 0.221610958490543
22.297021151 0.21552136057296
22.378168344 0.215550463314454
22.449243784 0.218082381711673
22.530382395 0.213394944944809
22.621685982 0.21116144229062
22.723009586 0.216732090388186
22.854652166 0.213968284549982
22.955935001 0.214301033063429
23.077544689 0.214002065022394
23.148656845 0.214506880468631
23.240077257 0.213379101747082
23.3213346 0.213331173121102
23.422781467 0.211397273054864
23.514026403 0.209089302774732
23.585143328 0.20752876102195
23.716991663 0.207767035873561
23.777933359 0.205955012316246
23.86947298 0.206302749298375
23.950687885 0.202194338861132
24.021820545 0.208829918321082
24.153472423 0.217440176079213
24.264950514 0.21373284085367
24.346101522 0.207692479643458
24.417271614 0.21070644792932
24.508569956 0.21180113083074
24.579620123 0.209777930643103
24.660811186 0.207396897044804
24.752292633 0.204970258220914
24.863802433 0.205138316759409
24.955062628 0.207281765645834
25.066570759 0.20597075411848
25.147779226 0.206719134344136
25.21885395 0.207071088892188
25.350507259 0.20637738634888
25.411503076 0.206255725564869
25.502901554 0.205856220211465
25.584325314 0.206644847219716
25.705877066 0.205304471830823
25.776982307 0.205066467696543
25.858206272 0.204822650258424
25.980067015 0.203899302279907
26.05139637 0.207731144366907
26.183240175 0.204368424389937
26.244215488 0.207031687970254
26.335502624 0.200923058006521
26.416719198 0.206890893217339
26.477708101 0.209313058048637
26.619707823 0.208543632363464
26.731155157 0.205907973101815
26.822526693 0.215479972483191
26.923908234 0.215609073231259
27.01519227 0.214395565610259
27.086347818 0.20676627985263
27.208006144 0.203026576148433
27.289249659 0.204776183119702
27.411138534 0.206308780377409
27.482315779 0.204833973612136
27.563636303 0.203976566873928
27.644851208 0.204446434979821
27.715904236 0.205087807825504
27.807189941 0.204570561102808
27.888353586 0.203764893681839
27.999786377 0.204496098447877
28.081002712 0.204617461697159
28.152292967 0.203632844768067
28.283961296 0.204466812032196
28.344907522 0.20550920123218
28.42610383 0.199219686070785
28.547697306 0.208352905465548
28.618802071 0.20396954861485
28.75050807 0.210124069883919
28.811588287 0.211595504218838
28.902893305 0.210342322025478
28.984147549 0.20982895326166
29.045192957 0.211156862116214
29.136545896 0.217240133031897
29.217670441 0.205835142172298
29.319195032 0.205723720266276
29.450890541 0.208034994820104
29.562416553 0.207121363339024
29.653707981 0.205765151810818
29.7550807 0.208339395800971
29.876724005 0.208725090208487
29.948034763 0.206224602389457
30.039405107 0.213262314693206
30.110585689 0.203685549156037
30.201971292 0.205705230635738
30.2932899 0.211960286412202
30.344251871 0.205741951489972
30.435790062 0.204426406687961
30.517178535 0.204791126098333
30.598567963 0.204589946940521
30.67977643 0.204591461521006
30.750858545 0.204046872151053
30.882576466 0.203771873657768
30.94378829 0.203641620459094
31.02513361 0.203645493904112
31.116649389 0.204611195167466
31.177718162 0.202981266715891
31.269541025 0.203026308184973
31.350744247 0.203150927556919
31.432051897 0.201632542459173
31.543591976 0.203641630291774
31.624850511 0.202403287691922
31.746440887 0.210202824850465
31.817703247 0.209029944150653
31.94948864 0.205495485116947
32.010649443 0.206816351373691
32.112061024 0.204731137479373
32.183124304 0.210678443196666
32.254178286 0.213431216267949
32.345468759 0.211402344227865
32.416581392 0.2078375276304
32.558401585 0.206070073804609
32.659913063 0.20330224393536
32.751293659 0.20600442439828
32.862970114 0.206012867115924
32.95425725 0.205129156646854
33.065869331 0.204677529157112
33.147162199 0.206185187017938
33.218403101 0.206222803132991
33.31987977 0.205694818273101
33.43146348 0.205194197899857
33.512907267 0.204972829979463
33.584142685 0.204118362803688
33.665349722 0.205340326151878
33.787038803 0.202336363510317
33.908681869 0.207676334796562
33.989887237 0.20504563782397
34.111577034 0.20500480896338
34.182807207 0.210266943838393
34.264139414 0.209960652181955
34.345333338 0.213601418184365
34.4570086 0.209744439261529
34.54838109 0.206718160200688
34.639740467 0.20413783201061
34.791864157 0.20550402710603
34.913450479 0.205545190543037
34.984593153 0.205390674388845
35.116267443 0.204931211657685
35.177255154 0.205892321656155
35.319267034 0.206705023730817
35.430798531 0.205559554702436
35.5120399 0.206353384695806
35.583136558 0.204328592663091
35.674642324 0.203832170675496
35.745771408 0.2072533483537
35.816863775 0.204327762290886
35.908101082 0.207040903301834
35.979290962 0.198187393898111
36.060506582 0.202605726086136
36.151916265 0.20886944887245
36.263616085 0.205348157834706
36.355079651 0.204077650599978
36.456537485 0.203971989577723
36.548088074 0.206040791418658
36.619627714 0.206810327996829
36.751798153 0.206131294228856
36.863384008 0.205475345750982
36.944591045 0.207592732754253
37.015696526 0.207470525667182
37.107084751 0.208252934807505
37.178186417 0.205887907046551
37.269515753 0.208932074002499
37.35068655 0.211463515208895
37.411700964 0.203212010296308
37.503101349 0.204241196654239
37.584325075 0.207954323575005
37.685910225 0.213091105101448
37.787325382 0.209767017664223
37.888715506 0.208388991393752
37.980068445 0.204824003425606
38.051229715 0.20311018467436
38.183126211 0.204484441507747
38.244180918 0.204639979447991
38.335587263 0.204798316943807
38.416853905 0.206152729076864
38.477903604 0.203788747612384
38.569523811 0.203014435142707
38.650883198 0.203631825858355
38.772616386 0.201864431064914
38.853896856 0.201818697150504
38.914894104 0.202969534405453
39.006242752 0.202493491558047
39.077589989 0.202312802770553
39.158942938 0.201187564384551
39.250485182 0.203364362554635
39.311570644 0.211375831271414
39.402840376 0.212527394672403
39.484099388 0.20906182767238
39.545132637 0.208074381673212
39.636439323 0.20353666080573
39.727715492 0.20507314375054
39.82909441 0.203904538994274
39.950762272 0.2040969186659
40.011919975 0.201714530447794
40.10332942 0.20217954088527
40.184502602 0.203737295641242
40.265665531 0.201309669237382
40.346873999 0.209274060559831
40.417939186 0.207013937823102
40.549581528 0.216433498983053
40.620654345 0.209159313638275
40.742346287 0.20676717602451
40.813573122 0.207481093075706
40.901032925 0.204617031427278
};
\end{axis}

\end{tikzpicture}

%% file: tex/realworld/e3p3.tex
% This file was created with tikzplotlib v0.10.1.
\begin{tikzpicture}

% \definecolor{darkslategray38}{RGB}{38,38,38}
% \definecolor{lavender234234242}{RGB}{234,234,242}
% \definecolor{mediumseagreen85168104}{RGB}{85,168,104}
% \definecolor{peru22113282}{RGB}{221,132,82}
% \definecolor{steelblue76114176}{RGB}{76,114,176}

% \begin{axis}[
%     width=\figurewidth,
%     height=\figureheight,
% legend style={at={(0.99,0.99)},anchor=north east, nodes={scale=0.8, transform shape}},
% axis background/.style={fill=lavender234234242},
% axis line style={white},
% tick align=outside,
% x grid style={white},
% major tick length=2.0,
% xmajorgrids,
% xmajorticks=true,
% % xmin=-1.96453596335, xmax=42.83319236035,
% xmin=-1, xmax=41,
% xtick style={color=darkslategray38},
% y grid style={white},
% % ylabel=\textcolor{darkslategray38}{Dist. to target (m)},
% yticklabels={},
% xlabel=\textcolor{darkslategray38}{Time (s)},
% ymajorgrids,
% ymajorticks=true,
% % ymin=0.37598902295362, ymax=3.71917413408861,
% ymin=-0.1, ymax=3.1,
% ytick style={color=darkslategray38},
% xtick distance=10,
% ytick distance=1
% ]

\definecolor{darkslategray38}{RGB}{38,38,38}
\definecolor{lightgray}{RGB}{192,192,192}
\definecolor{mediumseagreen85168104}{RGB}{85,168,104}
\definecolor{peru22113282}{RGB}{221,132,82}
\definecolor{steelblue76114176}{RGB}{76,114,176}
\begin{axis}[
    width=\figurewidth,
    height=\figureheight,
    axis background/.style={fill=white},
    axis line style={color=lightgray, line width=0.5pt},
    tick align=outside,
    x grid style={color=lightgray, opacity=0.3},
    y grid style={color=lightgray, opacity=0.3},
    major tick length=2.0,
    xmajorgrids,
    xmajorticks=true,
    xmin=-1, xmax=41,
    xtick style={color=lightgray},
    % xticklabels={},
    % ylabel=\textcolor{darkslategray38}{Dist. to target (m)},
    xlabel=\textcolor{darkslategray38}{Time (s)},
    ymajorgrids,
    ymajorticks=true,
    ymin=-0.1, ymax=3.1,
    ytick style={color=lightgray},
    xtick distance=10,
    ytick distance=1,
    % title=TARGET 1,
    % Scientific notation for y-axis if needed
    scaled y ticks=false,
    yticklabel style={
        /pgf/number format/fixed,
        /pgf/number format/precision=1
    },
]

%\addlegendentry{SARL}
\addplot [thick, steelblue76114176]
table {%
0.071724415 1.85194561252505
0.173231602 1.85203098124556
0.274574518 1.85202321351863
0.366084337 1.85257351577966
0.467458725 1.8526603406373
0.558732748 1.85126696207239
0.670309544 1.85174021873483
0.802333355 1.84900608881218
0.913858414 1.844134887987
1.005226135 1.83447465699484
1.116688728 1.82674180155119
1.238280773 1.82583892620448
1.349887848 1.81711926377343
1.441202164 1.80545709543183
1.542807579 1.79807279897001
1.664456129 1.79057453739525
1.74572444 1.78030167429947
1.867347717 1.76856853961741
1.928331852 1.75596072117643
2.029666424 1.75209764367897
2.100785971 1.73577224267675
2.232566357 1.72902439359637
2.303730726 1.718242267148
2.425456524 1.70648219510163
2.496614218 1.69578806649849
2.628310442 1.69125073038585
2.739787817 1.67516006897339
2.861327886 1.66289509096308
2.932612181 1.65015242262107
3.023890018 1.6430001930449
3.105118036 1.63409604021643
3.206689835 1.62439536930963
3.308115244 1.61184042472861
3.40956521 1.60296478625137
3.500914574 1.59312028148277
3.561922312 1.58215490781199
3.653232574 1.5740779013798
3.73444891 1.56598201729794
3.846027613 1.55717404667138
3.937556982 1.54395406216444
4.039197206 1.53438814284798
4.160977602 1.52238352754669
4.23217392 1.50990474166971
4.323533296 1.50469653123313
4.404682398 1.49457770641798
4.526286602 1.48385019117235
4.637919903 1.47288830879037
4.769782305 1.46004019681081
4.871256113 1.44568174107074
5.002973318 1.43843001004711
5.104353666 1.41967707276101
5.23622489 1.40817350806993
5.357816934 1.39706221346555
5.429092646 1.38038546115175
5.560770035 1.37356033146034
5.631811619 1.36238234243442
5.723044872 1.35343735518861
5.80419898 1.34482828419885
5.905589819 1.33680373404643
6.037261486 1.32556855773057
6.138811827 1.31092153573954
6.270653963 1.30335125201398
6.382239818 1.28306555425621
6.50388813 1.27462454146562
6.605402708 1.26063543710514
6.737170458 1.2469205374369
6.848749638 1.23673471984802
6.929927826 1.22089933774799
7.041383982 1.21131561368944
7.162965059 1.20147516921945
7.234078169 1.18504545981761
7.325454235 1.18014190883702
7.406766891 1.17242546359804
7.508184433 1.15805793844882
7.629737377 1.14558412079255
7.741327047 1.13176371615505
7.862930775 1.11823514118328
7.933951139 1.10590553528653
8.075833797 1.09559235847822
8.177241087 1.08196301821841
8.268575907 1.07030416625532
8.380153417 1.06161217879706
8.471558571 1.0512516755734
8.573143482 1.04436564608374
8.664488554 1.02743464508835
8.745796204 1.02153285518656
8.83708787 1.01329802441469
8.948714495 1.0018555064386
9.070395946 0.989566180249842
9.202201366 0.979643351867969
9.303639412 0.963170175279406
9.435664177 0.95549652634286
9.527144432 0.938499653382102
9.618673324 0.93313435105868
9.720078468 0.92499085127082
9.811460018 0.91267108361844
9.902779579 0.90437479949726
10.014237881 0.898219278257226
10.13590908 0.886156748005931
10.247391939 0.877408842875559
10.338649034 0.866282202241995
10.450085878 0.854117130630097
10.531255722 0.849021191025563
10.602373361 0.841206988302448
10.734097242 0.83359720359742
10.79518342 0.824162579750988
10.886454582 0.821615206286243
10.967713594 0.811574696186987
11.089354515 0.805763944363793
11.170614958 0.798558203228412
11.292221785 0.78824355361668
11.37354064 0.783575332947222
11.495146036 0.777657000361059
11.566216946 0.766447640655369
11.64762783 0.765806882646837
11.739164352 0.761644925237451
11.85063529 0.756213482031021
11.931854963 0.747825365782082
12.043357134 0.742935802011783
12.175032616 0.735631640748525
12.276677847 0.73146286073707
12.368052244 0.72318007745833
12.479692697 0.720273206760798
12.560863256 0.718534988599473
12.631971598 0.704669323264424
12.733324289 0.709322056649669
12.804437637 0.706476756376191
12.926072597 0.698313473725384
12.997215748 0.694937430288745
13.068366527 0.691120936276414
13.200060606 0.688547434658483
13.27109313 0.681357299828258
13.392660141 0.682568588838958
13.463706255 0.668604469012283
13.544871569 0.670411217003941
13.666649341 0.670915651842531
13.86913681 0.664358341222725
14.031311273 0.663244049707117
14.23371768 0.660284677323911
14.304781198 0.64549656980855
14.497112989 0.640217971164171
14.699430227 0.640184153956392
14.902345419 0.642281734590574
15.034217596 0.640144152560624
15.105530023 0.641605859389465
15.227189302 0.638872863805214
15.298568964 0.640004167771931
15.420265198 0.6379678960275
15.501471758 0.63557971675691
15.612941265 0.630499822541644
15.704160213 0.62513544396425
15.815708876 0.626676028032225
15.906984091 0.628722785268354
16.008312225 0.624113704761317
16.12989831 0.624269362811679
16.251458406 0.625186865381565
16.332784414 0.62350251614444
16.403927565 0.625256184128761
16.535809517 0.623130696799541
16.647230625 0.622289868243692
16.768951178 0.623516023526292
16.880489588 0.61960032200291
17.0023036 0.617299465590494
17.113780022 0.619521695362749
17.205046177 0.620442584576608
17.306491375 0.617920480709388
17.428225994 0.620052809130926
17.499399424 0.61931511097896
17.590723514 0.620335873386921
17.671904325 0.62119302422718
17.793506622 0.622479292496518
17.864623547 0.62212045756684
17.945833206 0.621218684268994
18.037109136 0.627241551762358
18.148637533 0.62438330092619
18.23995614 0.622719492077767
18.351384163 0.62629737371948
18.442718029 0.631622325420425
18.544070721 0.628289350116065
18.635756969 0.629293214155895
18.747252941 0.635365982835721
18.838528395 0.630087071526892
18.949982881 0.636488782724898
19.031194925 0.634438889459059
19.102301359 0.636488366334774
19.23395133 0.63879247406394
19.304975748 0.639798999789351
19.436784267 0.642062638086988
19.548248768 0.645321508528785
19.639599562 0.647413365294461
19.791493893 0.653108599288518
19.872672319 0.652471169845209
20.004348755 0.655690605300424
20.126134157 0.6603896208569
20.197429895 0.658854758725893
20.269055605 0.661792776504344
20.400823116 0.666030925782535
20.461866617 0.66560653673572
20.543039083 0.668022275885352
20.634260416 0.66733289314727
20.705403566 0.669134288068416
20.837031126 0.673255115434493
20.928332567 0.677490749086983
21.029673815 0.677765825639162
21.100754261 0.67813323782726
21.181982756 0.681315079981402
21.303580999 0.685215383380192
21.415062427 0.684750985225612
21.496332884 0.68698667440536
21.617779255 0.689528143340149
21.698996544 0.691896331938303
21.770147085 0.69553071241712
21.90196681 0.698068955953712
22.013448477 0.698184672083657
22.104650021 0.699452773963331
22.216115236 0.701734999883127
22.307465315 0.705097803929637
22.419013739 0.709081813728213
22.510318756 0.710082321405586
22.611813307 0.711483774080443
22.703179598 0.713214514451354
22.814729929 0.717205298172243
22.906009197 0.720725673646848
23.017437458 0.722315472567496
23.098589659 0.724051720766623
23.169666767 0.72807805798981
23.301326275 0.727997634911182
23.412872314 0.734266958003874
23.504260778 0.735045362240657
23.615972042 0.738596435721727
23.697132111 0.743438785485366
23.818951607 0.744691443990091
23.900255918 0.747277187205858
23.971444607 0.748422011207795
24.103221416 0.749768684764244
24.21487236 0.754864640645098
24.336662531 0.758004958867242
24.438150406 0.760957766338696
24.570245027 0.762617890151767
24.672152996 0.764857725012065
24.804064512 0.768900856219095
24.905810833 0.770892892771548
25.037529945 0.775854274549058
25.149316549 0.781145834534897
25.230550289 0.781817407837643
25.301671028 0.78385654142315
25.433395624 0.785093946348428
25.545107365 0.790241196005294
25.666894913 0.797105612648926
25.788576126 0.798158060804698
25.869776964 0.80245997354122
25.981416225 0.802391802027822
26.103102922 0.807332262528099
26.214533806 0.810870903616158
26.305855513 0.813812491965789
26.417225838 0.81710315720523
26.50844121 0.81892645043814
26.610211611 0.821444080144035
26.711699009 0.824129278472467
26.81317234 0.826286289914289
26.904491424 0.82867694894113
27.016091823 0.832966101766067
27.137720823 0.834624908121458
27.249272823 0.838999863126522
27.330457687 0.842357465813168
27.401582003 0.844985599121347
27.533352852 0.847560159496368
27.604452133 0.849467716563106
27.726101398 0.853330190060529
27.807281017 0.859419755263076
27.908698559 0.861472908099979
28.030290365 0.867066804002167
28.101388931 0.869574457400883
28.233110428 0.871891107985497
28.294110775 0.875724286259281
28.395615101 0.87460069394437
28.466760397 0.870215821422021
28.588290453 0.878060572262312
28.66958189 0.880610811601446
28.791106224 0.877518284960997
28.862337112 0.886114028001113
28.943464041 0.887900044091594
29.034771919 0.886157795669021
29.146276712 0.887556839270176
29.267917871 0.892349960770434
29.379561424 0.895499365684843
29.470849037 0.898650526733268
29.582299471 0.902787213461269
29.703949451 0.904200227472042
29.815510034 0.906597107984294
29.906780481 0.910292058249862
30.018265009 0.914877739346555
30.109489202 0.918406106005288
30.210842132 0.920976693589182
30.302318096 0.922445910180665
30.413789987 0.921821510811138
30.505044937 0.924157080109343
30.606534719 0.927488854430412
30.697874308 0.924286765304738
30.769073725 0.92584517683744
30.901041269 0.932555247736237
31.012797117 0.934847704573986
31.13443923 0.934315198196694
31.245999575 0.939926452391683
31.367542744 0.942171681614661
31.438703537 0.944961823257105
31.560320139 0.947264942661646
31.631467342 0.950455916871903
31.752992391 0.954244472490359
31.834162474 0.956149214253219
31.905378818 0.956854550869487
32.006823063 0.95508535254052
32.108302593 0.952660394494756
32.199607372 0.960340964434351
32.270702124 0.961189475218162
32.392355442 0.958740321701654
32.463437557 0.964092620508878
32.544645548 0.966289453823802
32.666483402 0.96747138532878
32.778053999 0.9720034698261
32.859298706 0.975699114619649
32.930355549 0.977001019648354
33.021660089 0.97853172236053
33.102882147 0.982447113711786
33.214356422 0.985258335907264
33.305889606 0.979536530040486
33.417429447 0.979011038369234
33.508738041 0.984757671066042
33.610107183 0.983110574549365
33.701426983 0.984699161163088
33.802829742 0.990712353054368
33.93456316 0.991982890868694
34.046240568 0.99289561491602
34.137492657 0.996981733493952
34.249010324 1.00058311650239
34.330202103 0.997836551571062
34.401371956 0.996075927347768
34.533064842 1.00155384964013
34.604151249 1.00479821127237
34.736107111 1.00123505772249
34.847582817 1.00608187404737
34.969203949 1.00987580220728
35.090738773 1.01215252061725
35.171887875 1.0159393630778
35.30357027 1.0190037647589
35.415055275 1.02249202193908
35.506377458 1.02455719162479
35.61776042 1.03138788792943
35.698907852 1.02320108524539
35.769938707 1.02582472845986
35.901473999 1.03144990773698
36.002813816 1.02807135695442
36.134624243 1.02913724717149
36.246162176 1.03338613388794
36.367812157 1.03147252084028
36.479300976 1.0386064901318
36.570567608 1.03372880943874
36.682118177 1.03629732033577
36.773533106 1.0433477136163
36.87517786 1.04009109705033
36.966669321 1.04061891447243
37.037811994 1.04308072800578
37.169562101 1.0453659039033
37.281036854 1.04832309092475
37.402531624 1.05381849629047
37.513968468 1.05799193570535
37.605551243 1.05818581891316
37.717061758 1.05655233350621
37.798227549 1.05864653031623
37.86936903 1.06208416385232
38.001063585 1.06118623302028
38.112744331 1.05710474318561
38.204064608 1.06496427726984
38.305658102 1.06480973456648
38.437379599 1.05963396129518
38.53885436 1.0671791248058
38.670562982 1.0672798771403
38.782125473 1.06407589710028
38.903749227 1.07105413954239
39.005200148 1.06298688800817
39.137006998 1.06875356757657
39.248550653 1.07138423453542
39.329730749 1.07005482169509
39.400811672 1.07478084600619
39.492094517 1.07736890467939
39.563174963 1.07763675071783
39.644364357 1.07765935867415
39.735668898 1.07854136743588
39.847147226 1.08099635130537
39.938516617 1.08484694531863
40.050209522 1.08625129245175
40.161833525 1.08740942350286
40.273386001 1.08315537232577
40.364731073 1.08836709302756
40.435882091 1.08834326887442
40.567749023 1.08186905208977
40.669136047 1.08818682816831
40.796931982 1.09013121053692
};
%\addlegendentry{MARL (global)}
\addplot [thick, peru22113282]
table {%
0.092583179 1.86451778855232
0.194922447 1.86465501219478
0.286361933 1.86411313272608
0.387781381 1.86149263387757
0.489301205 1.86451036167983
0.621103525 1.86516940983382
0.712264061 1.8649545591572
0.803450823 1.86669600625529
0.884753465 1.86735134759409
0.955881834 1.86856426346313
1.087512493 1.8698300407516
1.198828459 1.87076534639198
1.279999018 1.87550951437899
1.351009369 1.87436823006394
1.442220449 1.87747711950387
1.513380527 1.87766416497493
1.594554186 1.876436164885
1.686106205 1.88125083457364
1.797661781 1.87938004302064
1.919471025 1.88189922390739
2.030974626 1.88569908444492
2.152626276 1.89042764962415
2.264149427 1.89391572262658
2.386070013 1.8953463881897
2.497653007 1.89649247929854
2.589056492 1.89854321872955
2.700544596 1.90489793664204
2.781845093 1.90932612202526
2.8529706 1.91264555126764
2.984636545 1.91029619923837
3.045664072 1.91563231609092
3.13692522 1.91702179096586
3.218178272 1.91902327435748
3.279244184 1.92397521194269
3.370567083 1.92836635091637
3.451758146 1.92938115136499
3.512748241 1.92869519998597
3.604014635 1.93501437651811
3.685192108 1.93331048221931
3.766499996 1.9357975529706
3.888127327 1.94432885977345
3.989520311 1.94736851471317
4.121195793 1.95115034457564
4.232689142 1.95394543186556
4.354331016 1.95678994688426
4.455869913 1.96476780217966
4.557431698 1.9668560501495
4.658877611 1.97044788830899
4.790553093 1.97430002182433
4.891908407 1.97763514686479
4.983268499 1.98200181009464
5.054379463 1.98721257815411
5.186335325 1.98464352641144
5.307927847 1.99335719605988
5.379223108 1.99712221810824
5.470500469 1.99993984248077
5.551693439 2.00285243471205
5.632869482 2.00668988926817
5.724238396 2.00929120105494
5.836004257 2.01417125666324
5.917248964 2.01992035296611
6.028691292 2.02369437430432
6.119920015 2.02921293095269
6.231420994 2.03294866405794
6.353084802 2.03808839368664
6.464566231 2.04438379665099
6.586373091 2.05043863176934
6.697846412 2.05688688055267
6.789254904 2.06320999927817
6.90071702 2.06922063060877
6.981879711 2.07313876629341
7.052934646 2.07626447280828
7.184619903 2.07939697363951
7.255828619 2.08739253555816
7.387626171 2.09188543522824
7.489121675 2.09735654075369
7.580407381 2.10108185411076
7.651523113 2.10483469814608
7.742842197 2.10855932991044
7.824065685 2.11451931211718
7.94572711 2.1219265332291
8.016999483 2.128028214903
8.098264694 2.13227952284727
8.189558983 2.13765968072961
8.301150083 2.14408891560395
8.38237071 2.15081056268221
8.453463554 2.15671522935914
8.585158586 2.16324547339191
8.656392813 2.16922202548423
8.778258085 2.173705868002
8.849342584 2.18284660842069
8.930520296 2.18693899684052
9.052144766 2.19188346785032
9.163594484 2.20240078998808
9.285203934 2.20748110799316
9.356425285 2.21844584747558
9.488125562 2.22236781251186
9.599686384 2.23394021371733
9.68097043 2.24145252011766
9.752342939 2.24606213307608
9.884162903 2.25214808323137
9.955358982 2.26244174826723
10.077052831 2.26743065925737
10.148160934 2.27721289749786
10.229406833 2.28233388328674
10.350970745 2.28817345369923
10.422013759 2.29763707051657
10.553750753 2.30333632229457
10.655285358 2.31527431428054
10.78730607 2.32227414396063
10.899082422 2.33271566657461
10.99045825 2.34230969564339
11.091817856 2.35013532492536
11.183095693 2.35978404453014
11.254199505 2.3673450144874
11.38602519 2.37245970576962
11.507660627 2.38487506203478
11.588860988 2.39644802052594
11.710455417 2.40321046496035
11.811947346 2.41442424679729
11.913345575 2.42409638960334
11.984448671 2.43347170170867
12.055624485 2.43880589247108
12.187230348 2.44493609735448
12.298676491 2.46088226016479
12.390090704 2.4707721064309
12.501526356 2.47801002718188
12.582874298 2.48884819667082
12.653935432 2.49775315648645
12.785721063 2.50486843740439
12.897306919 2.5177009622292
12.988752127 2.52785604662751
13.10020256 2.53985579258207
13.181320906 2.55426094572972
13.252589464 2.55918289755654
13.384256601 2.56832110448287
13.455346584 2.58146378026151
13.587188005 2.59294525470399
13.69866538 2.60723545298729
13.790193319 2.61709509415188
13.901680708 2.62921723328385
13.982895851 2.64286481772374
14.053942203 2.65243697531958
14.185819626 2.65989820460297
14.297421932 2.67470778962244
14.388719559 2.68956761397023
14.500314236 2.70165573765868
14.581524372 2.7136698374547
14.652650595 2.72518624529314
14.754004478 2.73336669528419
14.875699758 2.74445728655609
14.956979513 2.76123951361593
15.078654051 2.77076026922967
15.149805546 2.78706728143561
15.231040716 2.79423071677761
15.322323799 2.80600175271554
15.423646927 2.81950505755248
15.555289745 2.831256774436
15.656876325 2.84716008216586
15.778399944 2.86533703761608
15.849491596 2.87870670556257
15.940827846 2.88716253364849
16.032115459 2.90327436468475
16.143700599 2.91658659273936
16.214788675 2.93052154796713
16.29607439 2.93900788192026
16.417650223 2.95286713089964
16.47859931 2.96991121201829
16.57999134 2.97595988275997
16.651059627 2.99065363649891
16.722153902 3.00068454270807
16.853880167 3.01463371819164
16.965411901 3.03159746188118
17.08706522 3.04764327034625
17.188461304 3.06559272867519
17.32038331 3.08201814531564
17.431982517 3.10112143021638
17.513270378 3.11850535262918
17.584366321 3.13137237119972
17.675709963 3.14231519630357
17.756915808 3.15232795241165
17.858174801 3.16594844643589
17.979822635 3.18421016507607
18.050937176 3.19959153904758
18.152549267 3.208393111261
18.274341345 3.22948403313018
18.355551719 3.24715848171918
18.467091322 3.26241131837463
18.578905105 3.27951354569603
18.650107145 3.29404245744925
18.7414155 3.30582910265654
18.822632313 3.32095611276019
18.934117079 3.33175172468402
19.025413275 3.34954643448647
19.126973629 3.36528224129003
19.228595018 3.38142347168995
19.32998538 3.3960194083051
19.421276808 3.42438333567998
19.532901287 3.42610629016332
19.614053964 3.45733575492298
19.68518734 3.47578482017528
19.776506185 3.48909491467452
19.847593546 3.50118560622954
19.979306698 3.51898629199857
20.050417423 3.51748956129562
20.131651163 3.53027178711211
20.243569851 3.55277452948572
20.324832678 3.56721117449157
};
%\addlegendentry{MARL (partial)}
\addplot [thick, mediumseagreen85168104]
table {%
0.074190855 1.84996117034925
0.145467758 1.84941128666379
0.206692219 1.85526315142456
0.277895928 1.85530434083754
0.338966847 1.85186446805027
0.410219908 1.84584444266226
0.471180916 1.85461285219024
0.572656632 1.85519453275867
0.6437006 1.84973797804038
0.714805603 1.84661479096541
0.8062253 1.82979713534264
0.887481928 1.82441699961855
1.009076357 1.82342880269018
1.080456257 1.8095494134128
1.212171078 1.79726153868271
1.303436518 1.78331610434914
1.374527932 1.77862318683341
1.455718994 1.76886873455216
1.547148705 1.75514244820524
1.658670187 1.7469047784854
1.739805222 1.73697034232898
1.810903549 1.72268633466438
1.902219534 1.71419078683447
1.973470927 1.71350751651654
2.054763079 1.7053950139727
2.146140814 1.69443547604116
2.257860661 1.68552995676634
2.349377394 1.66401027247916
2.450792075 1.65895866955399
2.552173138 1.64969583244753
2.603026629 1.63718605182312
2.694271803 1.62147019983593
2.775494814 1.61943816106237
2.846761465 1.60753553895408
2.978490591 1.6032524258333
3.039622546 1.58746233596771
3.131084442 1.58011122369632
3.212327004 1.57296468059461
3.293496847 1.56468675898797
3.374803067 1.55189870339662
3.446065426 1.54619584923721
3.57788682 1.54055471816702
3.649019957 1.52196548223159
3.770645857 1.5174763027387
3.851837159 1.5020035761981
3.93311882 1.49254294699024
4.004155636 1.48695847576272
4.085351229 1.48126548291269
4.206969738 1.46804811535295
4.278044939 1.45918729587175
4.369474173 1.4566671490173
4.440628052 1.44072628351033
4.521886826 1.4293419626489
4.613203764 1.42786857037466
4.674168349 1.42459748365319
4.775538206 1.41121579192389
4.846650839 1.3968528561348
4.978406906 1.3954651085757
5.03947568 1.38308061646382
5.130960942 1.37048903907249
5.212265492 1.36275530998289
5.283394099 1.36214493435229
5.374685526 1.35437542001544
5.486276388 1.33547861176787
5.577632666 1.32886715863301
5.648843766 1.32073013350584
5.770698548 1.30834042051538
5.851950407 1.29846530116395
5.943308592 1.29346339654075
6.004299403 1.27723818132086
6.085615635 1.27150882473411
6.176920176 1.26601102283242
6.237912655 1.26038098146796
6.329417229 1.24953715088444
6.410704851 1.23720718836863
6.481777668 1.23335211449021
6.573021651 1.22702918939047
6.644077778 1.21501974521997
6.735323191 1.20700703034838
6.806362868 1.1994040539939
6.887485504 1.19302705203006
6.978727103 1.18205256629505
7.039724827 1.17135588736552
7.131052256 1.166159462982
7.212230683 1.15662102670114
7.293366909 1.14875670256163
7.374532461 1.13992979603533
7.445646525 1.13520503439335
7.536938191 1.12774014511435
7.608066559 1.11505373355273
7.689322949 1.10981655302423
7.780707836 1.10323505538336
7.892198086 1.09363513955921
7.973371029 1.08217804150242
8.04446888 1.07419772435792
8.145906687 1.06752729170352
8.247414351 1.05579836084166
8.379085303 1.04787072251722
8.440103293 1.03212913098602
8.531413317 1.02532541516881
8.612647057 1.01599264516101
8.693836689 1.013486387303
8.785112858 1.00189236077275
8.886554241 0.991762303327486
8.987914801 0.982566337673623
9.099446535 0.975180525490795
9.17062521 0.960045245021132
9.251813889 0.954246251682423
9.343163967 0.933700325030694
9.414239168 0.92696732297473
9.545915366 0.933243080384886
9.657427311 0.908346577369208
9.738601446 0.891328575125896
9.80980444 0.885687772208089
9.901167631 0.882791475636668
9.972225905 0.873785450177749
10.043537617 0.866804420241551
10.144949436 0.876967394076973
10.256539345 0.84443041445955
10.347922802 0.841105246326255
10.479605913 0.828354138696138
10.611343146 0.835832516425424
10.672317028 0.830768540445073
10.763592482 0.803610661318966
10.844761133 0.815261758621509
10.915879965 0.802043011668176
11.04769826 0.800033614729374
11.149151564 0.786353135200962
11.250658274 0.786276751340184
11.352048874 0.77321529777584
11.44327879 0.757221071145763
11.504221678 0.742451013236846
11.595515728 0.750100891717708
11.67673254 0.73527645181226
11.758028984 0.743489716218982
11.839241267 0.740469171008973
11.910378456 0.726559197710063
12.001828194 0.72627606938548
12.072985888 0.717721099889121
12.154200554 0.712068327852957
12.245630026 0.699708358174226
12.347764969 0.702363654449642
12.479614497 0.682580893931467
12.591109753 0.677223637943221
12.672286034 0.67916830478664
12.753472328 0.664453629633634
12.844750881 0.66684307312746
12.905730009 0.657658530647625
13.007056952 0.666676604416721
13.078104258 0.642400889207323
13.169428826 0.645227798093598
13.240558625 0.648182166029197
13.382366419 0.636375618249139
13.44354105 0.624169692111896
13.534839869 0.618704979672008
13.606016398 0.634704057288921
13.687325239 0.619089875606203
13.808913946 0.614712096486047
13.870020151 0.609257342702185
13.961326838 0.60310473283679
14.042526484 0.592858045000531
14.123739243 0.604183530114656
14.204987765 0.592997586262157
14.276096106 0.604501153388146
14.367440701 0.59246117800279
14.448615551 0.58323960842322
14.570300579 0.589502635706236
14.671675206 0.62018245795803
14.813417435 0.578148085819864
14.874373198 0.603531495955778
15.00609541 0.575268793399202
15.077159882 0.619458948600442
15.219058514 0.579011667410627
15.320463419 0.578786918946736
15.421971798 0.565949812989933
15.472990752 0.568997424292626
15.564289332 0.578609631404436
15.645507574 0.571267283322384
15.716706276 0.574027621016596
15.81808281 0.570170533683366
15.919490576 0.562629985327111
16.010770321 0.564598548466666
16.071717739 0.573186218457431
16.163039685 0.574135191115802
16.244361401 0.571192735928267
16.305442095 0.568681040063719
16.447494507 0.565161240697071
16.549242497 0.607105025250107
16.650666714 0.600665884631394
16.752016545 0.596085687986877
16.84340334 0.561552656466964
16.914613962 0.560946501768168
17.036338568 0.562621404368122
17.107478381 0.555283479711632
17.18871665 0.56185881544937
17.280267477 0.560024982194125
17.381739617 0.556808671505423
17.473166704 0.554716661947833
17.544224978 0.555371645603942
17.615268708 0.596590238913652
17.747141362 0.552717912775337
17.858821154 0.597171942308104
17.980581999 0.567925831039412
18.092142821 0.554772164533402
18.173377276 0.550219593416533
18.244439364 0.553231415262054
18.335988522 0.555570360831072
18.407109261 0.549303904246153
18.478160143 0.547429847123867
18.569489718 0.547062275540332
18.640607357 0.54883653548696
18.721847058 0.550379776540459
18.803209544 0.550130521940172
18.87434721 0.550264479801277
18.965696574 0.545699081227698
19.037023068 0.547467886869167
19.118381262 0.546912816846137
19.209814549 0.548875354484008
19.270881892 0.546788433531009
19.362166167 0.54545350729624
19.443457842 0.542229060114395
19.514583588 0.545843395589087
19.646290541 0.544766660664758
19.747716666 0.541681301547164
19.838980675 0.541487547810784
19.91027999 0.540137482381533
20.00158 0.542131061539391
20.072677851 0.543166692166961
20.15394187 0.539992100801357
20.245338917 0.541939514234873
20.30649972 0.541778136633941
20.398004771 0.54071213551243
20.479269266 0.541274977294753
20.60095501 0.542045152336214
20.672122479 0.540250129729333
20.75334096 0.540467026887745
20.844611883 0.541016279272779
20.905616522 0.539936741175913
20.997120858 0.535255873065163
21.10863638 0.537195920932029
21.210023403 0.541546379086416
21.270970106 0.538209318843306
21.3521595 0.538847734333849
21.443413019 0.538518869271132
21.504485846 0.53896888932098
21.585955859 0.538621053216342
21.67732358 0.536402858364927
21.738361836 0.541634348788984
21.859978915 0.54113629564708
21.951317072 0.539191129502229
22.042573691 0.539311156715766
22.103580714 0.540073307685017
22.194856167 0.539693841086305
22.2861197 0.538952449955849
22.387528897 0.539202074539972
22.478813172 0.539061788642792
22.539885283 0.538957415161368
22.63148427 0.538355113056924
22.712762595 0.540004985665242
22.793960333 0.538024849039959
22.875211716 0.539631447115889
22.94634676 0.538994762445168
23.037672043 0.537564229611761
23.108800888 0.539455001756329
23.200233221 0.538945990216461
23.271395922 0.53955741614594
23.352838755 0.539268124833935
23.444102526 0.538699817235401
23.505207777 0.538020167503461
23.596497298 0.53936103677203
23.677660466 0.538391584250589
23.799224139 0.53731391444963
23.870402336 0.539717228314746
23.95165205 0.54034551565275
24.042960882 0.532565313890514
24.103899002 0.528843383519869
24.245644808 0.539007470976586
24.3066504 0.541549755091897
24.38787055 0.540686992055502
24.479239226 0.539747360812813
24.641306877 0.53724974076176
24.712409973 0.537792973633199
24.813777685 0.536893634962292
24.925242901 0.536295412850929
25.006397724 0.534352084558058
25.07751298 0.532588063994222
25.178927422 0.532785848933059
25.239917279 0.540518641406709
25.321060181 0.537314842181328
25.442614794 0.530633187296098
25.503702879 0.532470311536793
25.594968319 0.534495022899437
25.676189423 0.540136698523535
25.747292281 0.540013791239291
25.83850956 0.539228292901172
25.909794569 0.539059068311121
26.001285076 0.539959363346955
26.072511435 0.540317614124418
26.143621445 0.539568709856406
26.235003233 0.536986818889221
26.306129456 0.537747126377413
26.387433768 0.537578578167421
26.479015351 0.536669172479087
26.540152788 0.535013602907982
26.631542921 0.536645814830442
26.713052273 0.534648981786363
26.834722758 0.534737291950012
26.905847073 0.534100323155443
26.997264386 0.534213531347969
27.078539849 0.539708765705505
27.139666558 0.53534087247225
27.231034518 0.530591124216845
27.312294007 0.531066752962113
27.383411408 0.534302745833612
27.47467494 0.54071446352578
27.545756817 0.540284851973317
27.63710928 0.53846173325782
27.708242178 0.538076882627638
27.789485216 0.539771304229275
27.911150217 0.540653482601716
27.972135067 0.537887143826251
28.063385725 0.53644546860328
28.144551754 0.536711880271764
28.205509901 0.537358602283235
28.296802759 0.536758466865576
28.378067494 0.53453245506833
28.439019203 0.531723913084679
28.530359268 0.531653267288069
28.611569882 0.531790376403914
28.672550917 0.537675129517968
28.763774395 0.533216111546361
28.8449049 0.533039814730414
28.915958405 0.532239657557945
29.007196427 0.534415969500802
29.078233719 0.536927737740292
29.159399748 0.537695307306907
29.25067997 0.536409056615086
29.352063418 0.536400913099615
29.443382502 0.539135825366207
29.504373074 0.53933320377318
29.595671892 0.53981728687278
29.676854849 0.539932896400271
29.758089304 0.540265634289845
29.839317322 0.537239803326463
29.910439015 0.53643427531432
30.011793614 0.53697554069561
30.072839976 0.536989850284462
30.164184809 0.536677551577977
30.245357514 0.536132048154792
30.326581002 0.531608870357337
30.407811165 0.531680603557111
30.4889853 0.531338514013464
30.580328942 0.539674453886255
30.691994906 0.53661535440694
30.783336401 0.527951982550665
30.84431696 0.533143639804656
30.925568581 0.53031411421191
31.016955376 0.536380369959519
31.138545037 0.535798083369881
31.209778786 0.54061811506127
31.290971756 0.540473817990877
31.38233757 0.539792517043175
31.443384171 0.539402942307707
31.534670353 0.539466610364194
31.605757952 0.536658765055377
31.677041292 0.536818281219643
31.758198023 0.536507272637546
31.839527607 0.536006170011697
31.91059804 0.532115559943749
31.991719961 0.532592318615538
32.072873831 0.531333924745808
32.14394784 0.531255218479053
32.23541832 0.530811596939591
32.306786061 0.534729374657278
32.388156414 0.536499469276109
32.5101645 0.529880208448462
32.581271649 0.530529335738721
32.672554255 0.536005416226723
32.743840218 0.535143275968427
32.825163603 0.535852268492084
32.906515599 0.53545605412747
32.977775097 0.536658750844258
33.069406033 0.538530294336321
33.140600682 0.539003217876611
33.221794129 0.538857911809986
33.303006649 0.539074858147839
33.374055624 0.540085411262749
33.465319872 0.540156478143012
33.546539307 0.539889677337482
33.61768198 0.536988910727732
33.708946944 0.536672216950233
33.780213118 0.537142982293772
33.911978484 0.536820180500567
33.983115912 0.536244667217068
34.07438016 0.532320787008367
34.145457506 0.531454318870857
34.226649285 0.531270115504943
34.338098765 0.534417633674625
34.429544688 0.534863981085461
34.510756493 0.531718863158408
34.591965914 0.534095707882625
34.673187256 0.540091436883023
34.744281292 0.54143231892603
34.835732699 0.538243499707522
34.906927586 0.538918645023226
34.998215437 0.539865572232814
35.079508543 0.540304131598423
35.190917254 0.540105715796614
35.282129765 0.537234611184459
35.343178034 0.536930317787637
35.434369564 0.537242081926531
35.505393744 0.53686687420821
35.596734047 0.533828370572402
35.677871704 0.532990677136831
35.738826752 0.535758080622328
35.830284357 0.534646451402182
35.911458493 0.534309081769602
35.982540608 0.53415022698082
36.073803425 0.537968509899565
36.144892216 0.532182892169761
36.236174107 0.530402973313887
36.30727911 0.531565634269692
36.388461113 0.533211002216626
36.469652176 0.540249326109722
36.540900469 0.540256575765666
36.632295609 0.539348982052792
36.703382254 0.538204420312923
36.784675837 0.540535765594581
36.876067877 0.53927469909522
36.937059164 0.537154374588537
37.038471937 0.535291936514799
37.112486363 0.536612476709715
37.183218479 0.536718160145818
};
\end{axis}

\end{tikzpicture}

%% file: tex/Gazebo/e3p2.tex
% This file was created with tikzplotlib v0.10.1.
\begin{tikzpicture}

\definecolor{darkgray176}{RGB}{176,176,176}
\definecolor{darkorange25512714}{RGB}{255,127,14}
\definecolor{forestgreen4416044}{RGB}{44,160,44}
\definecolor{lightgray204}{RGB}{204,204,204}
\definecolor{steelblue31119180}{RGB}{31,119,180}

% \definecolor{darkslategray38}{RGB}{38,38,38}
% \definecolor{lavender234234242}{RGB}{234,234,242}
% \definecolor{mediumseagreen85168104}{RGB}{85,168,104}
% \definecolor{peru22113282}{RGB}{221,132,82}
% \definecolor{steelblue76114176}{RGB}{76,114,176}

% \begin{axis}[
%     width=\figurewidth,
%     height=\figureheight,
% axis background/.style={fill=lavender234234242},
% axis line style={white},
% tick align=outside,
% x grid style={white},
% major tick length=2.0,
% xmajorgrids,
% xmajorticks=true,
% % xmin=-2.24838079205, xmax=49.03273762505,
% xmin=-1, xmax=41,
% xtick style={color=darkslategray38},
% y grid style={white},
% % ylabel=\textcolor{darkslategray38}{Dist. to target (m)},
% xlabel=\textcolor{darkslategray38}{Time (s)},
% yticklabels={},
% ymajorgrids,
% ymajorticks=true,
% % ymin=0.0298281469811177, ymax=3.06259763603752,
% ymin=-0.1, ymax=5.1,
% ytick style={color=darkslategray38},
% xtick distance=10,
% ytick distance=1
% ]
\definecolor{darkslategray38}{RGB}{38,38,38}
\definecolor{lightgray}{RGB}{192,192,192}
\definecolor{mediumseagreen85168104}{RGB}{85,168,104}
\definecolor{peru22113282}{RGB}{221,132,82}
\definecolor{steelblue76114176}{RGB}{76,114,176}
\begin{axis}[
    width=\figurewidth,
    height=\figureheight,
    axis background/.style={fill=white},
    axis line style={color=lightgray, line width=0.5pt},
    tick align=outside,
    x grid style={color=lightgray, opacity=0.3},
    y grid style={color=lightgray, opacity=0.3},
    major tick length=2.0,
    xmajorgrids,
    xmajorticks=true,
    xmin=-1, xmax=41,
    xtick style={color=lightgray},
    % xticklabels={},
    % ylabel=\textcolor{darkslategray38}{Dist. to target (m)},
    xlabel=\textcolor{darkslategray38}{Time (s)},
    ymajorgrids,
    ymajorticks=true,
    ymin=-0.1, ymax=5.1,
    ytick style={color=lightgray},
    xtick distance=10,
    ytick distance=1,
    % title=TARGET 1,
    % Scientific notation for y-axis if needed
    scaled y ticks=false,
    yticklabel style={
        /pgf/number format/fixed,
        /pgf/number format/precision=1
    },
]
\path [fill=steelblue31119180, fill opacity=0.2]
(axis cs:0.04,2.90258873982827)
--(axis cs:0.04,2.15124284406598)
--(axis cs:0.44,2.15089418534762)
--(axis cs:0.84,2.15474674388731)
--(axis cs:1.24,2.15874377696271)
--(axis cs:1.64,2.16286980663182)
--(axis cs:2.04,2.17038666566863)
--(axis cs:2.44,2.17429152199096)
--(axis cs:2.84,2.18388086469015)
--(axis cs:3.24,2.18927203464334)
--(axis cs:3.64,2.19529085519317)
--(axis cs:4.04,2.20432255579187)
--(axis cs:4.44,2.21415145395031)
--(axis cs:4.84,2.21930547949659)
--(axis cs:5.24,2.23556916097158)
--(axis cs:5.64,2.24034848417065)
--(axis cs:6.04,2.25357498704478)
--(axis cs:6.44,2.26827587336557)
--(axis cs:6.84,2.27372762410062)
--(axis cs:7.2402,2.28822872616516)
--(axis cs:7.6402,2.30576035996632)
--(axis cs:8.0402,2.31776981395157)
--(axis cs:8.4402,2.3333614715502)
--(axis cs:8.8402,2.34496072454544)
--(axis cs:9.2402,2.36598321709249)
--(axis cs:9.6402,2.38221614382569)
--(axis cs:10.0402,2.39684244525907)
--(axis cs:10.4402,2.41726033859302)
--(axis cs:10.8402,2.43359913972901)
--(axis cs:11.2402,2.45457621610202)
--(axis cs:11.6402,2.47134136493227)
--(axis cs:12.0402,2.49149590085408)
--(axis cs:12.4402,2.5109195613125)
--(axis cs:12.8402,2.53310805245021)
--(axis cs:13.2402,2.55343867597477)
--(axis cs:13.6402,2.57029870347163)
--(axis cs:14.0402,2.58586983076375)
--(axis cs:14.4404,2.60585618479351)
--(axis cs:14.8404,2.62296175168395)
--(axis cs:15.2404,2.63327168730673)
--(axis cs:15.6404,2.64191720185626)
--(axis cs:16.0404,2.65328566018274)
--(axis cs:16.4404,2.65892871984045)
--(axis cs:16.8404,2.66921831049865)
--(axis cs:17.2404,2.67534762255646)
--(axis cs:17.6404,2.67767938934042)
--(axis cs:18.0404,2.68137057202401)
--(axis cs:18.4404,2.68091324314188)
--(axis cs:18.8404,2.67946359632546)
--(axis cs:19.2404,2.68169734936054)
--(axis cs:19.6404,2.68010858500935)
--(axis cs:20.0404,2.67780553097959)
--(axis cs:20.4404,2.67435489368889)
--(axis cs:20.8404,2.67253292220658)
--(axis cs:21.2404,2.66834566904717)
--(axis cs:21.6404,2.66392120086646)
--(axis cs:22.0404,2.66227643026587)
--(axis cs:22.4404,2.65727891371481)
--(axis cs:22.8404,2.64881837102227)
--(axis cs:23.2404,2.64128557016591)
--(axis cs:23.6404,2.62883407067985)
--(axis cs:24.0404,2.61543044836288)
--(axis cs:24.4404,2.6047356177031)
--(axis cs:24.8404,2.59107722840801)
--(axis cs:25.2404,2.57635344102641)
--(axis cs:25.6404,2.55782195546303)
--(axis cs:26.0404,2.53922434522095)
--(axis cs:26.4404,2.5224435840503)
--(axis cs:26.8404,2.50144386918003)
--(axis cs:27.2404,2.48495472508508)
--(axis cs:27.6404,2.46544367363706)
--(axis cs:28.0404,2.44364981976807)
--(axis cs:28.4404,2.4232450773403)
--(axis cs:28.8404,2.40216557004301)
--(axis cs:29.2408,2.37612177968576)
--(axis cs:29.6414,2.35623881893311)
--(axis cs:30.0414,2.33059861301461)
--(axis cs:30.4414,2.30701597548508)
--(axis cs:30.8414,2.2791578374341)
--(axis cs:31.2414,2.25228419968689)
--(axis cs:31.6414,2.22810008344207)
--(axis cs:32.0414,2.20212690863908)
--(axis cs:32.4414,2.17470428154285)
--(axis cs:32.8416,2.14618337565667)
--(axis cs:33.2416,2.12214381854302)
--(axis cs:33.6416,2.09367911049123)
--(axis cs:34.0416,2.06494174151949)
--(axis cs:34.4416,2.03611408309673)
--(axis cs:34.8416,2.00414641760703)
--(axis cs:35.2416,1.97501255921494)
--(axis cs:35.6416,1.94397841134135)
--(axis cs:36.0416,1.90835835967875)
--(axis cs:36.4416,1.87361762960763)
--(axis cs:36.8416,1.84228164920328)
--(axis cs:37.2416,1.80609507690035)
--(axis cs:37.6402,1.7686697005533)
--(axis cs:38.0402,1.72631532637375)
--(axis cs:38.4402,1.68932835288141)
--(axis cs:38.8402,1.64717306510859)
--(axis cs:39.2402,1.60686723258339)
--(axis cs:39.6404,1.56333125378296)
--(axis cs:40.0412,1.52329701372557)
--(axis cs:40.4412,1.48073708175327)
--(axis cs:40.4412,6.66495932645202)
--(axis cs:40.4412,6.66495932645202)
--(axis cs:40.0412,6.68794326676086)
--(axis cs:39.6404,6.71315564113228)
--(axis cs:39.2402,6.73200317807735)
--(axis cs:38.8402,6.75479787394246)
--(axis cs:38.4402,6.77474828705446)
--(axis cs:38.0402,6.80266476667051)
--(axis cs:37.6402,6.82103521392914)
--(axis cs:37.2416,6.84432130795279)
--(axis cs:36.8416,6.86261041351008)
--(axis cs:36.4416,6.88908737175298)
--(axis cs:36.0416,6.90737817786447)
--(axis cs:35.6416,6.93030351901518)
--(axis cs:35.2416,6.94802846875979)
--(axis cs:34.8416,6.96688445892611)
--(axis cs:34.4416,6.98501210839801)
--(axis cs:34.0416,6.9989374818253)
--(axis cs:33.6416,7.01434763599084)
--(axis cs:33.2416,7.02870867627014)
--(axis cs:32.8416,7.04159630085741)
--(axis cs:32.4414,7.05337275233124)
--(axis cs:32.0414,7.06186286311624)
--(axis cs:31.6414,7.0735322385273)
--(axis cs:31.2414,7.07637190152936)
--(axis cs:30.8414,7.08002865325661)
--(axis cs:30.4414,7.08836112069784)
--(axis cs:30.0414,7.08252617857632)
--(axis cs:29.6414,7.08398365648419)
--(axis cs:29.2408,7.07842116928195)
--(axis cs:28.8404,7.0753402433365)
--(axis cs:28.4404,7.06631144930853)
--(axis cs:28.0404,7.05663910453563)
--(axis cs:27.6404,7.04626970305431)
--(axis cs:27.2404,7.03418894381274)
--(axis cs:26.8404,7.01901558572222)
--(axis cs:26.4404,7.0032181430885)
--(axis cs:26.0404,6.97850497667481)
--(axis cs:25.6404,6.95808535498892)
--(axis cs:25.2404,6.93782429745172)
--(axis cs:24.8404,6.91098085198299)
--(axis cs:24.4404,6.88712076768992)
--(axis cs:24.0404,6.86059154896362)
--(axis cs:23.6404,6.83484646058887)
--(axis cs:23.2404,6.80532503794755)
--(axis cs:22.8404,6.77481260659607)
--(axis cs:22.4404,6.74490944419916)
--(axis cs:22.0404,6.71404069791795)
--(axis cs:21.6404,6.68100232066134)
--(axis cs:21.2404,6.64717838074174)
--(axis cs:20.8404,6.6100887762738)
--(axis cs:20.4404,6.5732118311522)
--(axis cs:20.0404,6.53554079903058)
--(axis cs:19.6404,6.49645188689466)
--(axis cs:19.2404,6.45230328265044)
--(axis cs:18.8404,6.41191200700089)
--(axis cs:18.4404,6.36789050659406)
--(axis cs:18.0404,6.32299795699942)
--(axis cs:17.6404,6.27537612521761)
--(axis cs:17.2404,6.22830614164053)
--(axis cs:16.8404,6.18041587525993)
--(axis cs:16.4404,6.12827786834022)
--(axis cs:16.0404,6.07788904274606)
--(axis cs:15.6404,6.02381144457596)
--(axis cs:15.2404,5.9697605659067)
--(axis cs:14.8404,5.91216680090665)
--(axis cs:14.4404,5.85611764710887)
--(axis cs:14.0402,5.79584014290287)
--(axis cs:13.6402,5.73892980772173)
--(axis cs:13.2402,5.67462226703763)
--(axis cs:12.8402,5.61360653102337)
--(axis cs:12.4402,5.54688834913073)
--(axis cs:12.0402,5.48097949000068)
--(axis cs:11.6402,5.41480084242785)
--(axis cs:11.2402,5.34244689697282)
--(axis cs:10.8402,5.27136255999368)
--(axis cs:10.4402,5.19940249230729)
--(axis cs:10.0402,5.12458100740393)
--(axis cs:9.6402,5.04772104894112)
--(axis cs:9.2402,4.96912831866859)
--(axis cs:8.8402,4.89062597311169)
--(axis cs:8.4402,4.80665872772676)
--(axis cs:8.0402,4.72393491252427)
--(axis cs:7.6402,4.63967702201818)
--(axis cs:7.2402,4.55348158386203)
--(axis cs:6.84,4.46598291791058)
--(axis cs:6.44,4.37635466717872)
--(axis cs:6.04,4.28732228498222)
--(axis cs:5.64,4.19550036888402)
--(axis cs:5.24,4.10216639334981)
--(axis cs:4.84,4.00934076039117)
--(axis cs:4.44,3.91240309501397)
--(axis cs:4.04,3.81762702465077)
--(axis cs:3.64,3.7196294697832)
--(axis cs:3.24,3.62215820434178)
--(axis cs:2.84,3.52378399864742)
--(axis cs:2.44,3.42685813792715)
--(axis cs:2.04,3.32844506507702)
--(axis cs:1.64,3.23030854196707)
--(axis cs:1.24,3.13553778852796)
--(axis cs:0.84,3.03890877683368)
--(axis cs:0.44,2.94594736684357)
--(axis cs:0.04,2.90258873982827)
--cycle;

\path [fill=darkorange25512714, fill opacity=0.2]
(axis cs:0.04,2.87183881894133)
--(axis cs:0.04,2.4630503775569)
--(axis cs:0.44,2.42987710195295)
--(axis cs:0.84,2.32696777624854)
--(axis cs:1.24,2.22329521336877)
--(axis cs:1.6406,2.12094088201388)
--(axis cs:2.0406,2.01814578824772)
--(axis cs:2.4408,1.91466878169426)
--(axis cs:2.8408,1.81169075129448)
--(axis cs:3.241,1.70780404457903)
--(axis cs:3.641,1.60304125724754)
--(axis cs:4.041,1.50033201388901)
--(axis cs:4.441,1.3960468551702)
--(axis cs:4.841,1.29166033201018)
--(axis cs:5.241,1.18890546798458)
--(axis cs:5.641,1.08409408769243)
--(axis cs:6.041,0.984477159072902)
--(axis cs:6.441,0.879014314665322)
--(axis cs:6.841,0.78014857765414)
--(axis cs:7.241,0.680281585245633)
--(axis cs:7.641,0.591991898641127)
--(axis cs:8.0412,0.508617755127986)
--(axis cs:8.4412,0.440095369046577)
--(axis cs:8.8412,0.375746484083938)
--(axis cs:9.2412,0.327167293845289)
--(axis cs:9.6412,0.283854813885865)
--(axis cs:10.0412,0.250565996494934)
--(axis cs:10.4412,0.222750367065376)
--(axis cs:10.8414,0.19433874013876)
--(axis cs:11.2404,0.173055253801733)
--(axis cs:11.6324,0.149691856824868)
--(axis cs:12.0352,0.132541392841703)
--(axis cs:12.4352,0.122107161508304)
--(axis cs:12.8354,0.120444004846427)
--(axis cs:13.2316,0.123358538879449)
--(axis cs:13.6236,0.126845325615009)
--(axis cs:14.0198,0.130207804706314)
--(axis cs:14.4118,0.133482187584683)
--(axis cs:14.7998,0.135464864129435)
--(axis cs:15.1918,0.139277993409215)
--(axis cs:15.5878,0.143989432276829)
--(axis cs:15.9798,0.148965629417282)
--(axis cs:16.3786,0.157129226505282)
--(axis cs:16.7706,0.167149794241098)
--(axis cs:17.1658,0.179480090923243)
--(axis cs:17.5658,0.193816177917262)
--(axis cs:17.9658,0.210128818607473)
--(axis cs:18.3658,0.227078230647591)
--(axis cs:18.7658,0.244360822469044)
--(axis cs:19.1658,0.267841435745614)
--(axis cs:19.5658,0.28867304573846)
--(axis cs:19.9658,0.31247016419474)
--(axis cs:20.366,0.3375826972924)
--(axis cs:20.767,0.364767141354438)
--(axis cs:21.167,0.393420203640417)
--(axis cs:21.5672,0.423679004524133)
--(axis cs:21.9672,0.454788980955422)
--(axis cs:22.3672,0.487139645300552)
--(axis cs:22.7674,0.52069295844573)
--(axis cs:23.1594,0.551171507393689)
--(axis cs:23.5594,0.584060798456978)
--(axis cs:23.9514,0.614509972488648)
--(axis cs:24.3394,0.641861531715552)
--(axis cs:24.7354,0.666232536187859)
--(axis cs:25.1274,0.68488760271057)
--(axis cs:25.5274,0.706547615144087)
--(axis cs:25.9194,0.7254004847733)
--(axis cs:26.3194,0.744564569804197)
--(axis cs:26.6894,0.760600028118151)
--(axis cs:27.0784,0.781534735095209)
--(axis cs:27.4784,0.801420242050921)
--(axis cs:27.8786,0.821759521885798)
--(axis cs:28.2786,0.84231018695861)
--(axis cs:28.6746,0.863992813758894)
--(axis cs:29.0746,0.887000889008376)
--(axis cs:29.4706,0.908358548844331)
--(axis cs:29.866,0.931470290834588)
--(axis cs:30.262,0.954914733928666)
--(axis cs:30.654,0.979327670056638)
--(axis cs:31.05,1.00415976126608)
--(axis cs:31.45,1.02888455200668)
--(axis cs:31.846,1.05450939292717)
--(axis cs:32.246,1.08001876161196)
--(axis cs:32.646,1.10792763470223)
--(axis cs:33.0422,1.13678276426997)
--(axis cs:33.4422,1.16714968112438)
--(axis cs:33.8422,1.19832456991923)
--(axis cs:34.2302,1.23100367242618)
--(axis cs:34.6302,1.26421252840089)
--(axis cs:35.0302,1.29848927677199)
--(axis cs:35.4262,1.33345967970654)
--(axis cs:35.8262,1.36946721678113)
--(axis cs:36.2262,1.40659959515133)
--(axis cs:36.6264,1.44485827102777)
--(axis cs:37.0264,1.48384512891613)
--(axis cs:37.4264,1.52387290222411)
--(axis cs:37.8264,1.56375345479707)
--(axis cs:38.2224,1.60512769106124)
--(axis cs:38.6224,1.6465812609987)
--(axis cs:39.0224,1.68971474467408)
--(axis cs:39.4186,1.73304484014235)
--(axis cs:39.8174,1.77771605692607)
--(axis cs:40.2174,1.82308925998449)
--(axis cs:40.2174,3.60348024625164)
--(axis cs:40.2174,3.60348024625164)
--(axis cs:39.8174,3.51920881070879)
--(axis cs:39.4186,3.43761894491794)
--(axis cs:39.0224,3.35488232941012)
--(axis cs:38.6224,3.27140804433002)
--(axis cs:38.2224,3.19014350262405)
--(axis cs:37.8264,3.10557463026697)
--(axis cs:37.4264,3.01784450553811)
--(axis cs:37.0264,2.93118530795187)
--(axis cs:36.6264,2.8442043107105)
--(axis cs:36.2262,2.75465128717529)
--(axis cs:35.8262,2.66631652121853)
--(axis cs:35.4262,2.57522811531484)
--(axis cs:35.0302,2.48678617195407)
--(axis cs:34.6302,2.39702348964216)
--(axis cs:34.2302,2.3050194214905)
--(axis cs:33.8422,2.22100592869207)
--(axis cs:33.4422,2.13105906579796)
--(axis cs:33.0422,2.0427850117286)
--(axis cs:32.646,1.95809414280284)
--(axis cs:32.246,1.87221990822774)
--(axis cs:31.846,1.78855138438469)
--(axis cs:31.45,1.71002598293318)
--(axis cs:31.05,1.62929796585831)
--(axis cs:30.654,1.55408298047217)
--(axis cs:30.262,1.48288340955355)
--(axis cs:29.866,1.41330765058632)
--(axis cs:29.4706,1.34534353069802)
--(axis cs:29.0746,1.28020218811667)
--(axis cs:28.6746,1.21644957275609)
--(axis cs:28.2786,1.15541548356826)
--(axis cs:27.8786,1.09599470810281)
--(axis cs:27.4784,1.03926599568794)
--(axis cs:27.0784,0.985613139666369)
--(axis cs:26.6894,0.940550755498924)
--(axis cs:26.3194,0.893000587037406)
--(axis cs:25.9194,0.84809999532616)
--(axis cs:25.5274,0.804992909023939)
--(axis cs:25.1274,0.764221031383933)
--(axis cs:24.7354,0.727747557311822)
--(axis cs:24.3394,0.695565768439597)
--(axis cs:23.9514,0.669848676071675)
--(axis cs:23.5594,0.649012222813983)
--(axis cs:23.1594,0.6302024055161)
--(axis cs:22.7674,0.613638934154473)
--(axis cs:22.3672,0.600190767022067)
--(axis cs:21.9672,0.588726924255354)
--(axis cs:21.5672,0.577702071488674)
--(axis cs:21.167,0.566616621361144)
--(axis cs:20.767,0.556135563828658)
--(axis cs:20.366,0.544861116978649)
--(axis cs:19.9658,0.533455479315975)
--(axis cs:19.5658,0.521117444924741)
--(axis cs:19.1658,0.5084219785374)
--(axis cs:18.7658,0.496091320967006)
--(axis cs:18.3658,0.484039365586864)
--(axis cs:17.9658,0.471461568393346)
--(axis cs:17.5658,0.460612507318782)
--(axis cs:17.1658,0.449686223710709)
--(axis cs:16.7706,0.439465051415652)
--(axis cs:16.3786,0.42965688296281)
--(axis cs:15.9798,0.419838659398913)
--(axis cs:15.5878,0.409373371503822)
--(axis cs:15.1918,0.400575817121255)
--(axis cs:14.7998,0.39385178449962)
--(axis cs:14.4118,0.38718248973879)
--(axis cs:14.0198,0.381865098423376)
--(axis cs:13.6236,0.377386234907315)
--(axis cs:13.2316,0.37371799037788)
--(axis cs:12.8354,0.373417017306002)
--(axis cs:12.4352,0.373525258462158)
--(axis cs:12.0352,0.373374161578031)
--(axis cs:11.6324,0.379545532727509)
--(axis cs:11.2404,0.390318182890116)
--(axis cs:10.8414,0.410184049896533)
--(axis cs:10.4412,0.443862778172683)
--(axis cs:10.0412,0.491619730903549)
--(axis cs:9.6412,0.5546562462972)
--(axis cs:9.2412,0.626891722307905)
--(axis cs:8.8412,0.713690344166474)
--(axis cs:8.4412,0.802907270064861)
--(axis cs:8.0412,0.895693620903675)
--(axis cs:7.641,0.994259040769824)
--(axis cs:7.241,1.09440000528148)
--(axis cs:6.841,1.19827760412086)
--(axis cs:6.441,1.29862405710544)
--(axis cs:6.041,1.40002954010978)
--(axis cs:5.641,1.50365721797936)
--(axis cs:5.241,1.60677905774388)
--(axis cs:4.841,1.70835297476165)
--(axis cs:4.441,1.81188317860942)
--(axis cs:4.041,1.91344988733176)
--(axis cs:3.641,2.01603516264711)
--(axis cs:3.241,2.11778136126946)
--(axis cs:2.8408,2.21961073906128)
--(axis cs:2.4408,2.32119843319276)
--(axis cs:2.0406,2.42209718359677)
--(axis cs:1.6406,2.52402496366697)
--(axis cs:1.24,2.62488220994698)
--(axis cs:0.84,2.72547432697591)
--(axis cs:0.44,2.82592246364463)
--(axis cs:0.04,2.87183881894133)
--cycle;

\path [fill=forestgreen4416044, fill opacity=0.2]
(axis cs:0.04,2.87383052673912)
--(axis cs:0.04,2.41310996654103)
--(axis cs:0.44,2.36371732782988)
--(axis cs:0.84,2.26502110054338)
--(axis cs:1.228,2.16858859443546)
--(axis cs:1.6244,2.07191630096227)
--(axis cs:2.0136,1.97870880198691)
--(axis cs:2.4096,1.88155148779103)
--(axis cs:2.7976,1.78883611265221)
--(axis cs:3.1978,1.6897788426839)
--(axis cs:3.5938,1.59491392106064)
--(axis cs:3.9858,1.5013005657572)
--(axis cs:4.3778,1.4110372945205)
--(axis cs:4.7718,1.31667928405203)
--(axis cs:5.1698,1.22028052004653)
--(axis cs:5.5648,1.12412288702986)
--(axis cs:5.9568,1.03136339771092)
--(axis cs:6.3528,0.939923446543882)
--(axis cs:6.7488,0.849721844714774)
--(axis cs:7.1408,0.760063634871125)
--(axis cs:7.5372,0.673115760396259)
--(axis cs:7.9292,0.587695761887987)
--(axis cs:8.3132,0.505473140900811)
--(axis cs:8.7092,0.426248334522335)
--(axis cs:9.1052,0.353409014965465)
--(axis cs:9.5,0.299388421422248)
--(axis cs:9.9,0.263352498808025)
--(axis cs:10.292,0.241152087360839)
--(axis cs:10.688,0.228554174320443)
--(axis cs:11.088,0.222359106049517)
--(axis cs:11.484,0.219329892660704)
--(axis cs:11.8842,0.217609685430336)
--(axis cs:12.2842,0.217399353456936)
--(axis cs:12.6842,0.216330833187705)
--(axis cs:13.0802,0.215256858776141)
--(axis cs:13.4762,0.213167562281326)
--(axis cs:13.8762,0.21108171773547)
--(axis cs:14.2762,0.208599592684431)
--(axis cs:14.6762,0.205910515833175)
--(axis cs:15.0762,0.20301357442018)
--(axis cs:15.4762,0.199997218324155)
--(axis cs:15.8722,0.19687591494453)
--(axis cs:16.2682,0.193833511054324)
--(axis cs:16.6682,0.190751596755434)
--(axis cs:17.0642,0.187630964052071)
--(axis cs:17.463,0.184521926379327)
--(axis cs:17.863,0.181444036150507)
--(axis cs:18.263,0.178405365939438)
--(axis cs:18.663,0.175424400303612)
--(axis cs:19.063,0.172515053978969)
--(axis cs:19.467,0.169686360452957)
--(axis cs:19.867,0.166982372071792)
--(axis cs:20.267,0.16439308204269)
--(axis cs:20.667,0.161921530310467)
--(axis cs:21.067,0.159567089653222)
--(axis cs:21.467,0.157396489412505)
--(axis cs:21.867,0.155293451091331)
--(axis cs:22.267,0.15344683354982)
--(axis cs:22.667,0.151622219461543)
--(axis cs:23.067,0.149890539029375)
--(axis cs:23.467,0.148250035118675)
--(axis cs:23.867,0.146679372641904)
--(axis cs:24.267,0.145196066840243)
--(axis cs:24.667,0.143776254514833)
--(axis cs:25.067,0.142453639290233)
--(axis cs:25.467,0.141195375902485)
--(axis cs:25.867,0.139945873181392)
--(axis cs:26.267,0.138894534110961)
--(axis cs:26.6672,0.137867749784987)
--(axis cs:27.0632,0.136894451241064)
--(axis cs:27.4592,0.135981407168888)
--(axis cs:27.8472,0.135135843841888)
--(axis cs:28.2472,0.134319206667027)
--(axis cs:28.6472,0.133546741868045)
--(axis cs:29.0392,0.132818916356135)
--(axis cs:29.4352,0.132115819051671)
--(axis cs:29.8352,0.131451994108747)
--(axis cs:30.2312,0.130820173849812)
--(axis cs:30.6312,0.130200885826539)
--(axis cs:31.0312,0.129623187919205)
--(axis cs:31.4312,0.129075318956511)
--(axis cs:31.8272,0.128559609872286)
--(axis cs:32.2232,0.128072092199154)
--(axis cs:32.6152,0.12761838396163)
--(axis cs:33.0112,0.127183682986967)
--(axis cs:33.4112,0.126778713746256)
--(axis cs:33.8072,0.126395771504233)
--(axis cs:34.2072,0.126028230268593)
--(axis cs:34.6072,0.125650255001329)
--(axis cs:34.9992,0.125306462265605)
--(axis cs:35.3946,0.124967644941408)
--(axis cs:35.7952,0.124638193245271)
--(axis cs:36.188,0.124296357271492)
--(axis cs:36.588,0.124073594299025)
--(axis cs:36.98,0.123848751239076)
--(axis cs:37.376,0.123633374516794)
--(axis cs:37.776,0.123429019983125)
--(axis cs:38.176,0.123228480909762)
--(axis cs:38.576,0.123035955522233)
--(axis cs:38.976,0.122850063428113)
--(axis cs:39.376,0.122672281158992)
--(axis cs:39.776,0.122501300782997)
--(axis cs:40.1708,0.122339323839102)
--(axis cs:40.5712,0.122185297359875)
--(axis cs:40.5712,0.258497338320355)
--(axis cs:40.5712,0.258497338320355)
--(axis cs:40.1708,0.258619757794475)
--(axis cs:39.776,0.258751755761332)
--(axis cs:39.376,0.258895106606532)
--(axis cs:38.976,0.259053220064479)
--(axis cs:38.576,0.259222946289424)
--(axis cs:38.176,0.259404470998624)
--(axis cs:37.776,0.259597742411314)
--(axis cs:37.376,0.259805925695988)
--(axis cs:36.98,0.260030937119902)
--(axis cs:36.588,0.260272963344979)
--(axis cs:36.188,0.260536653759824)
--(axis cs:35.7952,0.260833931942114)
--(axis cs:35.3946,0.261156566584362)
--(axis cs:34.9992,0.261498138608976)
--(axis cs:34.6072,0.261837980578545)
--(axis cs:34.2072,0.262277245922727)
--(axis cs:33.8072,0.262682454041318)
--(axis cs:33.4112,0.263093759858589)
--(axis cs:33.0112,0.263529376599284)
--(axis cs:32.6152,0.263991492200644)
--(axis cs:32.2232,0.264467301407877)
--(axis cs:31.8272,0.264992606251549)
--(axis cs:31.4312,0.26554585821501)
--(axis cs:31.0312,0.266142136626957)
--(axis cs:30.6312,0.266776254017642)
--(axis cs:30.2312,0.267464211646633)
--(axis cs:29.8352,0.268190095100952)
--(axis cs:29.4352,0.268972045745748)
--(axis cs:29.0392,0.269817192204775)
--(axis cs:28.6472,0.270700849931335)
--(axis cs:28.2472,0.271675389864279)
--(axis cs:27.8472,0.2727354310036)
--(axis cs:27.4592,0.273853754039741)
--(axis cs:27.0632,0.27508027923653)
--(axis cs:26.6672,0.276417768089995)
--(axis cs:26.267,0.277870151052388)
--(axis cs:25.867,0.27944887035208)
--(axis cs:25.467,0.281142004080194)
--(axis cs:25.067,0.28302212403445)
--(axis cs:24.667,0.28508332498998)
--(axis cs:24.267,0.287348877557656)
--(axis cs:23.867,0.289813048671322)
--(axis cs:23.467,0.292508348547253)
--(axis cs:23.067,0.295421139917121)
--(axis cs:22.667,0.298586204269534)
--(axis cs:22.267,0.302029678386984)
--(axis cs:21.867,0.305738603105888)
--(axis cs:21.467,0.309805937280276)
--(axis cs:21.067,0.314214274214925)
--(axis cs:20.667,0.319028631931607)
--(axis cs:20.267,0.324253146125433)
--(axis cs:19.867,0.32992617375545)
--(axis cs:19.467,0.33608599382636)
--(axis cs:19.063,0.342857923975053)
--(axis cs:18.663,0.350040691158152)
--(axis cs:18.263,0.357772217011006)
--(axis cs:17.863,0.366074333181008)
--(axis cs:17.463,0.374969043519584)
--(axis cs:17.0642,0.384405484565907)
--(axis cs:16.6682,0.39432253438769)
--(axis cs:16.2682,0.405039454678061)
--(axis cs:15.8722,0.416165457133554)
--(axis cs:15.4762,0.42796949109109)
--(axis cs:15.0762,0.4405938304309)
--(axis cs:14.6762,0.45394829636623)
--(axis cs:14.2762,0.468013676039928)
--(axis cs:13.8762,0.482489948127441)
--(axis cs:13.4762,0.497554352346826)
--(axis cs:13.0802,0.513578622746844)
--(axis cs:12.6842,0.528305542246627)
--(axis cs:12.2842,0.544809314388734)
--(axis cs:11.8842,0.560286043134684)
--(axis cs:11.484,0.582986400424489)
--(axis cs:11.088,0.604536591737278)
--(axis cs:10.688,0.631395460701833)
--(axis cs:10.292,0.665879658079664)
--(axis cs:9.9,0.700680205639563)
--(axis cs:9.5,0.748314512436579)
--(axis cs:9.1052,0.80182459305381)
--(axis cs:8.7092,0.867172766467482)
--(axis cs:8.3132,0.937060723705467)
--(axis cs:7.9292,1.00881284054316)
--(axis cs:7.5372,1.08953111128861)
--(axis cs:7.1408,1.17801502212529)
--(axis cs:6.7488,1.26715788756437)
--(axis cs:6.3528,1.36039815465993)
--(axis cs:5.9568,1.45538121783727)
--(axis cs:5.5648,1.54834826796063)
--(axis cs:5.1698,1.64237732596149)
--(axis cs:4.7718,1.73700668545302)
--(axis cs:4.3778,1.83482940436361)
--(axis cs:3.9858,1.93500981495952)
--(axis cs:3.5938,2.03420281658486)
--(axis cs:3.1978,2.13335755913497)
--(axis cs:2.7976,2.23218535918845)
--(axis cs:2.4096,2.33137076881479)
--(axis cs:2.0136,2.43113061005971)
--(axis cs:1.6244,2.53072584278285)
--(axis cs:1.228,2.63121448265718)
--(axis cs:0.84,2.7273497813815)
--(axis cs:0.44,2.82704182130988)
--(axis cs:0.04,2.87383052673912)
--cycle;

\addplot [semithick, steelblue31119180]
table {%
0.04 2.52691579194712
0.44 2.5484207760956
0.84 2.59682776036049
1.24 2.64714078274533
1.64 2.69658917429945
2.04 2.74941586537282
2.44 2.80057482995905
2.84 2.85383243166879
3.24 2.90571511949256
3.64 2.95746016248818
4.04 3.01097479022132
4.44 3.06327727448214
4.84 3.11432311994388
5.24 3.16886777716069
5.64 3.21792442652734
6.04 3.2704486360135
6.44 3.32231527027215
6.84 3.3698552710056
7.2402 3.42085515501359
7.6402 3.47271869099225
8.0402 3.52085236323792
8.4402 3.57001009963848
8.8402 3.61779334882856
9.2402 3.66755576788054
9.6402 3.7149685963834
10.0402 3.7607117263315
10.4402 3.80833141545016
10.8402 3.85248084986134
11.2402 3.89851155653742
11.6402 3.94307110368006
12.0402 3.98623769542738
12.4402 4.02890395522162
12.8402 4.07335729173679
13.2402 4.1140304715062
13.6402 4.15461425559668
14.0402 4.19085498683331
14.4404 4.23098691595119
14.8404 4.2675642762953
15.2404 4.30151612660672
15.6404 4.33286432321611
16.0404 4.3655873514644
16.4404 4.39360329409034
16.8404 4.42481709287929
17.2404 4.4518268820985
17.6404 4.47652775727901
18.0404 4.50218426451172
18.4404 4.52440187486797
18.8404 4.54568780166317
19.2404 4.56700031600549
19.6404 4.58828023595201
20.0404 4.60667316500509
20.4404 4.62378336242054
20.8404 4.64131084924019
21.2404 4.65776202489446
21.6404 4.6724617607639
22.0404 4.68815856409191
22.4404 4.70109417895699
22.8404 4.71181548880917
23.2404 4.72330530405673
23.6404 4.73184026563436
24.0404 4.73801099866325
24.4404 4.74592819269651
24.8404 4.7510290401955
25.2404 4.75708886923907
25.6404 4.75795365522598
26.0404 4.75886466094788
26.4404 4.7628308635694
26.8404 4.76022972745112
27.2404 4.75957183444891
27.6404 4.75585668834568
28.0404 4.75014446215185
28.4404 4.74477826332441
28.8404 4.73875290668976
29.2408 4.72727147448386
29.6414 4.72011123770865
30.0414 4.70656239579547
30.4414 4.69768854809146
30.8414 4.67959324534536
31.2414 4.66432805060813
31.6414 4.65081616098469
32.0414 4.63199488587766
32.4414 4.61403851693705
32.8416 4.59388983825704
33.2416 4.57542624740658
33.6416 4.55401337324104
34.0416 4.53193961167239
34.4416 4.51056309574737
34.8416 4.48551543826657
35.2416 4.46152051398737
35.6416 4.43714096517827
36.0416 4.40786826877161
36.4416 4.3813525006803
36.8416 4.35244603135668
37.2416 4.32520819242657
37.6402 4.29485245724122
38.0402 4.26449004652213
38.4402 4.23203831996794
38.8402 4.20098546952552
39.2402 4.16943520533037
39.6404 4.13824344745762
40.0412 4.10562014024322
40.4412 4.07284820410264
};
% \addlegendentry{baseline}
\addplot [semithick, darkorange25512714]
table {%
0.04 2.66744459824911
0.44 2.62789978279879
0.84 2.52622105161222
1.24 2.42408871165787
1.6406 2.32248292284043
2.0406 2.22012148592224
2.4408 2.11793360744351
2.8408 2.01565074517788
3.241 1.91279270292424
3.641 1.80953820994732
4.041 1.70689095061039
4.441 1.60396501688981
4.841 1.50000665338591
5.241 1.39784226286423
5.641 1.2938756528359
6.041 1.19225334959134
6.441 1.08881918588538
6.841 0.989213090887502
7.241 0.887340795263558
7.641 0.793125469705475
8.0412 0.70215568801583
8.4412 0.621501319555719
8.8412 0.544718414125206
9.2412 0.477029508076597
9.6412 0.419255530091532
10.0412 0.371092863699242
10.4412 0.33330657261903
10.8414 0.302261395017647
11.2404 0.281686718345925
11.6324 0.264618694776189
12.0352 0.252957777209867
12.4352 0.247816209985231
12.8354 0.246930511076214
13.2316 0.248538264628664
13.6236 0.252115780261162
14.0198 0.256036451564845
14.4118 0.260332338661736
14.7998 0.264658324314527
15.1918 0.269926905265235
15.5878 0.276681401890326
15.9798 0.284402144408098
16.3786 0.293393054734046
16.7706 0.303307422828375
17.1658 0.314583157316976
17.5658 0.327214342618022
17.9658 0.340795193500409
18.3658 0.355558798117228
18.7658 0.370226071718025
19.1658 0.388131707141507
19.5658 0.4048952453316
19.9658 0.422962821755357
20.366 0.441221907135524
20.767 0.460451352591548
21.167 0.48001841250078
21.5672 0.500690538006404
21.9672 0.521757952605388
22.3672 0.543665206161309
22.7674 0.567165946300102
23.1594 0.590686956454894
23.5594 0.616536510635481
23.9514 0.642179324280162
24.3394 0.668713650077575
24.7354 0.69699004674984
25.1274 0.724554317047251
25.5274 0.755770262084013
25.9194 0.78675024004973
26.3194 0.818782578420802
26.6894 0.850575391808537
27.0784 0.883573937380789
27.4784 0.92034311886943
27.8786 0.958877114994303
28.2786 0.998862835263434
28.6746 1.04022119325749
29.0746 1.08360153856252
29.4706 1.12685103977118
29.866 1.17238897071045
30.262 1.21889907174111
30.654 1.2667053252644
31.05 1.3167288635622
31.45 1.36945526746993
31.846 1.42153038865593
32.246 1.47611933491985
32.646 1.53301088875253
33.0422 1.58978388799929
33.4422 1.64910437346117
33.8422 1.70966524930565
34.2302 1.76801154695834
34.6302 1.83061800902152
35.0302 1.89263772436303
35.4262 1.95434389751069
35.8262 2.01789186899983
36.2262 2.08062544116331
36.6264 2.14453129086914
37.0264 2.207515218434
37.4264 2.27085870388111
37.8264 2.33466404253202
38.2224 2.39763559684265
38.6224 2.45899465266436
39.0224 2.5222985370421
39.4186 2.58533189253014
39.8174 2.64846243381743
40.2174 2.71328475311806
};
% \addlegendentry{Mean M1}
\addplot [semithick, forestgreen4416044]
table {%
0.04 2.64347024664008
0.44 2.59537957456988
0.84 2.49618544096244
1.228 2.39990153854632
1.6244 2.30132107187256
2.0136 2.20491970602331
2.4096 2.10646112830291
2.7976 2.01051073592033
3.1978 1.91156820090944
3.5938 1.81455836882275
3.9858 1.71815519035836
4.3778 1.62293334944206
4.7718 1.52684298475252
5.1698 1.43132892300401
5.5648 1.33623557749524
5.9568 1.2433723077741
6.3528 1.1501608006019
6.7488 1.05843986613957
7.1408 0.969039328498207
7.5372 0.881323435842434
7.9292 0.798254301215573
8.3132 0.721266932303139
8.7092 0.646710550494909
9.1052 0.577616804009638
9.5 0.523851466929413
9.9 0.482016352223794
10.292 0.453515872720251
10.688 0.429974817511138
11.088 0.413447848893397
11.484 0.401158146542597
11.8842 0.38894786428251
12.2842 0.381104333922835
12.6842 0.372318187717166
13.0802 0.364417740761492
13.4762 0.355360957314076
13.8762 0.346785832931455
14.2762 0.338306634362179
14.6762 0.329929406099703
15.0762 0.32180370242554
15.4762 0.313983354707622
15.8722 0.306520686039042
16.2682 0.299436482866192
16.6682 0.292537065571562
17.0642 0.286018224308989
17.463 0.279745484949456
17.863 0.273759184665757
18.263 0.268088791475222
18.663 0.262732545730882
19.063 0.257686488977011
19.467 0.252886177139658
19.867 0.248454272913621
20.267 0.244323114084061
20.667 0.240475081121037
21.067 0.236890681934073
21.467 0.23360121334639
21.867 0.230516027098609
22.267 0.227738255968402
22.667 0.225104211865538
23.067 0.222655839473248
23.467 0.220379191832964
23.867 0.218246210656613
24.267 0.216272472198949
24.667 0.214429789752407
25.067 0.212737881662342
25.467 0.21116868999134
25.867 0.209697371766736
26.267 0.208382342581674
26.6672 0.207142758937491
27.0632 0.205987365238797
27.4592 0.204917580604315
27.8472 0.203935637422744
28.2472 0.202997298265653
28.6472 0.20212379589969
29.0392 0.201318054280455
29.4352 0.200543932398709
29.8352 0.19982104460485
30.2312 0.199142192748223
30.6312 0.198488569922091
31.0312 0.197882662273081
31.4312 0.19731058858576
31.8272 0.196776108061918
32.2232 0.196269696803515
32.6152 0.195804938081137
33.0112 0.195356529793126
33.4112 0.194936236802423
33.8072 0.194539112772775
34.2072 0.19415273809566
34.6072 0.193744117789937
34.9992 0.19340230043729
35.3946 0.193062105762885
35.7952 0.192736062593693
36.188 0.192416505515658
36.588 0.192173278822002
36.98 0.191939844179489
37.376 0.191719650106391
37.776 0.191513381197219
38.176 0.191316475954193
38.576 0.191129450905829
38.976 0.190951641746296
39.376 0.190783693882762
39.776 0.190626528272164
40.1708 0.190479540816789
40.5712 0.190341317840115
};
% \addlegendentry{Mean M2}
\end{axis}

\end{tikzpicture}

%% file: tex/Gazebo/e3p3.tex
% This file was created with tikzplotlib v0.10.1.
\begin{tikzpicture}

\definecolor{darkgray176}{RGB}{176,176,176}
\definecolor{darkorange25512714}{RGB}{255,127,14}
\definecolor{forestgreen4416044}{RGB}{44,160,44}
\definecolor{lightgray204}{RGB}{204,204,204}
\definecolor{steelblue31119180}{RGB}{31,119,180}

% \definecolor{darkslategray38}{RGB}{38,38,38}
% \definecolor{lavender234234242}{RGB}{234,234,242}
% \definecolor{lightgray204}{RGB}{204,204,204}
% \definecolor{mediumseagreen85168104}{RGB}{85,168,104}
% \definecolor{peru22113282}{RGB}{221,132,82}
% \definecolor{steelblue76114176}{RGB}{76,114,176}

% \begin{axis}[
%     width=\figurewidth,
%     height=\figureheight,
% legend style={at={(0.99,0.99)},anchor=north east, nodes={scale=0.8, transform shape}},
% axis background/.style={fill=lavender234234242},
% axis line style={white},
% tick align=outside,
% x grid style={white},
% major tick length=2.0,
% xmajorgrids,
% xmajorticks=true,
% % xmin=-1.96453596335, xmax=42.83319236035,
% xmin=-1, xmax=41,
% xtick style={color=darkslategray38},
% y grid style={white},
% % ylabel=\textcolor{darkslategray38}{Dist. to target (m)},
% yticklabels={},
% xlabel=\textcolor{darkslategray38}{Time (s)},
% ymajorgrids,
% ymajorticks=true,
% % ymin=0.37598902295362, ymax=3.71917413408861,
% ymin=-0.1, ymax=5.1,
% ytick style={color=darkslategray38},
% xtick distance=10,
% ytick distance=1
% ]
\definecolor{darkslategray38}{RGB}{38,38,38}
\definecolor{lightgray}{RGB}{192,192,192}
\definecolor{mediumseagreen85168104}{RGB}{85,168,104}
\definecolor{peru22113282}{RGB}{221,132,82}
\definecolor{steelblue76114176}{RGB}{76,114,176}
\begin{axis}[
    width=\figurewidth,
    height=\figureheight,
    axis background/.style={fill=white},
    axis line style={color=lightgray, line width=0.5pt},
    tick align=outside,
    x grid style={color=lightgray, opacity=0.3},
    y grid style={color=lightgray, opacity=0.3},
    major tick length=2.0,
    xmajorgrids,
    xmajorticks=true,
    xmin=-1, xmax=41,
    xtick style={color=lightgray},
    % xticklabels={},
    % ylabel=\textcolor{darkslategray38}{Dist. to target (m)},
    xlabel=\textcolor{darkslategray38}{Time (s)},
    ymajorgrids,
    ymajorticks=true,
    ymin=-0.1, ymax=5.1,
    ytick style={color=lightgray},
    xtick distance=10,
    ytick distance=1,
    % title=TARGET 1,
    % Scientific notation for y-axis if needed
    scaled y ticks=false,
    yticklabel style={
        /pgf/number format/fixed,
        /pgf/number format/precision=1
    },
]
\path [fill=steelblue31119180, fill opacity=0.2]
(axis cs:0.042,2.13344316046089)
--(axis cs:0.042,1.73540126731771)
--(axis cs:0.44,1.7030986193245)
--(axis cs:0.84,1.63509095272045)
--(axis cs:1.242,1.56679626956781)
--(axis cs:1.64,1.49931437253996)
--(axis cs:2.04,1.43078064516257)
--(axis cs:2.442,1.36016812610311)
--(axis cs:2.8408,1.29632051776635)
--(axis cs:3.2408,1.22586753734353)
--(axis cs:3.6368,1.15662428960087)
--(axis cs:4.0368,1.09233043196128)
--(axis cs:4.4368,1.02592270312074)
--(axis cs:4.8368,0.965158290326952)
--(axis cs:5.2368,0.898783444834554)
--(axis cs:5.6376,0.843977124684006)
--(axis cs:6.0376,0.798087686103069)
--(axis cs:6.4338,0.750738177582305)
--(axis cs:6.8338,0.710135880982879)
--(axis cs:7.2338,0.677560032861248)
--(axis cs:7.6298,0.638453434251657)
--(axis cs:8.0298,0.606700325056311)
--(axis cs:8.4298,0.579156314648603)
--(axis cs:8.8178,0.552376644197038)
--(axis cs:9.2178,0.522660502450719)
--(axis cs:9.6138,0.499199046103782)
--(axis cs:10.0124,0.471624132276934)
--(axis cs:10.4124,0.447528257215866)
--(axis cs:10.8084,0.424529909525977)
--(axis cs:11.2084,0.401133365153401)
--(axis cs:11.6084,0.382919812773626)
--(axis cs:12.0044,0.366116173766127)
--(axis cs:12.4044,0.355657505634578)
--(axis cs:12.8044,0.351699059226949)
--(axis cs:13.2004,0.349667170604011)
--(axis cs:13.6004,0.34986216524893)
--(axis cs:14.0002,0.350613246551732)
--(axis cs:14.4002,0.353367919237719)
--(axis cs:14.8002,0.355840759542283)
--(axis cs:15.2002,0.356148313729015)
--(axis cs:15.6002,0.357794042916049)
--(axis cs:16.0002,0.356114130928923)
--(axis cs:16.4002,0.361085543515999)
--(axis cs:16.8002,0.354601686790465)
--(axis cs:17.2002,0.357039527616187)
--(axis cs:17.6002,0.355874038008989)
--(axis cs:18.0006,0.352841047467329)
--(axis cs:18.4006,0.351812079506121)
--(axis cs:18.8008,0.346023794326135)
--(axis cs:19.1968,0.345763505689013)
--(axis cs:19.5928,0.340214268392593)
--(axis cs:19.9928,0.339402200508402)
--(axis cs:20.3928,0.33643704328585)
--(axis cs:20.7928,0.332401212356168)
--(axis cs:21.1882,0.32661204716872)
--(axis cs:21.5888,0.328113246476092)
--(axis cs:21.9848,0.328715822531535)
--(axis cs:22.3848,0.326276958621655)
--(axis cs:22.7808,0.325628491145786)
--(axis cs:23.1804,0.332833098798741)
--(axis cs:23.5764,0.339236717244629)
--(axis cs:23.9764,0.352213240048345)
--(axis cs:24.3764,0.368186493847169)
--(axis cs:24.7724,0.387213996917333)
--(axis cs:25.1686,0.407288694731727)
--(axis cs:25.5686,0.424334426143564)
--(axis cs:25.9686,0.446190576908211)
--(axis cs:26.3686,0.467113651450171)
--(axis cs:26.7686,0.492026507673792)
--(axis cs:27.169,0.514987183916729)
--(axis cs:27.569,0.536033380328545)
--(axis cs:27.969,0.558948997274925)
--(axis cs:28.357,0.582333577116546)
--(axis cs:28.757,0.604191530808404)
--(axis cs:29.157,0.624368369726678)
--(axis cs:29.553,0.643414121936873)
--(axis cs:29.953,0.661045536453518)
--(axis cs:30.353,0.67746084991828)
--(axis cs:30.753,0.693717478945255)
--(axis cs:31.153,0.706334203995765)
--(axis cs:31.553,0.719319438485913)
--(axis cs:31.9502,0.72919287452926)
--(axis cs:32.3502,0.742177967422014)
--(axis cs:32.7522,0.750660699743245)
--(axis cs:33.1502,0.760456131839013)
--(axis cs:33.551,0.766811954310668)
--(axis cs:33.951,0.773737259124931)
--(axis cs:34.351,0.779031394369028)
--(axis cs:34.751,0.784343814637771)
--(axis cs:35.151,0.787996598609148)
--(axis cs:35.551,0.790861931098183)
--(axis cs:35.951,0.792562340665763)
--(axis cs:36.351,0.793720379872323)
--(axis cs:36.751,0.795087456203374)
--(axis cs:37.1522,0.794723261652277)
--(axis cs:37.5482,0.791860352864885)
--(axis cs:37.9402,0.789389959588138)
--(axis cs:38.3362,0.786455162909111)
--(axis cs:38.7362,0.782131020057541)
--(axis cs:39.1364,0.778357312840569)
--(axis cs:39.5364,0.773139191564886)
--(axis cs:39.9324,0.766999811641072)
--(axis cs:40.3324,0.761015228880543)
--(axis cs:40.7336,0.754668913556809)
--(axis cs:40.7336,3.28195552414542)
--(axis cs:40.7336,3.28195552414542)
--(axis cs:40.3324,3.26390495405467)
--(axis cs:39.9324,3.24825882194941)
--(axis cs:39.5364,3.23357429585266)
--(axis cs:39.1364,3.21627901987893)
--(axis cs:38.7362,3.20775729238951)
--(axis cs:38.3362,3.192793809567)
--(axis cs:37.9402,3.18297096860384)
--(axis cs:37.5482,3.17153798364937)
--(axis cs:37.1522,3.15659313751821)
--(axis cs:36.751,3.14743463047434)
--(axis cs:36.351,3.14238684320959)
--(axis cs:35.951,3.13407411222555)
--(axis cs:35.551,3.12688939915205)
--(axis cs:35.151,3.11803170791351)
--(axis cs:34.751,3.10815523102368)
--(axis cs:34.351,3.10076130869056)
--(axis cs:33.951,3.08390502998657)
--(axis cs:33.551,3.07748853244356)
--(axis cs:33.1502,3.05825204860101)
--(axis cs:32.7522,3.04405723102763)
--(axis cs:32.3502,3.02850872647153)
--(axis cs:31.9502,3.00606057698415)
--(axis cs:31.553,2.98190869645717)
--(axis cs:31.153,2.96352525180944)
--(axis cs:30.753,2.9367452609774)
--(axis cs:30.353,2.91635909836247)
--(axis cs:29.953,2.88765497566874)
--(axis cs:29.553,2.8607929955709)
--(axis cs:29.157,2.83233349002473)
--(axis cs:28.757,2.79915645217477)
--(axis cs:28.357,2.77206953997876)
--(axis cs:27.969,2.73680494110164)
--(axis cs:27.569,2.70106984469106)
--(axis cs:27.169,2.66950263464685)
--(axis cs:26.7686,2.62764367813155)
--(axis cs:26.3686,2.59729741788319)
--(axis cs:25.9686,2.55782516367961)
--(axis cs:25.5686,2.51669276489164)
--(axis cs:25.1686,2.4834457083174)
--(axis cs:24.7724,2.4405640750613)
--(axis cs:24.3764,2.40639889886268)
--(axis cs:23.9764,2.36031476300873)
--(axis cs:23.5764,2.3214591812414)
--(axis cs:23.1804,2.27732999920685)
--(axis cs:22.7808,2.23370290568451)
--(axis cs:22.3848,2.18864141413039)
--(axis cs:21.9848,2.14319831527653)
--(axis cs:21.5888,2.09894604557815)
--(axis cs:21.1882,2.05234249966088)
--(axis cs:20.7928,2.00829464114672)
--(axis cs:20.3928,1.95943373366695)
--(axis cs:19.9928,1.91868245881559)
--(axis cs:19.5928,1.87293332287106)
--(axis cs:19.1968,1.83191067713144)
--(axis cs:18.8008,1.78393832326178)
--(axis cs:18.4006,1.74509487561801)
--(axis cs:18.0006,1.69947102724614)
--(axis cs:17.6002,1.66460200573319)
--(axis cs:17.2002,1.62807413124185)
--(axis cs:16.8002,1.59020938611883)
--(axis cs:16.4002,1.56010459103622)
--(axis cs:16.0002,1.53163460149936)
--(axis cs:15.6002,1.50403529946909)
--(axis cs:15.2002,1.47532456635101)
--(axis cs:14.8002,1.45607488770517)
--(axis cs:14.4002,1.43577801523465)
--(axis cs:14.0002,1.41806565508724)
--(axis cs:13.6004,1.40184554407401)
--(axis cs:13.2004,1.39341530290257)
--(axis cs:12.8044,1.38626296919901)
--(axis cs:12.4044,1.37996325697911)
--(axis cs:12.0044,1.37473773473342)
--(axis cs:11.6084,1.37498330170916)
--(axis cs:11.2084,1.37017687284546)
--(axis cs:10.8084,1.37140832945808)
--(axis cs:10.4124,1.37329313785072)
--(axis cs:10.0124,1.37625909772427)
--(axis cs:9.6138,1.38863662160564)
--(axis cs:9.2178,1.39313036969549)
--(axis cs:8.8178,1.41049735845951)
--(axis cs:8.4298,1.42869677290296)
--(axis cs:8.0298,1.4437290302103)
--(axis cs:7.6298,1.47092893056241)
--(axis cs:7.2338,1.49771170649892)
--(axis cs:6.8338,1.52449548857788)
--(axis cs:6.4338,1.55876578602753)
--(axis cs:6.0376,1.58921505420674)
--(axis cs:5.6376,1.6274987009364)
--(axis cs:5.2368,1.65430990692489)
--(axis cs:4.8368,1.68964211762349)
--(axis cs:4.4368,1.72225707193068)
--(axis cs:4.0368,1.75894600194045)
--(axis cs:3.6368,1.7956065978044)
--(axis cs:3.2408,1.83309425600147)
--(axis cs:2.8408,1.87077278622449)
--(axis cs:2.442,1.91154139678749)
--(axis cs:2.04,1.9510895050846)
--(axis cs:1.64,1.988348663684)
--(axis cs:1.242,2.03707716406007)
--(axis cs:0.84,2.06976785537093)
--(axis cs:0.44,2.12057464254101)
--(axis cs:0.042,2.13344316046089)
--cycle;

\path [fill=darkorange25512714, fill opacity=0.2]
(axis cs:0.042,2.01388211553534)
--(axis cs:0.042,1.80410339251337)
--(axis cs:0.4372,1.79568844089504)
--(axis cs:0.8292,1.75947980080339)
--(axis cs:1.2292,1.73386614777389)
--(axis cs:1.6252,1.70435105364037)
--(axis cs:2.0252,1.67269481289346)
--(axis cs:2.4132,1.63816563375342)
--(axis cs:2.808,1.60276795418112)
--(axis cs:3.208,1.56661480244991)
--(axis cs:3.604,1.52581262066667)
--(axis cs:4,1.48560951523798)
--(axis cs:4.4,1.44259630646184)
--(axis cs:4.796,1.39954850669505)
--(axis cs:5.196,1.3558016652533)
--(axis cs:5.596,1.31455307326293)
--(axis cs:5.9948,1.2703049048127)
--(axis cs:6.3948,1.22639154984747)
--(axis cs:6.795,1.17425091504063)
--(axis cs:7.195,1.12620143257382)
--(axis cs:7.595,1.07625062863326)
--(axis cs:7.9954,1.02658495699311)
--(axis cs:8.3954,0.976886022893812)
--(axis cs:8.7954,0.930562347053734)
--(axis cs:9.1954,0.885070123557606)
--(axis cs:9.5954,0.839317279993477)
--(axis cs:9.9954,0.797812916713276)
--(axis cs:10.3954,0.758242058492885)
--(axis cs:10.7954,0.720976986027001)
--(axis cs:11.1954,0.685853810139537)
--(axis cs:11.5954,0.651946242128368)
--(axis cs:11.9954,0.621634376186115)
--(axis cs:12.3954,0.591663290069817)
--(axis cs:12.7954,0.563526174211367)
--(axis cs:13.1954,0.534158847165807)
--(axis cs:13.5954,0.508865315261207)
--(axis cs:13.9954,0.484160144947322)
--(axis cs:14.3954,0.461599048249015)
--(axis cs:14.7954,0.445950796916491)
--(axis cs:15.1954,0.430217303214525)
--(axis cs:15.5954,0.420918251530191)
--(axis cs:15.9954,0.41466355574625)
--(axis cs:16.3954,0.411528451086255)
--(axis cs:16.7954,0.411773444955125)
--(axis cs:17.1958,0.414020351313273)
--(axis cs:17.5958,0.422860175192184)
--(axis cs:17.9958,0.431886837252641)
--(axis cs:18.3958,0.444404182423416)
--(axis cs:18.7958,0.454049989423522)
--(axis cs:19.1958,0.47149425673568)
--(axis cs:19.5958,0.486135029449628)
--(axis cs:19.9958,0.504188429228758)
--(axis cs:20.3958,0.520820765151157)
--(axis cs:20.7958,0.539155931581844)
--(axis cs:21.1958,0.556941440712306)
--(axis cs:21.5958,0.576057820989314)
--(axis cs:21.996,0.598733274232019)
--(axis cs:22.3962,0.617444027548383)
--(axis cs:22.7962,0.639187914293647)
--(axis cs:23.1962,0.658616362888408)
--(axis cs:23.5962,0.679441453451102)
--(axis cs:23.9964,0.700451463698289)
--(axis cs:24.3964,0.724852779583358)
--(axis cs:24.7964,0.747543227242916)
--(axis cs:25.1964,0.772374224434627)
--(axis cs:25.5964,0.794974515344464)
--(axis cs:25.9964,0.821483614470215)
--(axis cs:26.3964,0.84622502292666)
--(axis cs:26.7964,0.871989682648303)
--(axis cs:27.1964,0.89749234835183)
--(axis cs:27.5966,0.920171755366374)
--(axis cs:27.9966,0.945175888443161)
--(axis cs:28.3966,0.972533625554371)
--(axis cs:28.7968,0.998203003776189)
--(axis cs:29.1968,1.02518717791402)
--(axis cs:29.5968,1.05100593175847)
--(axis cs:29.9988,1.07628711943072)
--(axis cs:30.3968,1.10009742584879)
--(axis cs:30.7968,1.12580964496244)
--(axis cs:31.1968,1.14958199374777)
--(axis cs:31.5968,1.17332722065711)
--(axis cs:31.9968,1.19556725932973)
--(axis cs:32.3968,1.21741056435321)
--(axis cs:32.7972,1.23865467864178)
--(axis cs:33.1972,1.25863087062929)
--(axis cs:33.5972,1.27701778889831)
--(axis cs:33.9972,1.29219605623374)
--(axis cs:34.3972,1.30454793688381)
--(axis cs:34.7972,1.30639881652089)
--(axis cs:35.1972,1.30007065818825)
--(axis cs:35.5972,1.29066806587446)
--(axis cs:35.9972,1.27942154945024)
--(axis cs:36.4012,1.26664943679419)
--(axis cs:36.8012,1.25328033863985)
--(axis cs:37.2012,1.23889521435289)
--(axis cs:37.6012,1.22340745421329)
--(axis cs:38.0012,1.2073117770291)
--(axis cs:38.4012,1.19040741659899)
--(axis cs:38.8012,1.17384377175326)
--(axis cs:39.2012,1.15700648047786)
--(axis cs:39.6012,1.14040147352183)
--(axis cs:40.0012,1.1239604064856)
--(axis cs:40.4014,1.1070812065719)
--(axis cs:40.4014,1.34544888457382)
--(axis cs:40.4014,1.34544888457382)
--(axis cs:40.0012,1.35744563089689)
--(axis cs:39.6012,1.36681855598772)
--(axis cs:39.2012,1.37422141194007)
--(axis cs:38.8012,1.37951659763305)
--(axis cs:38.4012,1.38284683896032)
--(axis cs:38.0012,1.38553024401586)
--(axis cs:37.6012,1.38608630403912)
--(axis cs:37.2012,1.38458914269381)
--(axis cs:36.8012,1.38072822777066)
--(axis cs:36.4012,1.37403837168435)
--(axis cs:35.9972,1.36620891501533)
--(axis cs:35.5972,1.35688792540395)
--(axis cs:35.1972,1.34573427194953)
--(axis cs:34.7972,1.33481315146416)
--(axis cs:34.3972,1.33105498565201)
--(axis cs:33.9972,1.33366630811299)
--(axis cs:33.5972,1.33664267073321)
--(axis cs:33.1972,1.33907859696359)
--(axis cs:32.7972,1.34039956285438)
--(axis cs:32.3968,1.3403639682941)
--(axis cs:31.9968,1.33883080627985)
--(axis cs:31.5968,1.33624203810677)
--(axis cs:31.1968,1.33227031459484)
--(axis cs:30.7968,1.32646121770001)
--(axis cs:30.3968,1.31902494216899)
--(axis cs:29.9988,1.31074589597213)
--(axis cs:29.5968,1.30080982611544)
--(axis cs:29.1968,1.28960130068886)
--(axis cs:28.7968,1.27674421180846)
--(axis cs:28.3966,1.26361147241629)
--(axis cs:27.9966,1.2486457226498)
--(axis cs:27.5966,1.23412474038775)
--(axis cs:27.1964,1.21919504983752)
--(axis cs:26.7964,1.20337743351559)
--(axis cs:26.3964,1.18579873353205)
--(axis cs:25.9964,1.16681389164525)
--(axis cs:25.5964,1.1466086710932)
--(axis cs:25.1964,1.12694867194322)
--(axis cs:24.7964,1.10492651930591)
--(axis cs:24.3964,1.08334904224236)
--(axis cs:23.9964,1.06157551483601)
--(axis cs:23.5962,1.03861140255073)
--(axis cs:23.1962,1.01557256105792)
--(axis cs:22.7962,0.993096699901151)
--(axis cs:22.3962,0.969547140153667)
--(axis cs:21.996,0.94725497850036)
--(axis cs:21.5958,0.922885590272947)
--(axis cs:21.1958,0.899877719483015)
--(axis cs:20.7958,0.877357207232875)
--(axis cs:20.3958,0.855334899299443)
--(axis cs:19.9958,0.832769033755225)
--(axis cs:19.5958,0.810925996076047)
--(axis cs:19.1958,0.787183353671905)
--(axis cs:18.7958,0.769589487620005)
--(axis cs:18.3958,0.745099495043372)
--(axis cs:17.9958,0.726653478806574)
--(axis cs:17.5958,0.711698744963489)
--(axis cs:17.1958,0.694903841842296)
--(axis cs:16.7954,0.67955584264043)
--(axis cs:16.3954,0.665284649289947)
--(axis cs:15.9954,0.651869834359784)
--(axis cs:15.5954,0.641120068679036)
--(axis cs:15.1954,0.633042276146299)
--(axis cs:14.7954,0.62809726365885)
--(axis cs:14.3954,0.6257102562758)
--(axis cs:13.9954,0.627332262629257)
--(axis cs:13.5954,0.633969367302509)
--(axis cs:13.1954,0.646138571734754)
--(axis cs:12.7954,0.664765553538043)
--(axis cs:12.3954,0.687421862022126)
--(axis cs:11.9954,0.716493648598848)
--(axis cs:11.5954,0.752743864750018)
--(axis cs:11.1954,0.792539464843772)
--(axis cs:10.7954,0.838469094990131)
--(axis cs:10.3954,0.885857281642522)
--(axis cs:9.9954,0.934873991118465)
--(axis cs:9.5954,0.984323435999316)
--(axis cs:9.1954,1.03517495820811)
--(axis cs:8.7954,1.08404816058441)
--(axis cs:8.3954,1.13515987936691)
--(axis cs:7.9954,1.18410670251493)
--(axis cs:7.595,1.23377069667299)
--(axis cs:7.195,1.2833424715761)
--(axis cs:6.795,1.32875028714038)
--(axis cs:6.3948,1.37459481668844)
--(axis cs:5.9948,1.41838834232537)
--(axis cs:5.596,1.46372616626824)
--(axis cs:5.196,1.5095868843054)
--(axis cs:4.796,1.55355320838625)
--(axis cs:4.4,1.59749422053879)
--(axis cs:4,1.6394261495712)
--(axis cs:3.604,1.6811390922425)
--(axis cs:3.208,1.72163568863954)
--(axis cs:2.808,1.76187112350531)
--(axis cs:2.4132,1.80221545948843)
--(axis cs:2.0252,1.84134911381464)
--(axis cs:1.6252,1.88167822467323)
--(axis cs:1.2292,1.91962791720687)
--(axis cs:0.8292,1.95626371695254)
--(axis cs:0.4372,1.99604944751776)
--(axis cs:0.042,2.01388211553534)
--cycle;

\path [fill=forestgreen4416044, fill opacity=0.2]
(axis cs:0.042,2.10476015648029)
--(axis cs:0.042,1.81461567506895)
--(axis cs:0.4408,1.79039280840269)
--(axis cs:0.8408,1.74715687018268)
--(axis cs:1.2408,1.70250767828498)
--(axis cs:1.6408,1.65722455691572)
--(axis cs:2.0414,1.61812944670062)
--(axis cs:2.4414,1.56901275642689)
--(axis cs:2.8414,1.52749126389168)
--(axis cs:3.2414,1.48514279236622)
--(axis cs:3.6414,1.44493441032184)
--(axis cs:4.0414,1.40179174137808)
--(axis cs:4.4414,1.36588693593486)
--(axis cs:4.8416,1.32808172121541)
--(axis cs:5.2416,1.29514172544722)
--(axis cs:5.6416,1.25479509842415)
--(axis cs:6.0422,1.22208590401152)
--(axis cs:6.4424,1.18880168812945)
--(axis cs:6.8424,1.15779785323538)
--(axis cs:7.2424,1.11574914931094)
--(axis cs:7.6424,1.0864590381744)
--(axis cs:8.0424,1.04735532683921)
--(axis cs:8.4424,1.01325585843725)
--(axis cs:8.8424,0.974568451035328)
--(axis cs:9.2424,0.941040242808971)
--(axis cs:9.6424,0.901883701804583)
--(axis cs:10.0424,0.870778781161164)
--(axis cs:10.4424,0.832589235177935)
--(axis cs:10.8424,0.79689991413984)
--(axis cs:11.2424,0.760700759319291)
--(axis cs:11.6424,0.723311956184207)
--(axis cs:12.0424,0.685940694876804)
--(axis cs:12.4424,0.655074310569962)
--(axis cs:12.8424,0.619659157133012)
--(axis cs:13.2424,0.589572725268469)
--(axis cs:13.6424,0.559724548042716)
--(axis cs:14.0424,0.52992939631336)
--(axis cs:14.4424,0.502338686670827)
--(axis cs:14.8424,0.475388573856437)
--(axis cs:15.2424,0.454057291191549)
--(axis cs:15.6424,0.430024106416485)
--(axis cs:16.0424,0.407401305194907)
--(axis cs:16.4424,0.384932733365017)
--(axis cs:16.8424,0.364960289540622)
--(axis cs:17.2424,0.34546300248053)
--(axis cs:17.6424,0.327683249734986)
--(axis cs:18.0424,0.309861186725603)
--(axis cs:18.4424,0.295458258634714)
--(axis cs:18.8424,0.280933188286171)
--(axis cs:19.2424,0.267787086271641)
--(axis cs:19.6424,0.255397433659149)
--(axis cs:20.0424,0.243915759639212)
--(axis cs:20.4424,0.233075700837009)
--(axis cs:20.8424,0.22225452597421)
--(axis cs:21.2424,0.212443495609435)
--(axis cs:21.6424,0.204094381357686)
--(axis cs:22.0424,0.196141538095501)
--(axis cs:22.4424,0.188592142379519)
--(axis cs:22.8424,0.1817720048317)
--(axis cs:23.2424,0.175354205162362)
--(axis cs:23.6424,0.16945053910882)
--(axis cs:24.0424,0.164066173726513)
--(axis cs:24.4424,0.15918445183224)
--(axis cs:24.8424,0.154541525008844)
--(axis cs:25.2426,0.150930376005834)
--(axis cs:25.6426,0.147720283650351)
--(axis cs:26.0426,0.145050178825683)
--(axis cs:26.4426,0.142833316122736)
--(axis cs:26.8426,0.140440058957205)
--(axis cs:27.2426,0.138979863833007)
--(axis cs:27.6426,0.137392549095334)
--(axis cs:28.0426,0.136006630263732)
--(axis cs:28.4426,0.134988447131156)
--(axis cs:28.8426,0.133852824171496)
--(axis cs:29.2426,0.133277709503039)
--(axis cs:29.6426,0.132797331556418)
--(axis cs:30.0426,0.13240174553981)
--(axis cs:30.4426,0.132078241379177)
--(axis cs:30.8428,0.132014271173138)
--(axis cs:31.2346,0.132003066452182)
--(axis cs:31.6346,0.132080003931294)
--(axis cs:32.0346,0.132220976588512)
--(axis cs:32.4346,0.132409825811046)
--(axis cs:32.8346,0.132635665256202)
--(axis cs:33.2308,0.13289849910491)
--(axis cs:33.6308,0.133169327098913)
--(axis cs:34.0188,0.133458950207987)
--(axis cs:34.4148,0.133755679821104)
--(axis cs:34.8108,0.134083920465353)
--(axis cs:35.2108,0.134437241625237)
--(axis cs:35.6108,0.134765604714342)
--(axis cs:36.0108,0.135182257749867)
--(axis cs:36.4108,0.135672516922679)
--(axis cs:36.8108,0.136103492116571)
--(axis cs:37.2028,0.136506721110399)
--(axis cs:37.6028,0.136877856154908)
--(axis cs:38.0028,0.137269355910398)
--(axis cs:38.3988,0.137636125085163)
--(axis cs:38.7988,0.137988362780923)
--(axis cs:39.1936,0.138311306425398)
--(axis cs:39.5936,0.138633004913778)
--(axis cs:39.9936,0.13894977951532)
--(axis cs:40.39,0.13924969345571)
--(axis cs:40.7896,0.139528424857173)
--(axis cs:40.7896,0.324884637474081)
--(axis cs:40.7896,0.324884637474081)
--(axis cs:40.39,0.324701924608217)
--(axis cs:39.9936,0.324519728195134)
--(axis cs:39.5936,0.324314705330941)
--(axis cs:39.1936,0.324106977291303)
--(axis cs:38.7988,0.323886747238786)
--(axis cs:38.3988,0.323631324712527)
--(axis cs:38.0028,0.323399968374811)
--(axis cs:37.6028,0.323154484610415)
--(axis cs:37.2028,0.322893679229455)
--(axis cs:36.8108,0.322635718328295)
--(axis cs:36.4108,0.322350937552546)
--(axis cs:36.0108,0.322066709579833)
--(axis cs:35.6108,0.321785944862235)
--(axis cs:35.2108,0.321517006024325)
--(axis cs:34.8108,0.321252259195641)
--(axis cs:34.4148,0.320995467964826)
--(axis cs:34.0188,0.320754841944665)
--(axis cs:33.6308,0.320519270986383)
--(axis cs:33.2308,0.320300444612274)
--(axis cs:32.8346,0.320093869386462)
--(axis cs:32.4346,0.319925990875453)
--(axis cs:32.0346,0.319782905248203)
--(axis cs:31.6346,0.319678997480294)
--(axis cs:31.2346,0.319638392516202)
--(axis cs:30.8428,0.319599582608547)
--(axis cs:30.4426,0.319781422205739)
--(axis cs:30.0426,0.319616678736289)
--(axis cs:29.6426,0.319734308324498)
--(axis cs:29.2426,0.320263295912979)
--(axis cs:28.8426,0.320711039463901)
--(axis cs:28.4426,0.321394271422873)
--(axis cs:28.0426,0.322054690671142)
--(axis cs:27.6426,0.323031593566109)
--(axis cs:27.2426,0.324138729519466)
--(axis cs:26.8426,0.325405299852062)
--(axis cs:26.4426,0.327097761327833)
--(axis cs:26.0426,0.328938663746838)
--(axis cs:25.6426,0.331115900106483)
--(axis cs:25.2426,0.333631617481527)
--(axis cs:24.8424,0.336520140794905)
--(axis cs:24.4424,0.339995793480414)
--(axis cs:24.0424,0.343619536115442)
--(axis cs:23.6424,0.348316073241254)
--(axis cs:23.2424,0.353688424891654)
--(axis cs:22.8424,0.359797611013549)
--(axis cs:22.4424,0.366520984986141)
--(axis cs:22.0424,0.374264884677784)
--(axis cs:21.6424,0.382653014643906)
--(axis cs:21.2424,0.391904178025262)
--(axis cs:20.8424,0.401834978700173)
--(axis cs:20.4424,0.413215495804753)
--(axis cs:20.0424,0.426502075432538)
--(axis cs:19.6424,0.440169192834124)
--(axis cs:19.2424,0.456145294290648)
--(axis cs:18.8424,0.472715163366499)
--(axis cs:18.4424,0.491807798644165)
--(axis cs:18.0424,0.509817780355156)
--(axis cs:17.6424,0.534657002360905)
--(axis cs:17.2424,0.554622461184782)
--(axis cs:16.8424,0.581360869751982)
--(axis cs:16.4424,0.606574384467854)
--(axis cs:16.0424,0.63449043902364)
--(axis cs:15.6424,0.662554555646542)
--(axis cs:15.2424,0.693182170482618)
--(axis cs:14.8424,0.724444652170529)
--(axis cs:14.4424,0.757425577538895)
--(axis cs:14.0424,0.79113580366617)
--(axis cs:13.6424,0.823892429216531)
--(axis cs:13.2424,0.861649475098361)
--(axis cs:12.8424,0.89731465207091)
--(axis cs:12.4424,0.934019527380285)
--(axis cs:12.0424,0.9722976460073)
--(axis cs:11.6424,1.01046379581891)
--(axis cs:11.2424,1.04786294697775)
--(axis cs:10.8424,1.08767408982573)
--(axis cs:10.4424,1.12387702123272)
--(axis cs:10.0424,1.16318655561061)
--(axis cs:9.6424,1.20084669609413)
--(axis cs:9.2424,1.23488997045868)
--(axis cs:8.8424,1.27171486457102)
--(axis cs:8.4424,1.30868710849935)
--(axis cs:8.0424,1.34650091682195)
--(axis cs:7.6424,1.38168870547041)
--(axis cs:7.2424,1.41981154281096)
--(axis cs:6.8424,1.45537554352199)
--(axis cs:6.4424,1.49120361573157)
--(axis cs:6.0422,1.52940617161736)
--(axis cs:5.6416,1.56662094924431)
--(axis cs:5.2416,1.60014716153345)
--(axis cs:4.8416,1.6417982957173)
--(axis cs:4.4414,1.67616775570031)
--(axis cs:4.0414,1.7162905329963)
--(axis cs:3.6414,1.75217427032498)
--(axis cs:3.2414,1.79445981342164)
--(axis cs:2.8414,1.83540755413754)
--(axis cs:2.4414,1.8769713687929)
--(axis cs:2.0414,1.92185233077819)
--(axis cs:1.6408,1.96327963150634)
--(axis cs:1.2408,2.00708272954032)
--(axis cs:0.8408,2.04622398432934)
--(axis cs:0.4408,2.08663429467326)
--(axis cs:0.042,2.10476015648029)
--cycle;

\addplot [semithick, steelblue31119180]
table {%
0.042 1.9344222138893
0.44 1.91183663093275
0.84 1.85242940404569
1.242 1.80193671681394
1.64 1.74383151811198
2.04 1.69093507512359
2.442 1.6358547614453
2.8408 1.58354665199542
3.2408 1.5294808966725
3.6368 1.47611544370264
4.0368 1.42563821695087
4.4368 1.37408988752571
4.8368 1.32740020397522
5.2368 1.27654667587972
5.6376 1.2357379128102
6.0376 1.19365137015491
6.4338 1.15475198180492
6.8338 1.11731568478038
7.2338 1.08763586968008
7.6298 1.05469118240703
8.0298 1.02521467763331
8.4298 1.00392654377578
8.8178 0.981437001328275
9.2178 0.957895436073107
9.6138 0.94391783385471
10.0124 0.9239416150006
10.4124 0.910410697533294
10.8084 0.897969119492028
11.2084 0.885655118999431
11.6084 0.878951557241394
12.0044 0.870426954249775
12.4044 0.867810381306843
12.8044 0.868981014212979
13.2004 0.871541236753292
13.6004 0.875853854661472
14.0002 0.884339450819487
14.4002 0.894572967236186
14.8002 0.905957823623727
15.2002 0.915736440040014
15.6002 0.930914671192568
16.0002 0.94387436621414
16.4002 0.960595067276107
16.8002 0.972405536454649
17.2002 0.992556829429021
17.6002 1.01023802187109
18.0006 1.02615603735673
18.4006 1.04845347756207
18.8008 1.06498105879396
19.1968 1.08883709141023
19.5928 1.10657379563183
19.9928 1.129042329662
20.3928 1.1479353884764
20.7928 1.17034792675144
21.1882 1.1894772734148
21.5888 1.21352964602712
21.9848 1.23595706890403
22.3848 1.25745918637602
22.7808 1.27966569841515
23.1804 1.3050815490028
23.5764 1.33034794924301
23.9764 1.35626400152854
24.3764 1.38729269635492
24.7724 1.41388903598932
25.1686 1.44536720152456
25.5686 1.4705135955176
25.9686 1.50200787029391
26.3686 1.53220553466668
26.7686 1.55983509290267
27.169 1.59224490928179
27.569 1.6185516125098
27.969 1.64787696918828
28.357 1.67720155854765
28.757 1.70167399149159
29.157 1.7283509298757
29.553 1.75210355875389
29.953 1.77435025606113
30.353 1.79690997414038
30.753 1.81523136996133
31.153 1.8349297279026
31.553 1.85061406747154
31.9502 1.86762672575671
32.3502 1.88534334694677
32.7522 1.89735896538544
33.1502 1.90935409022001
33.551 1.92215024337711
33.951 1.92882114455575
34.351 1.93989635152979
34.751 1.94624952283073
35.151 1.95301415326133
35.551 1.95887566512512
35.951 1.96331822644566
36.351 1.96805361154096
36.751 1.97126104333886
37.1522 1.97565819958524
37.5482 1.98169916825713
37.9402 1.98618046409599
38.3362 1.98962448623805
38.7362 1.99494415622353
39.1364 1.99731816635975
39.5364 2.00335674370877
39.9324 2.00762931679524
40.3324 2.01246009146761
40.7336 2.01831221885111
};
% \addlegendentry{baseline}
\addplot [semithick, darkorange25512714]
table {%
0.042 1.90899275402436
0.4372 1.8958689442064
0.8292 1.85787175887797
1.2292 1.82674703249038
1.6252 1.7930146391568
2.0252 1.75702196335405
2.4132 1.72019054662092
2.808 1.68231953884321
3.208 1.64412524554473
3.604 1.60347585645459
4 1.56251783240459
4.4 1.52004526350032
4.796 1.47655085754065
5.196 1.43269427477935
5.596 1.38913961976558
5.9948 1.34434662356903
6.3948 1.30049318326795
6.795 1.2515006010905
7.195 1.20477195207496
7.595 1.15501066265312
7.9954 1.10534582975402
8.3954 1.05602295113036
8.7954 1.00730525381907
9.1954 0.96012254088286
9.5954 0.911820357996397
9.9954 0.866343453915871
10.3954 0.822049670067704
10.7954 0.779723040508566
11.1954 0.739196637491655
11.5954 0.702345053439193
11.9954 0.669064012392481
12.3954 0.639542576045971
12.7954 0.614145863874705
13.1954 0.59014870945028
13.5954 0.571417341281858
13.9954 0.555746203788289
14.3954 0.543654652262407
14.7954 0.53702403028767
15.1954 0.531629789680412
15.5954 0.531019160104614
15.9954 0.533266695053017
16.3954 0.538406550188101
16.7954 0.545664643797778
17.1958 0.554462096577784
17.5958 0.567279460077836
17.9958 0.579270158029608
18.3958 0.594751838733394
18.7958 0.611819738521764
19.1958 0.629338805203793
19.5958 0.648530512762837
19.9958 0.668478731491992
20.3958 0.6880778322253
20.7958 0.708256569407359
21.1958 0.728409580097661
21.5958 0.74947170563113
21.996 0.772994126366189
22.3962 0.793495583851025
22.7962 0.816142307097399
23.1962 0.837094461973166
23.5962 0.859026428000917
23.9964 0.881013489267152
24.3964 0.904100910912861
24.7964 0.926234873274413
25.1964 0.949661448188922
25.5964 0.970791593218833
25.9964 0.994148753057733
26.3964 1.01601187822936
26.7964 1.03768355808195
27.1964 1.05834369909468
27.5966 1.07714824787706
27.9966 1.09691080554648
28.3966 1.11807254898533
28.7968 1.13747360779233
29.1968 1.15739423930144
29.5968 1.17590787893696
29.9988 1.19351650770143
30.3968 1.20956118400889
30.7968 1.22613543133123
31.1968 1.24092615417131
31.5968 1.25478462938194
31.9968 1.26719903280479
32.3968 1.27888726632365
32.7972 1.28952712074808
33.1972 1.29885473379644
33.5972 1.30683022981576
33.9972 1.31293118217337
34.3972 1.31780146126791
34.7972 1.32060598399252
35.1972 1.32290246506889
35.5972 1.32377799563921
35.9972 1.32281523223279
36.4012 1.32034390423927
36.8012 1.31700428320525
37.2012 1.31174217852335
37.6012 1.30474687912621
38.0012 1.29642101052248
38.4012 1.28662712777965
38.8012 1.27668018469315
39.2012 1.26561394620897
39.6012 1.25361001475477
40.0012 1.24070301869125
40.4014 1.22626504557286
};
% \addlegendentry{Mean M1}
\addplot [semithick, forestgreen4416044]
table {%
0.042 1.95968791577462
0.4408 1.93851355153797
0.8408 1.89669042725601
1.2408 1.85479520391265
1.6408 1.81025209421103
2.0414 1.76999088873941
2.4414 1.7229920626099
2.8414 1.68144940901461
3.2414 1.63980130289393
3.6414 1.59855434032341
4.0414 1.55904113718719
4.4414 1.52102734581759
4.8416 1.48494000846635
5.2416 1.44764444349034
5.6416 1.41070802383423
6.0422 1.37574603781444
6.4424 1.34000265193051
6.8424 1.30658669837868
7.2424 1.26778034606095
7.6424 1.23407387182241
8.0424 1.19692812183058
8.4424 1.1609714834683
8.8424 1.12314165780317
9.2424 1.08796510663382
9.6424 1.05136519894936
10.0424 1.01698266838589
10.4424 0.978233128205329
10.8424 0.942287001982784
11.2424 0.904281853148519
11.6424 0.866887876001558
12.0424 0.829119170442052
12.4424 0.794546918975123
12.8424 0.758486904601961
13.2424 0.725611100183415
13.6424 0.691808488629624
14.0424 0.660532599989765
14.4424 0.629882132104861
14.8424 0.599916613013483
15.2424 0.573619730837083
15.6424 0.546289331031513
16.0424 0.520945872109274
16.4424 0.495753558916435
16.8424 0.473160579646302
17.2424 0.450042731832656
17.6424 0.431170126047946
18.0424 0.40983948354038
18.4424 0.393633028639439
18.8424 0.376824175826335
19.2424 0.361966190281144
19.6424 0.347783313246636
20.0424 0.335208917535875
20.4424 0.323145598320881
20.8424 0.312044752337192
21.2424 0.302173836817349
21.6424 0.293373698000796
22.0424 0.285203211386642
22.4424 0.27755656368283
22.8424 0.270784807922624
23.2424 0.264521315027008
23.6424 0.258883306175037
24.0424 0.253842854920978
24.4424 0.249590122656327
24.8424 0.245530832901874
25.2426 0.242280996743681
25.6426 0.239418091878417
26.0426 0.236994421286261
26.4426 0.234965538725284
26.8426 0.232922679404633
27.2426 0.231559296676236
27.6426 0.230212071330721
28.0426 0.229030660467437
28.4426 0.228191359277015
28.8426 0.227281931817699
29.2426 0.226770502708009
29.6426 0.226265819940458
30.0426 0.22600921213805
30.4426 0.225929831792458
30.8428 0.225806926890843
31.2346 0.225820729484192
31.6346 0.225879500705794
32.0346 0.226001940918357
32.4346 0.22616790834325
32.8346 0.226364767321332
33.2308 0.226599471858592
33.6308 0.226844299042648
34.0188 0.227106896076326
34.4148 0.227375573892965
34.8108 0.227668089830497
35.2108 0.227977123824781
35.6108 0.228275774788289
36.0108 0.22862448366485
36.4108 0.229011727237613
36.8108 0.229369605222433
37.2028 0.229700200169927
37.6028 0.230016170382662
38.0028 0.230334662142604
38.3988 0.230633724898845
38.7988 0.230937555009854
39.1936 0.231209141858351
39.5936 0.23147385512236
39.9936 0.231734753855227
40.39 0.231975809031963
40.7896 0.232206531165627
};
% \addlegendentry{Mean M2}
\end{axis}

\end{tikzpicture}